\documentclass[sigconf, nonacm]{acmart}

\newcommand\vldbdoi{10.14778/3746405.3746407}
\newcommand\vldbpages{2775 - 2789}
\newcommand\vldbvolume{18}
\newcommand\vldbissue{9}
\newcommand\vldbyear{2025}

\newcommand\vldbavailabilityurl{URL_TO_YOUR_ARTIFACTS}
\newcommand\vldbpagestyle{empty} 
\newcommand\myproximity{\tikz\draw[fill=orange, draw=none] (0,0) -- (3,0) -- (1.5, 2.598) -- cycle;} 
\newcommand\myclustering{\tikz\draw[fill=orange, draw=none] (0,0) arc[start angle=0, end angle=180, radius=2];}
\newcommand\mydiscord{\tikz\draw[fill=orange, draw=none] (0,-1) rectangle (2,3) (-1,0) rectangle (3,2);}
\newcommand\mydistribution{\tikz\draw[fill=magenta, draw=none] (0,0) circle (1.5);}
\newcommand\mygraph{\tikz\draw[fill=magenta, draw=none] (-0.1,0) -- (2.1,0) -- (3,2) -- (-1,2) -- cycle;} 
\newcommand\mytree{\tikz\draw[fill=magenta, draw=none] (0,0) -- (3,3) -- (6,0) -- (3,-3) -- cycle;}
\newcommand\myencoding{\tikz\draw[fill=magenta, draw=none] (0,0) -- (3,0) -- (1.5, -2.598) -- cycle;}
\newcommand\myforecasting{{\tikz\draw[fill=cyan, draw=none] (0:2) 
    \foreach \i in {72,144,216,288} 
    {-- (\i:2)} -- cycle;}}
\newcommand\myreconstruction{\tikz\draw[fill=cyan, draw=none] (0,0) -- (3,0) -- (3,3) -- (0,3) -- cycle;} 
\newcommand\mycontrast{\tikz\draw[fill={rgb,255:red,0;green,128;blue,0}, draw=none] (0:2) 
    \foreach \i in {60,120,180,240,300} 
    {-- (\i:2)} -- cycle;}

\usepackage{makecell}
\usepackage{array}
\usepackage{algorithmicx,algorithm}
\usepackage{multirow}
\usepackage[normalem]{ulem}
\usepackage{subcaption}
\usepackage{graphicx}
\usepackage{tabularx}
\usepackage{diagbox}
\usepackage{balance}
\usepackage[flushleft]{threeparttable}
\usepackage{booktabs}
\usepackage{pifont}

\usepackage{amssymb}
\usepackage[misc]{ifsym} 
\usepackage{fontawesome}
\usepackage{color}
\usepackage[utf8]{inputenc}
\usepackage[T1]{fontenc}
\usepackage{CJKutf8}
\usepackage{enumitem}
\usepackage{tikz}
\usepackage{amsmath}

\definecolor{mydarkgreen}{RGB}{0,100,0} 
\newcommand{\cmark}{\textcolor{mydarkgreen}{\ding{51}}} 
\newcommand{\xmark}{\textcolor{red}{\ding{55}}} 

\usepackage{ulem}
\usepackage{xcolor}
\usepackage{color,xcolor}

\begin{document}
\title{TAB: Unified Benchmarking of Time Series Anomaly\\ Detection Methods}
 

\settopmatter{authorsperrow=4} 
 
\author{Xiangfei Qiu}
\affiliation{%
  \institution{East China Normal University, China}
}

\author{Zhe Li}
\affiliation{%
  \institution{East China Normal University, China}
}

\author{Wanghui Qiu}
\affiliation{%
  \institution{East China Normal University, China}
}

\author{Shiyan Hu}
\affiliation{%
  \institution{East China Normal University, China}
}

\author{Lekui Zhou}
\affiliation{%
  \institution{Huawei Cloud Availability Engineering Lab, China}
}

\author{Xingjian Wu}
\affiliation{%
  \institution{East China Normal University, China}
}

\author{Zhengyu Li}
\affiliation{%
  \institution{East China Normal University, China}
}

\author{Chenjuan Guo}
\affiliation{%
  \institution{East China Normal University, China}
}

\author{Aoying Zhou}
\affiliation{%
  \institution{East China Normal University, China}
}

\author{Zhenli Sheng}
\affiliation{%
  \institution{Huawei Cloud Availability Engineering Lab, China}
}


\author{Jilin Hu}
\affiliation{%
  \institution{East China Normal University, China \Letter}
}


\author{Christian S. Jensen}
\affiliation{%
  \institution{Aalborg University, Denmark}
}

\author{Bin Yang}
\affiliation{%
  \institution{East China Normal University, China}
}

\begin{abstract}
Time series anomaly detection (TSAD) plays an important role in many domains such as finance, transportation, and healthcare. With the ongoing instrumentation of reality, more time series data will be available, leading also to growing demands for TSAD. While many TSAD methods already exist, new and better methods are still desirable. However, effective progress hinges on the availability of reliable means of evaluating new methods and comparing them with existing methods. We address deficiencies in current evaluation procedures related to datasets and experimental settings and protocols. Specifically, we propose a new time series anomaly detection benchmark, called TAB. First, TAB encompasses 29 public multivariate datasets and 1,635 univariate time series from different domains to facilitate more comprehensive evaluations on diverse datasets. Second, TAB covers a variety of TSAD methods, including Non-learning, Machine learning, Deep learning, LLM-based, and Time-series pre-trained methods. Third, TAB features a unified and automated evaluation pipeline that enables fair and easy evaluation of TSAD methods. Finally, we employ TAB to evaluate existing TSAD methods and report on the outcomes, thereby offering a deeper insight into the performance of these methods.
\end{abstract}

\maketitle

\pagestyle{\vldbpagestyle}
\begingroup\small\noindent\raggedright\textbf{PVLDB Reference Format:}\\Xiangfei Qiu, Zhe Li, Wanghui Qiu, Shiyan Hu, Lekui Zhou, Xingjian Wu, Zhengyu Li, Chenjuan Guo, Aoying Zhou,  Zhenli Sheng, Jilin Hu, Christian S. Jensen and Bin Yang. TAB: Unified Benchmarking of Time Series Anomaly Detection Methods. PVLDB, \vldbvolume(\vldbissue): \vldbpages, \vldbyear.\\
\href{https://doi.org/\vldbdoi}{doi:\vldbdoi}
\endgroup
\begingroup
\renewcommand\thefootnote{}\footnote{\noindent
This work is licensed under the Creative Commons BY-NC-ND 4.0 International License. Visit \url{https://creativecommons.org/licenses/by-nc-nd/4.0/} to view a copy of this license. For any use beyond those covered by this license, obtain permission by emailing \href{mailto:info@vldb.org}{info@vldb.org}. Copyright is held by the owner/author(s). Publication rights licensed to the VLDB Endowment. \\
\raggedright Proceedings of the VLDB Endowment, Vol. \vldbvolume, No. \vldbissue\ %
ISSN 2150-8097. \\
\href{https://doi.org/\vldbdoi}{doi:\vldbdoi} \\
}\addtocounter{footnote}{-1}\endgroup

\ifdefempty{\vldbavailabilityurl}{}{
\vspace{.3cm}
\begingroup\small\noindent\raggedright\textbf{PVLDB Artifact Availability:}\\
The source code, data, and/or other artifacts have been made available at \url{https://github.com/decisionintelligence/TAB}.
\endgroup
}

\section{Introduction}
\label{sec:Introduction}
\begin{figure}[t]
    \centering
    \includegraphics[width=1\linewidth]{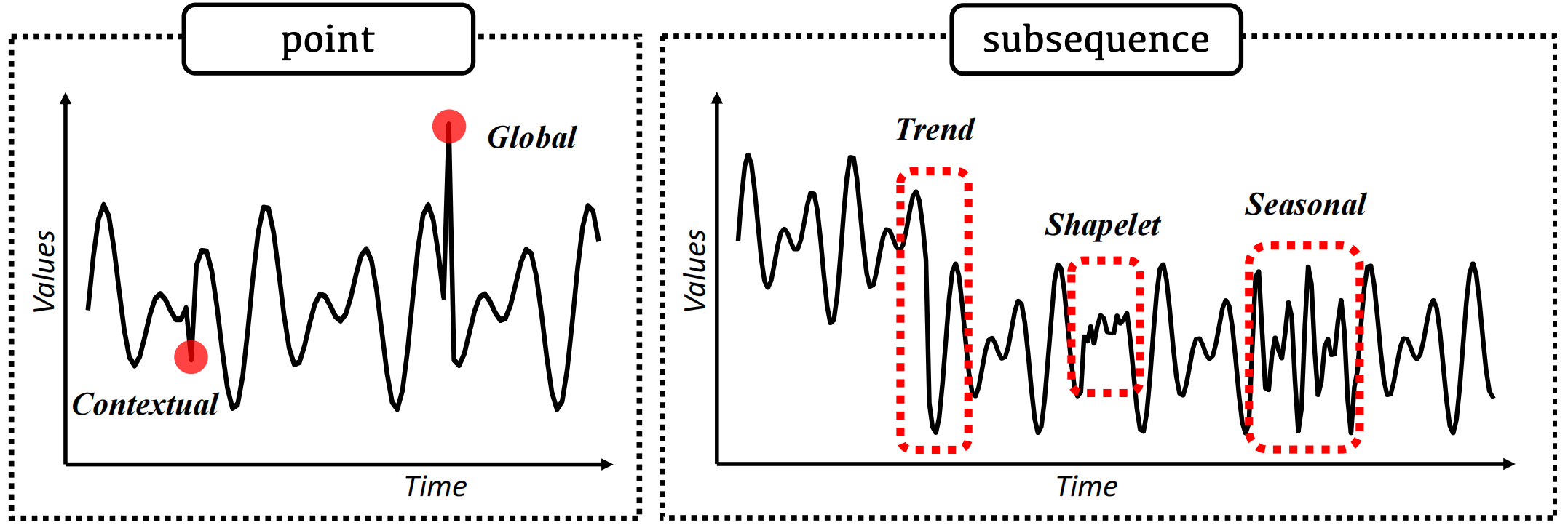}
    \caption{Types of time series anomalies. }
    \label{fig:pipeline}
\label{fig:anomaly-type}
\end{figure}

Due to the ongoing digitalization and instrumentation of processes with sensors, time series data is collected in numerous settings and fuels an increasing number of applications~\cite{wu2025k2vae,AutoCTS++,liu2025timebridge,liu2025rethinking,yu2024ginar,qiu2025easytime}. In recent years, there has been significant progress in time series analysis, with key tasks such as forecasting~\citep{wu2024multi,liu2023wftnet,qiu2025comprehensive,yu2025merlin,DSformer,HDMixer,yu2025ginarp}, classification~\citep{DBLP:conf/icde/YaoJC0GW24,DBLP:journals/pacmmod/0002Z0KGJ23,AimTS}, and imputation~\citep{wang2024lspt,gao2025ssdts,wang2024spot,escmtifs,wangoptimal}, among others~\citep{wang2025fredf,DBLP:conf/nips/HuangSZDWZW23,wang2024entire,wang2025tois}, gaining attention. Among these, Time Series Anomaly Detection (TSAD), which aims to identify data in time series that derive from different processes than the intended processes, stands out as a critical studied task. The objective is thus to invent a \textit{detector} that can detect the data that deviates from the expected or normal behavior~\cite{chandola2009anomaly,hu2024multirc,wang2023drift}. For example, financial firms aim to identify unusual transactions in time to prevent financial fraud, and manufacturing companies are eager to predict the impending malfunction of devices so that they can be replaced before they fail. Time series anomalies can be classified into \textbf{point} and \textbf{subsequence} anomalies~\cite{lai2021revisiting,zamanzadeh2024deep}. As is shown in Figure~\ref{fig:anomaly-type}, \textbf{point} anomalies can be further classified as \textit{contextual} or \textit{global} anomalies, while \textbf{subsequence} can be classified as \textit{trend}, \textit{shapelet}, or \textit{seasonal} anomalies. 

Recently, a variety of TSAD methods have been proposed. These methods include non-learning methods~\cite{breunig2000lof, goldstein2012histogram, yeh2016matrix}, machine learning methods~\cite{ramaswamy2000efficient, liu2008isolation, scholkopf1999support}, deep learning methods~\cite{yang2023dcdetector, xu2021anomaly, wang2024drift}, LLM-based methods~\cite{gpt4ts, unitime}, and Time-series pre-trained (TS pre-trained) methods~\cite{timer, units, moment}. 
According to the degree of supervision, TSAD methods can also be divided into unsupervised, semi-supervised, and supervised learning methods~\cite{chandola2009anomaly}. More specifically, unsupervised learning methods do not require labeled anomalies for training, 
semi-supervised learning methods are trained on time series without anomalous data points,
and supervised learning methods are trained on time series with all data points labeled as normal or abnormal. 
In reality, anomalies are rare, resulting in labeled data being rare, so supervised learning methods are seldom used in TSAD~\cite{schmidl2022anomaly}. 
Therefore, we focus on the evaluation of unsupervised and semi-supervised methods. 

Time series include different types of anomalies, and it is difficult to invent a method that is good at detecting all types of anomalies~\cite{schmidl2022anomaly}. We have surveyed a total of 100 studies\footnote{For details on the reviewed studies, please refer to: \url{https://github.com/decisionintelligence/TAB/blob/main/docs/paper_statistics.csv}.} that are published in recent years, yet we found that the existing works fall
short as follows.

\begin{figure}[t]
  \centering
  \subfloat[Number of datasets]
  {\includegraphics[width=0.24\textwidth]{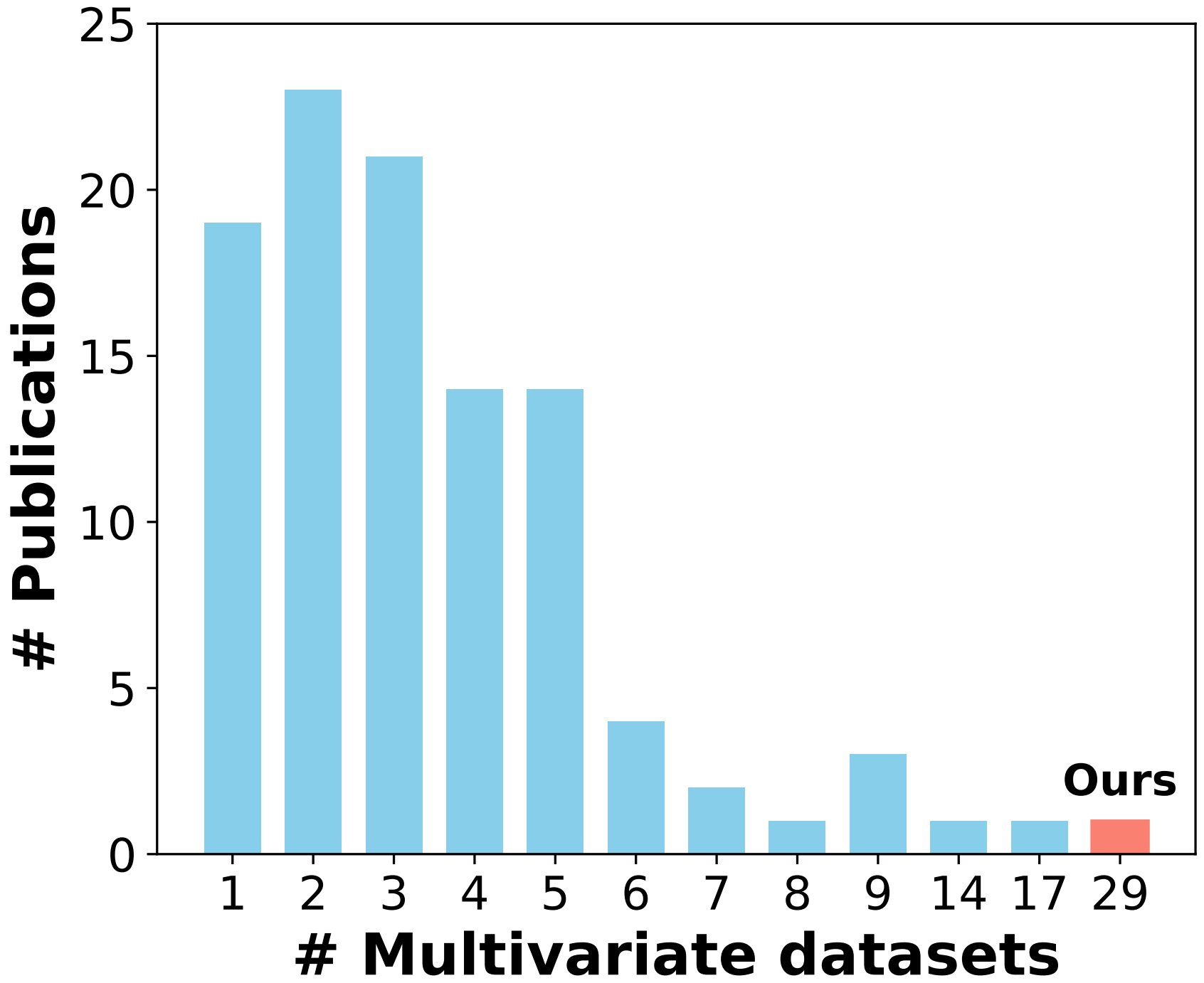}
  \label{fig:multi datasets}}
  \subfloat[Number of datasets domains]
  {\includegraphics[width=0.24\textwidth]{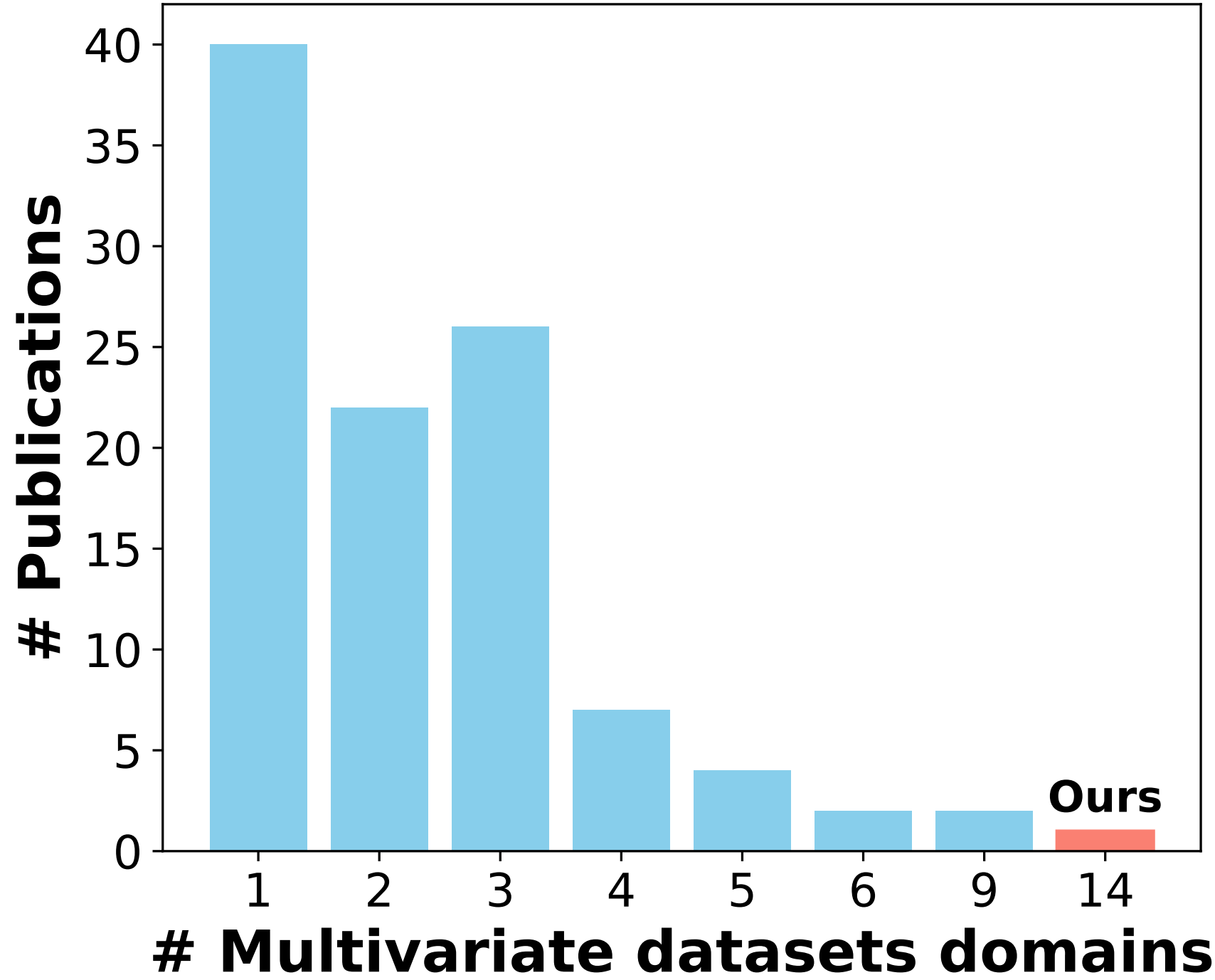}
  \label{fig:dataset domain}}
  \caption{limits in multivariate datasets.}
\label{Threshold problem description.}
\end{figure}

First, \textit{few of the most recent studies include a performance study that covers a broad variety of datasets.} Figure~\ref{fig:multi datasets} shows the numbers of multivariate datasets included in the existing TSAD studies. We observe that more than half of the studies include at most four datasets, and only one study covers 17 datasets. When considering the domains of datasets, we find that the number of domains is also limited---see Figure~\ref{fig:dataset domain}. 
The limited numbers of datasets and domains serve to question whether the proposed methods were evaluated comprehensively~\cite{dau2019ucr}. 

\begin{figure}[t]
  \centering
  \subfloat[SMAP]
  {\includegraphics[width=0.23\textwidth]{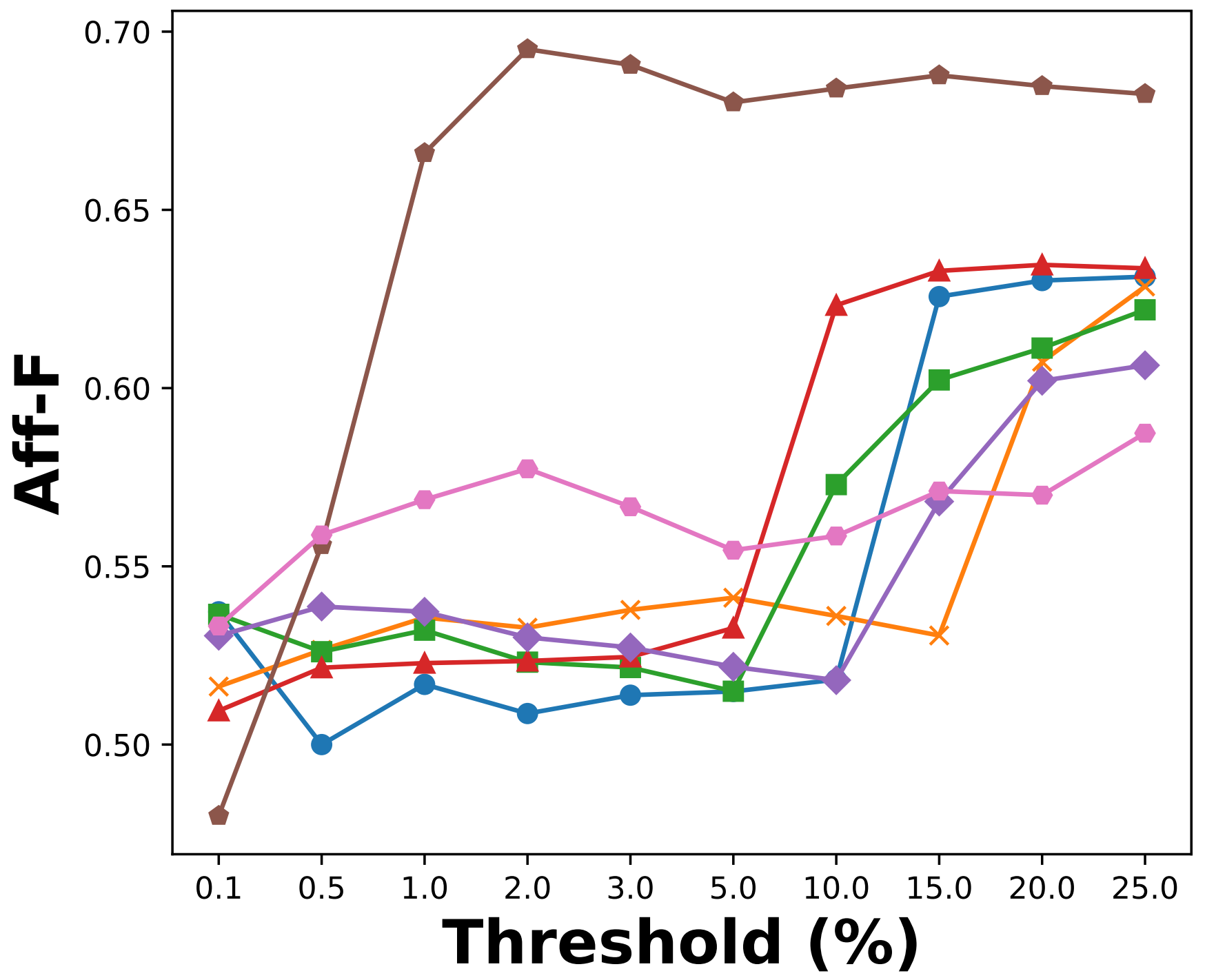}\label{fig:fixed}}
  \hspace{0.05mm} 
  \subfloat[PSM]
  {\includegraphics[width=0.233\textwidth]{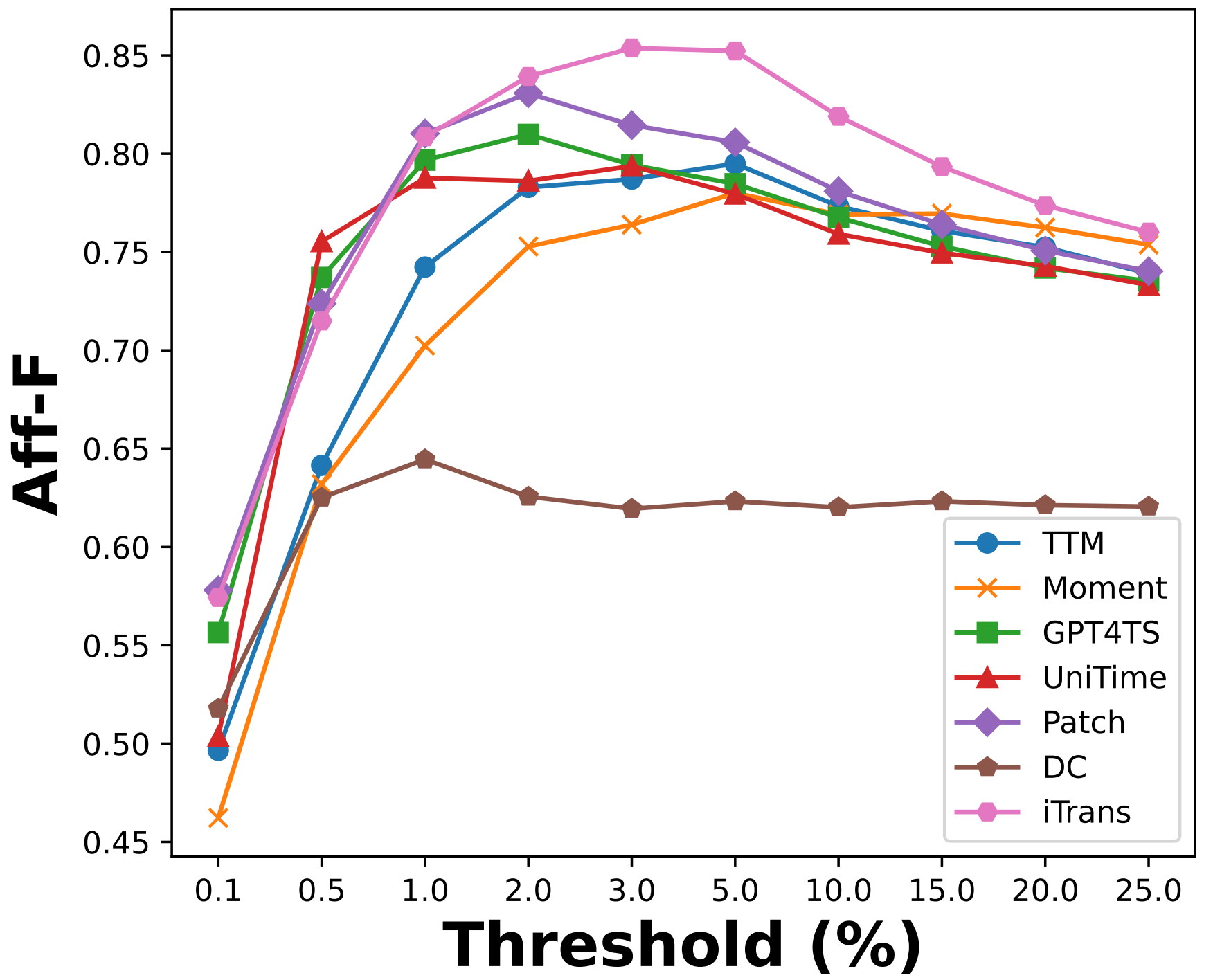}\label{fig:rolling}}
  \caption{Threshold problem description.}
\label{Threshold problem description.}
\end{figure}

Second, \textit{few of the most recent studies adopt consistent experimental settings and evaluation protocols, thus failing to offer reliable results to real-world applications}. 
Specifically, inconsistencies are observed in relation to the following aspects: (i)~\textit{Inconsistent Dataset Splitting. } Some methods use a validation set from the testing set~\cite{yang2023dcdetector,xu2021anomaly}, while others use a validation set from the training set~\cite{wang2023drift, wu2022timesnet}, resulting in inconsistent dataset splitting approaches.
(ii)~\textit{Drop-last Issue.} To accelerate the testing phase, it is common to split the data into batches. Unless all methods use the same batch size, discarding the last incomplete batch with fewer sample instances than the batch size is unfair, as the actual usage length of the testing set is inconsistent. ~\cite{qiu2024tfb,shi2024scaling}. (iii)~\textit{Point Adjustment.} This is a technique for modifying predictions to complement the point-wise F1-scores. Any anomaly window with at least one correctly predicted time step is considered predicted correctly~\cite{xu2018unsupervised}. However, even random methods have a good chance to predict at least one point in larger anomaly windows, the effect being that they can easily reach the performance of complex methods or even outperform them~\cite{kim2022towards}. Additionally, threshold-independent evaluation metrics such as $AUC$-$ROC$ should not be recalculated after point adjustment. (iv)~\textit{Threshold Problem.} When calculating some label-based metrics~(e.g., $precision$, $recall$), methods that output anomaly scores need to select a threshold to convert anomaly scores into anomaly labels. Then, as is shown in Figure~\ref{Threshold problem description.}, when the threshold changes, the performance can change considerably. Therefore, different threshold choices during the evaluation of the same dataset can impact performance comparisons. (v)~\textit{Inconsistent Algorithm Outputs.} For instance, TimesNet~\cite{wu2022timesnet} chooses overlapping windows during testing, while DCdetector~\cite{yang2023dcdetector}, Anomaly Transformer~\cite{xu2021anomaly}, and others choose non-overlapping windows. This difference in output lengths for different methods on the same dataset affects the direct comparison of method performance.

Third, \textit{recent studies have rarely systematically evaluated the performance of foundation methods, particularly time series pre-trained methods in TSAD.}
As mentioned above, the types of anomalies in time series are diverse, encompassing point anomalies~(global and contextual), subsequence anomalies~(seasonal, trend, and shapelet), as well as mixed types. Recall the ``No-Free-Lunch Theorem," stating that improving a method for one aspect leads to deterioration of another aspect~\cite{wolpert1997no}. This implies that no single TSAD method is universally best for all time series and anomaly types. Therefore, it is challenging to select a suitable method for a given time series. Foundation methods which include LLM-based and time series pre-trained methods are widely recognized for their strong generalization capabilities and may help address the limitations of specific methods\footnote{Most existing TSAD methods require training on specific datasets before they can perform inference on corresponding datasets. In this paper, these methods are referred to as ``specific methods," to distinguish them from ``foundation methods."}, which often struggle with generalization and fail to handle all types of anomaly data effectively. 
However, recent studies have seldom conducted systematic evaluations of the performance of these foundational methods, nor have they extensively compared their practical effectiveness with more specific methods.

\begin{figure}[t]
    \centering
    \includegraphics[width=0.8\linewidth]{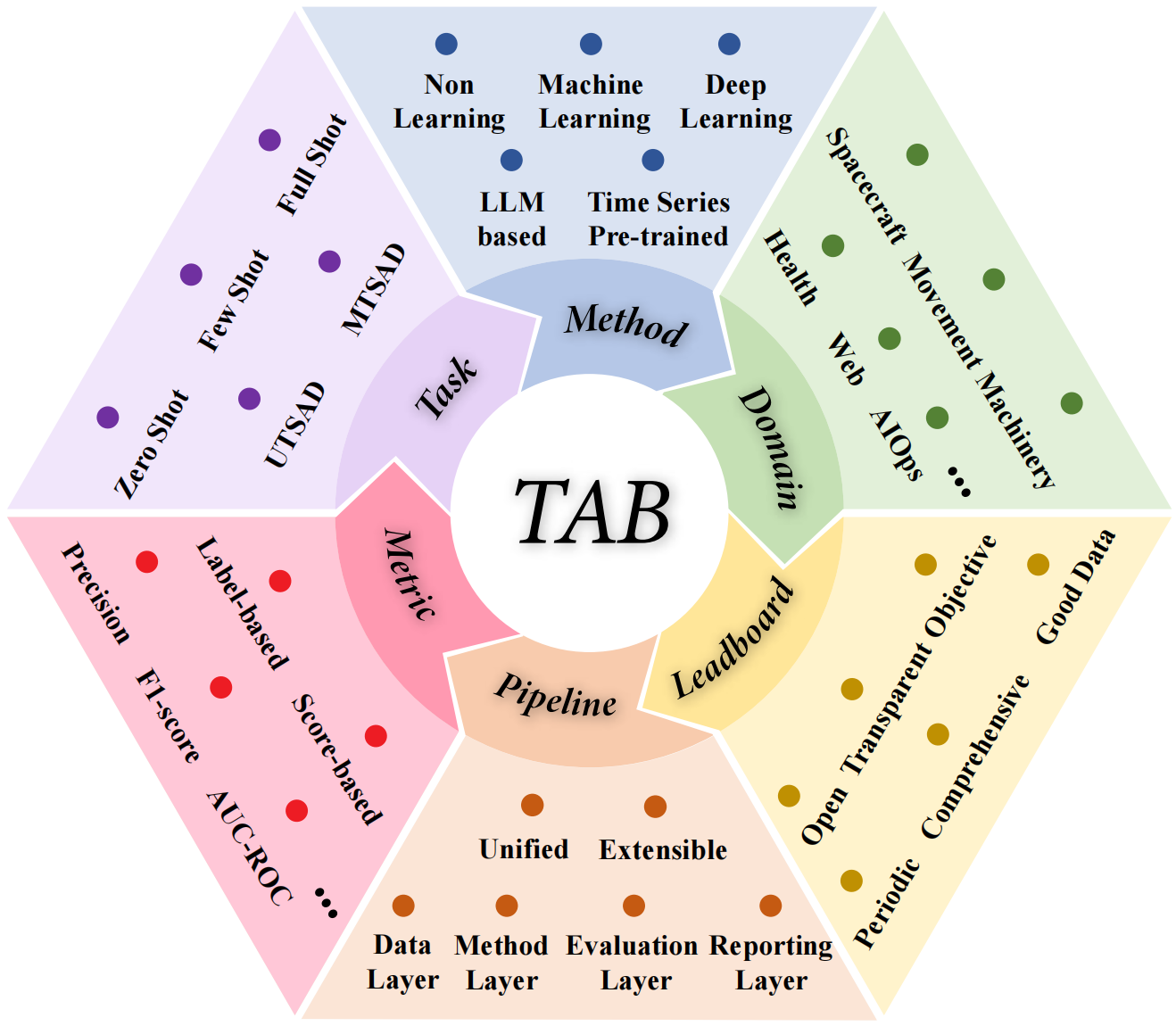}
    \caption{Key characteristics of TAB. }
    \label{fig:pipeline}
\label{Key characteristics of TAB.}
\end{figure}

To address the three challenges covered above, we propose a \textbf{T}ime series \textbf{A}nomaly detection \textbf{B}enchmark~(TAB). As shown in Figure~\ref{Key characteristics of TAB.}, TAB has the following key characteristics:

(1) Comprehensive datasets (to address \textbf{Challenge 1}): TAB collects and filters 29 public multivariate datasets and 1,635 univariate time series, spanning diverse domains.

(2) Unified and extensible pipeline (to address \textbf{Challenge 2}): To enable fair and efficient comparison of both univariate and multivariate TSAD methods, TAB establishes a unified, user-friendly, and extensible evaluation pipeline, enabling different methods to undergo consistent performance evaluation. 

(3) Broad coverage of existing methods (to address \textbf{Challenge 3}): TAB covers state-of-the-art TSAD methods, including Non-learning, Machine learning, Deep learning, LLM-based, and Time-series pre-trained methods.

(4) Multi evaluation strategies and tasks: TAB is specifically designed to further improve fair comparisons in TSAD, include univariate time series anomaly detection (UTSAD) and multivariate time series anomaly detection (MTSAD). Besides, it
 integrates zero-shot, few-shot, and full-shot evaluation strategies, facilitating improved assessment of method performance.

(5) Diversified evaluation metrics: TAB covers Label-based and Score-based metrics to ensure a comprehensive assessment of model performance.

(6) Continuously updated leaderboard: TAB has also launched an online TSAD leaderboard that covers LLM-based and time series pre-training methods. The leaderboard is open and transparent, and we welcome everyone to participate actively.



\textcolor{black}{Based on the experiments conducted using TAB, we make the following key observations:~(1) Non learning and machine learning methods exhibit strong performance on the UTSAD task, with KMeans, DWT, S2G, and, OCSVM performing the best. (2) On the MTSAD task, the deep learning and the machine learning methods demonstrate clear advantages, with methods such as TsNet, CATCH, DAGMM, KNN, and KMeans exhibiting excellent performance. (3) Global point anomalies are the easiest to detect, while mixed anomalies are the most challenging. (4) Deep learning as well as foundation methods are more suitable than other methods for finding point anomalies, while non learning and machine learning methods are better suited for finding subsequence anomalies. (5) Density based methods are appropriate for identifying trend anomalies, and distance based methods are suitable for identifying shapelet anomalies. (6) The application of foundation methods in TSAD needs further development, as their current performance remains relatively poor.}

In summary, we make the following main contributions:
\begin{itemize}[left=0.12cm]
\item We present TAB, a unified and extensible TSAD evaluation benchmark. TAB streamlines the process of evaluating TSAD methods and enables fair comparison of methods based on a unified experimental setup. Besides, all datasets and code are available at \url{https://github.com/decisionintelligence/TAB}. We have also launched \textcolor{black}{an online} leaderboard \url{https://decisionintelligence.github.io/OpenTS/Benchmarks/overview/} that covers time series pre-trained and LLM-based methods.

\item We survey recent TSAD studies, among which we select 29 multivariate datasets and 1,635 univariate time series. These datasets span various domains, and we classify them based on their characteristics and anomaly types. Besides, they are stored in a unified format for consistency and ease of use.
\item We cover state-of-the-art time series anomaly detection methods, including non-learning, machine learning, deep learning, LLM-based, and time-series pre-trained methods.
\item We use TAB to assess the performance of 40 multivariate TSAD methods on 29 multivariate datasets and 46 univariate TSAD methods on 1,635 univariate time series, covering diverse evaluation strategies and metrics. This provides deeper insights into the performance of these methods.
\end{itemize}

The rest of the paper is structured as follows. We first review related work in Section~\ref{RELATED WORK}. In Section ~\ref{HETEROGENEITY AMONG Time Series AD DATASETS}, we classify datasets, taking into consideration both the inherent characteristics of the datasets and the different types of anomalies. In Section~\ref{TAB: BENCHMARK DETAILS}, we cover the design of TAB, and we evaluate existing TSAD methods using TAB in Section~\ref{EXPERIMENTS}. Finally, we conclude in Section~\ref{sec:Conclusions}.

\begin{table*}[t!]
\caption{Time series anomaly detection benchmark comparison.}
\label{Time series anomaly detection benchmark comparison.}
\resizebox{1.9\columnwidth}{!}{
\begin{tabular}{ll|cccccccccccccc}
\toprule
\multicolumn{2}{c|}{\multirow{2}{*}{\diagbox{Property}{Benchmark}}} & NAB & Exathlon & TODS & TSB-UAD & TSB-AD & TimeEval & TimeseAD & Tadpak & TimeSeries & ADB & MTAD & OrionBench & \textcolor{black}{Experiment} &\textbf{TAB}  \\ & &~\cite{lavin2015evaluating} & ~\cite{jacob2020exathlon} & ~\cite{lai2021revisiting} & ~\cite{paparrizos2022tsb} & ~\cite{liuelephant} &~\cite{wenig2022timeeval} & ~\cite{wagner2023timesead} & ~\cite{kim2022towards} & -Bench~\cite{si2024timeseriesbench} & ~\cite{zhang2024advancing} & ~\cite{liu2024mtad} & ~\cite{alnegheimish2023making} & \textcolor{black}{-TSAD}~\cite{zhang2023experimental} & \textbf{(Ours)}\\ \midrule
\multicolumn{1}{l|}{\multirow{5}{*}{{\makecell{Evaluation \\ Datasets}}}} & Real-Word &  \cmark &  \cmark &  \cmark &  \cmark &  \cmark & \cmark &  \cmark &  \cmark &  \cmark &  \cmark &  \cmark &  \cmark &  \cmark &  \cmark \\
\multicolumn{1}{l|}{} & Synthetic &  \cmark & \xmark &  \cmark &  \cmark &  \cmark & \cmark & \xmark & \xmark &  \cmark &  \cmark & \xmark &  \cmark &  \cmark &  \cmark \\
\multicolumn{1}{l|}{} & Univariate  &  \cmark & \xmark &  \cmark &  \cmark &  \cmark & \cmark & \xmark & \xmark &  \cmark &  \cmark & \xmark &  \cmark &  \cmark &  \cmark \\
\multicolumn{1}{l|}{} & Multivariate  & \xmark &  \cmark &  \cmark & \xmark & \cmark &  \cmark &  \cmark &  \cmark & \xmark &  \cmark &  \cmark &  \cmark &  \cmark &  \cmark \\
\multicolumn{1}{l|}{} & Dataset taxonomy & \xmark &  \cmark &  \cmark &  \cmark &   \cmark &\cmark & \xmark & \xmark &  \cmark & \xmark & \xmark & \xmark &  \cmark &  \cmark \\\hline
\multicolumn{1}{l|}{\multirow{5}{*}{\makecell{Evaluation\\ Methods}}} & Non-learning &  \cmark & \xmark &  \cmark &  \cmark &  \cmark & \cmark & \xmark &  \cmark &  \cmark &  \cmark &  \cmark &  \cmark &  \cmark&  \cmark \\
\multicolumn{1}{l|}{}  & Machine-learning &  \cmark & \xmark &  \cmark &  \cmark &   \cmark &\cmark & \xmark & \xmark &  \cmark &  \cmark &  \cmark &  \cmark &  \cmark&  \cmark \\
\multicolumn{1}{l|}{}  & Deep-learning & \xmark &  \cmark &  \cmark &  \cmark &   \cmark &\cmark &  \cmark &  \cmark &  \cmark &  \cmark &  \cmark &  \cmark &  \cmark&  \cmark \\
 \multicolumn{1}{l|}{} & LLM-based & \xmark & \xmark & \xmark & \xmark &  \cmark &\xmark & \xmark & \xmark & \xmark & \xmark & \xmark & \xmark & \xmark&  \cmark \\
 \multicolumn{1}{l|}{} & TS pre-trained & \xmark & \xmark & \xmark & \xmark &  \cmark & \xmark & \xmark & \xmark & \xmark & \xmark & \xmark & \xmark& \xmark &  \cmark \\\hline
\multicolumn{1}{l|}{\multirow{3}{*}{\makecell{Evaluation \\Strategies}}} & Zero-shot & \xmark & \xmark & \xmark & \xmark & \xmark & \xmark & \xmark & \xmark &  \cmark & \xmark & \xmark & \xmark & \xmark&  \cmark \\
\multicolumn{1}{l|}{}  & Few-shot & \xmark & \xmark & \xmark & \xmark & \xmark & \xmark & \xmark & \xmark & \xmark & \xmark & \xmark & \xmark& \xmark &  \cmark \\
\multicolumn{1}{l|}{}  & Full-shot &  \cmark &  \cmark &  \cmark &  \cmark &  \cmark & \cmark &  \cmark &  \cmark &  \cmark &  \cmark &  \cmark &  \cmark& \cmark &  \cmark \\\hline
\multicolumn{1}{l|}{\multirow{2}{*}{\makecell{Evaluation\\ Metrics}}}  & Score-based & \xmark &  \cmark & \xmark &  \cmark &  \cmark & \cmark &  \cmark & \xmark &  \cmark &  \cmark & \xmark & \xmark &\xmark&  \cmark \\
\multicolumn{1}{l|}{}  & Label-based &  \cmark &  \cmark &  \cmark &  \cmark &  \cmark & \cmark &  \cmark &  \cmark &  \cmark &  \cmark &  \cmark &  \cmark &  \cmark&  \cmark \\\hline
\multicolumn{2}{c|}{List inconsistent phenomena} & \xmark & \xmark & \xmark &  \cmark  &\xmark & \xmark &  \cmark & \xmark & \xmark & \xmark &  \cmark & \xmark&  \cmark &  \cmark \\\hline
\multicolumn{2}{c|}{Accuracy, memory, and runtime trade-off}  & \xmark &  \cmark & \xmark & $\bigcirc$   & $\bigcirc$ &  \cmark & \xmark & \xmark &  \cmark & \xmark &  \cmark & $\bigcirc$ & $\bigcirc$ &  \cmark \\ \hline
\multicolumn{2}{c|}{Method recommendations} & \xmark & \xmark & \xmark & \xmark & \xmark & \cmark & \xmark & \xmark & \xmark & \xmark & \xmark & \xmark &  \cmark &  \cmark \\\hline
\multicolumn{2}{c|}{Unified and extensible pipeline} &  \cmark & $\bigcirc$ &  $\bigcirc$ & \cmark & \cmark & \cmark &  $\bigcirc$ & \xmark &  \cmark &  \cmark & $\bigcirc$ &  \cmark &  $\bigcirc$ &  \cmark \\\hline
\multicolumn{2}{c|}{Continuously updated leaderboard} & \xmark & \xmark & \xmark & \xmark & \cmark & \xmark & \xmark & \xmark &  \cmark & \xmark & \xmark &  \cmark &  \xmark &  \cmark \\
\bottomrule
\multicolumn{16}{l}{\xmark~indicates absent, \cmark~indicates present,$\bigcirc$ indicates incomplete.}
\end{tabular}
}
\end{table*}

\section{Related Works}
\label{RELATED WORK}
\subsection{Time Series Anomaly Detection}

We distinguish among five categories of time series anomaly detection methods: non-learning~(NL), machine learning~(ML), deep learning~(DL), LLM-based, and Time-series pre-trained~(TS pre-trained) methods. 

Among the non-learning methods, classical methods such as local outlier factor~(LOF) identify outliers in multidimensional datasets by using the notion of local density~\cite{breunig2000lof}. 
When the distances between data points are small or the data points are densely packed in dense datasets, LOF fails to perform well. HBOS~\cite{goldstein2012histogram} calculates anomaly scores based on histograms of feature values, which makes it easy to identify global anomalies but it is not good at solving contextual anomalies. Matrix Profile~\cite{yeh2016matrix} aims to find anomalies by measuring the distance between subsequences of the same length. If this distance of a subsequence to the most similar subsequences is large, the subsequence is likely an anomaly. 
In summary, non-learning methods are designed to solve a specific anomaly, falling short in different kinds of anomalies. 

Representative works in machine learning methods include Isolation Forest (IF)~\cite{liu2008isolation},  K-Nearest Neighbors~(KNN)~\cite{ramaswamy2000efficient}, One-Class SVM (OCSVM)~\cite{scholkopf1999support}, among others. 
IF assumes that anomalies have low density and can be easily separated based on tree models with a single split, while normal values have high density and require multiple splits. So, IF can process high-dimensional data efficiently but it struggles with datasets where anomalies are densely packed or with varying densities. 
KNN defines anomalies as the top-\textit{k} data points with the largest sum of distances to their \textit{k} nearest neighbors. 
OCSVM maps the data into a feature space according to a kernel and separates it from the origin with maximum margin. 

Recently, studies have shown that learned features may perform
better than human-designed features~\cite{wu2025affirm,he2025enhancing1,zhang2024distilling,yu2025prnet,yu2024scnet}. Leveraging the representation
learning of deep neural networks (DNNs), many deep learning-based methods emerge. These include TimesNet~\cite{wu2022timesnet}, Anomaly Transformer~\cite{xu2021anomaly}, DCdetector~\cite{yang2023dcdetector}, among others. Anomaly Transformer proposes a minimax strategy to amplify the normal-abnormal distinguishability of the Association Discrepancy and use a new Anomaly-Attention mechanism to compute the association discrepancy to identify anomalies~\cite{xu2021anomaly}. DCdetector introduces contrastive learning into anomaly detection tasks~\cite{yang2023dcdetector}. Its objective is to create an embedding space where similar data samples~(normal values) are close to each other, while dissimilar data samples~(abnormal values) are far away from each other. Different from non-learning and machine learning methods, deep learning methods do not focus on a specific anomaly only, but leverage the learned latent space to distinguish different anomalies. 

LLM-based methods leverage the strong representational capacity and sequential modeling capability of LLMs to capture complex patterns for time series modeling. GPT4TS \cite{gpt4ts} selectively adjust specific parameters such as positional encoding and layer normalization of LLMs, enabling the model to quickly adapt to time series while retaining most of pre-trained knowledge. Besides, methods such as UniTime \cite{unitime} focuses on designing prompts, such as learnable prompts, prompt pools, and domain-specific instructions to activate time series knowledge in LLMs. 

Recently, pre-training on multi-domain time series data has garnered significant attention. Reconstruction-based methods, such as UniTS \cite{units} and Moment \cite{moment}, focus on restoring the features of time series data, enabling the extraction of valuable information in an unsupervised manner. These methods primarily employ an encoder architecture. In contrast, autoregressive-based methods like Timer \cite{timer} utilize next-token prediction to learn time series representations, typically adopting a decoder architecture. Direct prediction methods, such as TTM \cite{ekambaram2024ttms}, unify the training process for pre-training and downstream tasks, allowing models to demonstrate strong adaptability when transitioning to forecasting tasks.

\textcolor{black}{Even though tremendous TSAD methods have been proposed, consistently evaluating and comparing them remains a challenge.}

\subsection{Time Series Anomaly Detection Benchmarking} In a field where new methods are proposed frequently, accurate assessment of method performance is crucial. A comprehensive benchmark is important for advancing the state-of-the-art by offering researchers the opportunity to evaluate their methods rigorously across diverse datasets~\cite{deng2009imagenet,qiu2024tfb,li2025TSFM-Bench,shao2023exploring}. In the field of TSAD, several benchmarks have been published. However, these benchmarks fall short in different aspects—see Table~\ref{Time series anomaly detection benchmark comparison.}.

First, some benchmarks consider either univariate or multivariate TSAD. NAB~\cite{lavin2015evaluating}, TSB-UAD~\cite{paparrizos2022tsb}, and TimeSeriesBench~\cite{si2024timeseriesbench} have only done excellent work in univariate TSAD. And other works, e.g., Exathlon~\cite{jacob2020exathlon}, TimeseAD~\cite{wagner2023timesead}, Tadpak~\cite{kim2022towards}, and MTAD~\cite{liu2024mtad} just consider multivariate. In addition, most benchmarks fail to evaluate both real-world and synthetic datasets, and they do not perform fine-grained classification on the datasets (for example, by dataset characteristics or anomaly types in the datasets).

Second, a notable issue arises regarding the assessment of diversity of evaluation methods and strategies. Except for NAB~\cite{lavin2015evaluating}, Exathlon~\cite{jacob2020exathlon}, TimeseAD~\cite{wagner2023timesead}, and Tadpak~\cite{kim2022towards}, other benchmarks include NL, ML, and DL methods, which make the evaluation results more comprehensive. However, apart from TSB-AD~\cite{liuelephant}, none of these benchmarks systematically assess the recently popular foundation methods (LLM-based and time series pre-trained methods). A fair and systematic comparison of the performance among these foundation methods, as well as their performance compared to specific methods, is crucial for advancing the future direction of TSAD. Additionally, existing benchmarks have not sufficiently focused on the generalization ability (i.e., inference capability on new or unseen data) of methods. Except for TimeSeriesBench~\cite{si2024timeseriesbench}, which evaluates the method's zero-shot capability, all other benchmarks focus solely on the method's full-shot ability, neglecting comparisons and evaluations of generalization abilities.



Third, some benchmarks lack extensible pipelines. Tadpak~\cite{kim2022towards} does not provide a pipeline in their benchmark at all, making it hard to integrate new methods. Exathlon~\cite{jacob2020exathlon}, TODS~\cite{lai2021revisiting}, TSB-UAD~\cite{paparrizos2022tsb}, TimeseAD~\cite{wagner2023timesead}, and MTAD~\cite{liu2024mtad} lack flexible interfaces for datasets, methods, and evaluation metrics, supporting only limited functionality. Consequently, these benchmarks lack sufficient extensibility, making it difficult to: i) support evaluations on new (proprietary) datasets, ii) effectively assess the performance of innovative methods using existing metrics, and iii) adapt to new evaluation metrics.

Unlike existing benchmarks, TAB provides a comprehensive evaluation of univariate and multivariate datasets, covering state-of-the-art TSAD methods, including NL, ML, DL, LLM-based, and TS pre-training methods.~\textcolor{black}{TAB evaluates the performance of 11 foundation methods on multivariate and univariate TSAD, and it covers three evaluation strategies: zero-shot, few-shot, and full-shot. TAB is the most comprehensive benchmark in terms of the number of foundation methods evaluated and the types of strategies covered.}
TAB offers comprehensive evaluation strategies and metrics, enabling a more accurate performance assessment over various methods and improvement fostering. TAB systematically organizes and unifies numerous inconsistencies found in current evaluations, analyzing the relationship among method accuracy performance, memory usage, and runtime. TAB also categorizes datasets to guide the selection of appropriate anomaly detection methods. Our goal is to deliver a fair and extensible pipeline for comprehensive assessments of diverse datasets and method performances. Additionally, TAB dynamically maintains a leaderboard, contributing consistently to the open-source community.



\section{Heterogeneity Among TSAD Datasets}
\label{HETEROGENEITY AMONG Time Series AD DATASETS}
This section classifies datasets by taking into consideration both the inherent characteristics of the datasets themselves~(in Section~\ref{Time Series Characteristics}) and the types of anomalies they contain~(in Section~\ref{Types of Time Series Anomalies}).

\subsection{Time Series Characteristics}
\label{Time Series Characteristics}
\begin{figure}[t]
\centering
\begin{subfigure}{0.3\linewidth}
    \includegraphics[width=1.0\textwidth]{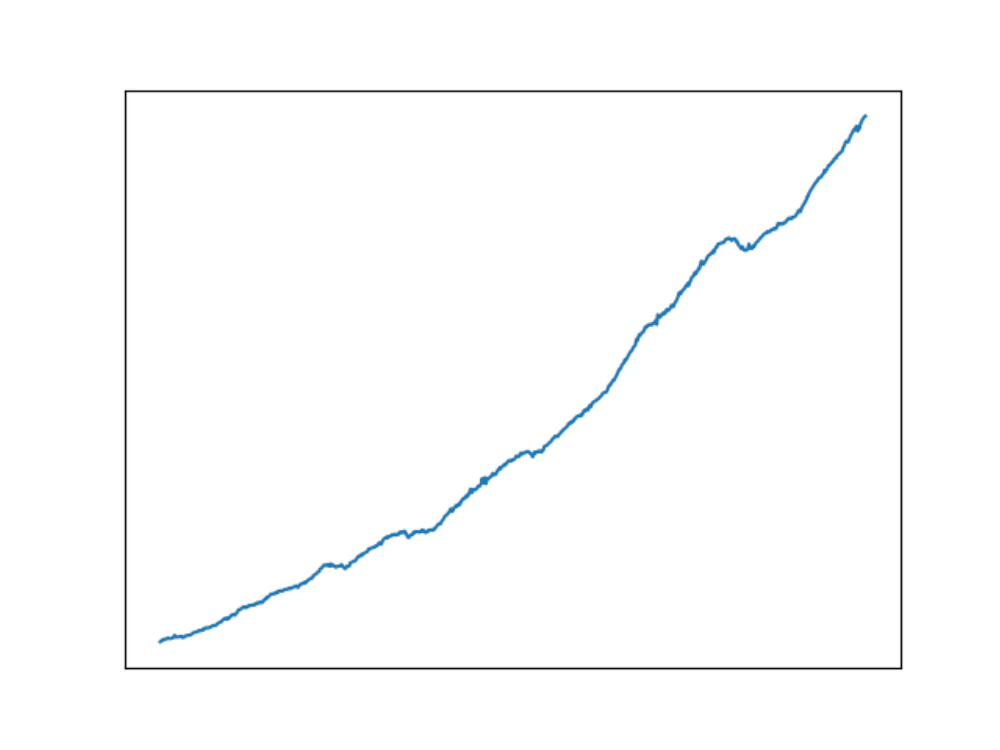}
    \caption{Trend}
    \label{fig:pattern_seansonal}
\end{subfigure}
\begin{subfigure}{0.3\linewidth}
    \includegraphics[width=1.0\textwidth]{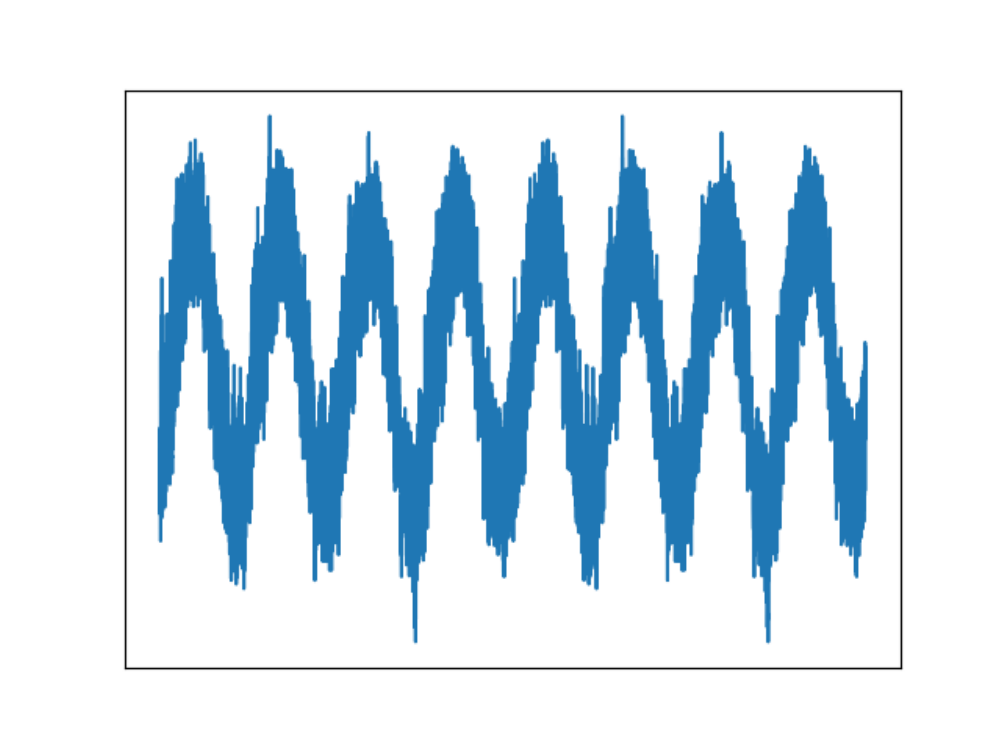}
    \caption{Seasonality}
    \label{fig:pattern_trend}
\end{subfigure}
\begin{subfigure}{0.3\linewidth}
    \includegraphics[width=1.0\textwidth]{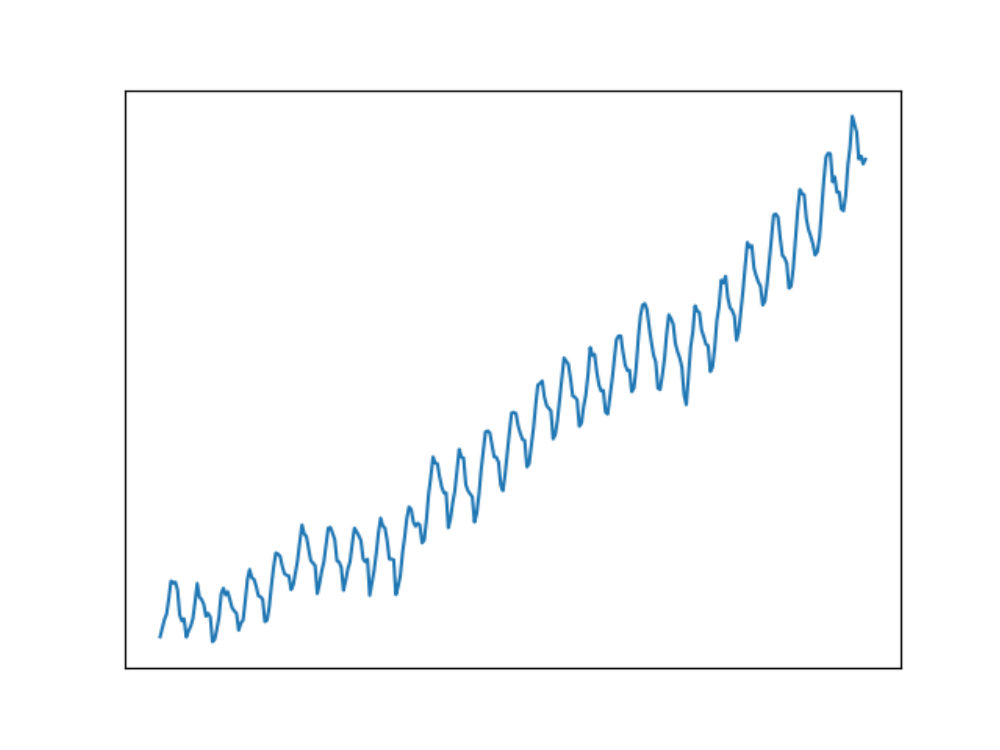}
    \caption{Transition}
    \label{fig:pattern_pattern}
\end{subfigure}
\begin{subfigure}{0.3\linewidth}
    \includegraphics[width=1.0\textwidth]{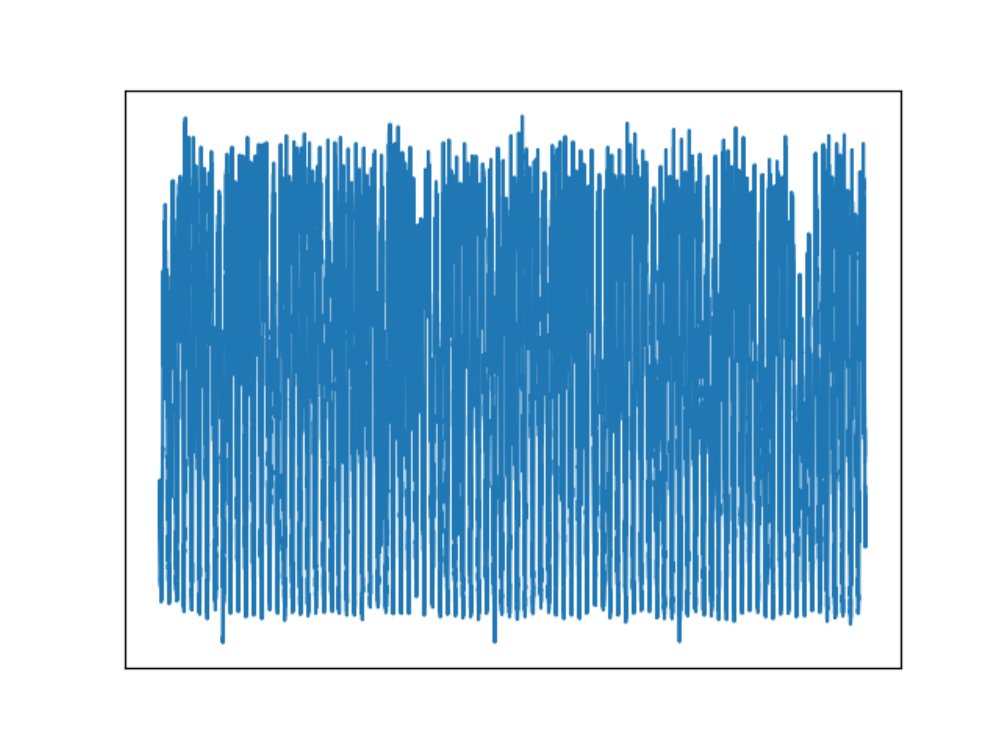}
    \caption{Stationarity}
    \label{fig:pattern_shift}
\end{subfigure}
\begin{subfigure}{0.3\linewidth}
    \includegraphics[width=1.0\textwidth]{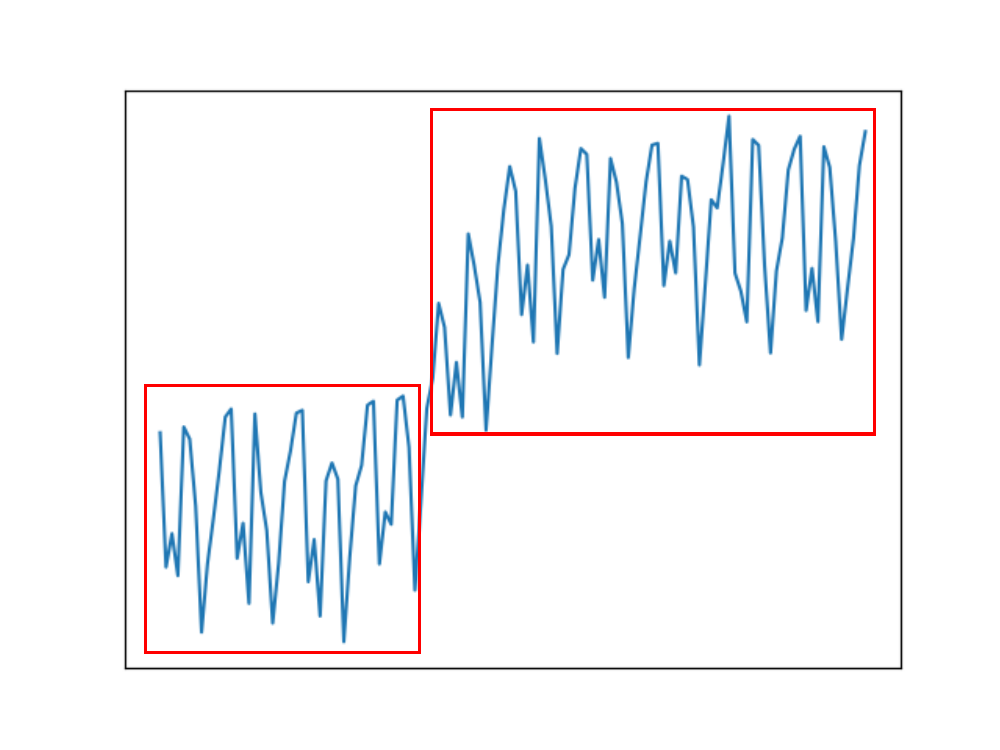}
    \caption{Shifting}
    \label{fig:pattern_shift}
\end{subfigure}
\caption{Visualization of data with different characteristics.}
\label{fig:ts_pattern_eg}
\end{figure}

In real-world scenarios, time series exhibit diverse characteristics, such as \textit{trend}, \textit{seasonality}, \textit{shifting}, \textit{transition}, and \textit{stationarity}, which are illustrated in Figure~\ref{fig:ts_pattern_eg}. 
\textit{Trend} refers to the long-term changes or patterns that occur over time and represents the change modality of the data. 
\textit{Seasonality} indicates changes in a time series that recur at specific intervals, with the mean of the time series being constant and the variance being finite for all observations. 
\textit{Shifting} refers to alterations in the probability distribution over time.
\textit{Transition} captures the regular and identifiable fixed features present in a time series, such as the clear manifestation of trends, periodicity, or the simultaneous presence of both seasonality and trend. 
\textit{Stationarity} refers to the mean of any observation in a time series is constant and the variance is finite for all observations.
\textcolor{black}{Code and pseudocode for computing these characteristics are available in our code repository.}

\subsection{Types of Time Series Anomalies}
\label{Types of Time Series Anomalies}
As is shown in Figure~\ref{fig:anomaly-type}, time series anomalies can be categorized into point anomalies and subsequence anomalies. Point anomalies include \textit{global anomalies} and \textit{contextual anomalies}, while subsequence anomalies encompass \textit{shapelet anomalies}, \textit{seasonal anomalies}, and \textit{trend anomalies}~\cite{lai2021revisiting}. \textcolor{black}{Additionally, there are \textit{fixed anomalies}, which are composed of a mixture of these anomalies.} Global anomalies refer to points that deviate significantly from the rest of the data points, often appearing as spikes. Next, contextual anomalies are points that deviate from their context, defined as neighboring time points within specific ranges, representing small deviations in sequential data. Shapelet anomalies involve subsequences with dissimilar basic shapelets compared to the normal shapelet component of the sequence. Seasonal anomalies exhibit unusual seasonal patterns compared to the overall seasonality, maintaining similar basic shapelets and trends. Trend anomalies involve subsequences that bring about a significant shift in the trend of the time series, resulting in a permanent change in the mean of the data. While retaining basic shapelets and seasonality of normal patterns, trend anomalies undergo drastic alterations in slopes.

\section{TAB: Benchmark Details}
\label{TAB: BENCHMARK DETAILS}
We proceed to cover the design of the TAB benchmark. First, we outline the process of dataset collection and filtering in TAB and their statistical informations~(Section~\ref{sec:Datasets}). Second, we categorize the existing methods supported~(Section~\ref{Comparison methods}). Third, we cover evaluation strategies and metrics~(Section~\ref{sec:Evaluation Settings}). Finally, we describe the full benchmark pipeline (Section~\ref{sec:Unified Pipeline}).


\subsection{Datasets}
\label{sec:Datasets}   
We equip TAB with a robust collection of \textcolor{black}{220 multivariate time series, which are derived from 29 multivariate datasets, along with 1,635 univariate time series that come from 15 univariate datasets}, sourced from community literature, providing researchers with a solid reference foundation. All datasets are formatted consistently, ensuring ease of use. This comprehensive collection spans a wide range of domains and characteristics. Detailed values of key characteristics and anomaly types for both multivariate and univariate datasets, along with their classification, are available in our code repository. This initiative enhances accessibility, tackling challenges such as inconsistent formats, varied documentation, and the time-consuming process of dataset collection.



\renewcommand{\arraystretch}{0.85} %

\begin{table}[t!]
\caption{\textcolor{black}{Statistics of multivariate datasets.}}
\label{Multivariate datasets}
\resizebox{1\columnwidth}{!}{
\begin{tabular}{@{}llrrrrr@{}}
\toprule
\textbf{Dataset}      & \textbf{Domain}   & \textbf{Dim}   & 
 \textbf{\begin{tabular}[c]{@{}c@{}}Avg. \\ AR~(\%) \end{tabular}} &
 \textbf{\begin{tabular}[c]{@{}c@{}}Avg Total\\ Length\end{tabular}}& 
 \textbf{\begin{tabular}[c]{@{}c@{}}Avg Test\\ Length\end{tabular}} & 
 \textbf{\begin{tabular}[c]{@{}c@{}}Series\\ Count\end{tabular}}\\ 
 \midrule
MSL~\cite{hundman2018detecting}      & Spacecraft  & 55   & 5.88    & 132,046            & 73,729 &1 \\
SMAP~\cite{hundman2018detecting}       & Spacecraft  & 25   & 9.72    & 562,800            & 427,617 &1\\
SWAN~\cite{DVN/EBCFKM_2020}        & Space Weather    & 38 & 23.80     & 120,000           & 60,000   &1\\

PSM~\cite{abdulaal2021practical}     & Server Machine  & 25   & 11.07    & 220,322            & 87,841  &1\\
SMD~\cite{su2019robust}       & Server Machine  & 38   & 2.08    & 1,416,825            & 708,420  &1\\

SWAT~\cite{mathur2016swat}      & Water treatment  & 51   & 5.78    & 944,919           & 449,919   &1\\
GECCO~\cite{moritz2018gecco}        & Water treatment  & 9 & 1.25   & 138,521              & 69,261   &1\\
PUMP~\cite{feng2021time}       & Water treatment     & 44    & 6.54   & 220,302          & 143,401  &1\\

Creditcard~\cite{dal2015calibrating}        & Finance  & 29 & 0.17    & 284,807             & 142,404   &1\\

CICIDS~\cite{sharafaldin2018toward}        & Web    & 72 & 1.28   & 170,231             & 85,116   &1\\
KDDcup99~\cite{kddcup99}        & Web    & 3 & 0.21   & 288,535             & 143,763   &2\\

CalIt2~\cite{asuncion2007uci}  & Visitors flowrate  & 2    & 4.09    & 5,040         & 2,520  &1\\

Genesis~\cite{von2018anomaly}        & Machinery   & 18     & 0.31   & 16,220          & 12,616 &1 \\
SKAB~\cite{skab}      & Machinery   & 8   & 3.65   & 10,505          & 1,100  &34\\
GHL~\cite{filonov2016multivariate}     & Machinery  & 19   & 1.13    & 199,001           & 149,653  &25\\
CATSv2~\cite{fleith2023controlled}  & Machinery & 17   & 2.60    & 172,500           & 144,028  &4\\

Daphnet~\cite{asuncion2007uci}         & Movement & 9  & 5.75   & 87,724             & 58,483 &7\\
OPP~\cite{roggen2010collecting}   & Movement & 248  & 1.75    & 27,518            & 21,313   &2\\

NYC~\cite{cui2016comparative}     & Transport  & 3  & 0.57    & 17,520            & 4,416  &1\\

Exathlon~\cite{jacob2020exathlon}        & Application Server & multi  & 9.72    & 67,952            & 52,147  &30\\
ASD~\cite{li2021multivariate}       & Application Server    & 19    & 1.55   & 12,848           & 4,320  &12\\

TODS~\cite{lai2021revisiting} & Synthetic & 5    & 1.27   & 25,000          & 5,000  &12 \\
GutenTAG~\cite{WenigEtAl2022TimeEval}    & Synthetic & multi  & 0.75    & 20,000            & 10,000   &48\\

MITDB~\cite{goldberger2000physiobank}     & Health  & 2    & 2.72    & 141,667           & 106,250  &6\\
SVDB~\cite{greenwald1990improved}     & Health  & 2    & 3.14    & 110,133           & 86,373  &6\\
LTDB~\cite{goldberger2000physiobank}     & Health  & 2    & 15.57   & 100,000           & 87,285   &5\\
DLR~\cite{yoon2020ultrafast}     & Health  & 9    & 2.28   & 23,130           &11,565    &1\\
ECG~\cite{yoon2020ultrafast}     & Health  & 32    & 16.27   & 112,209           &56,105    &1\\

TAO~\cite{tao2023}    & Climate & 3    & 8.33    & 10,000            & 8,000    &12\\
\bottomrule
\multicolumn{6}{l}{AR: anomaly ratio, ``multi'': the dimensionality varies across time series in the dataset.}
\end{tabular}
}
\end{table}
\renewcommand{\arraystretch}{1}

\begin{table}[t]
\footnotesize
\caption{Statistics of univariate datasets.}
\label{Univariate datasets}
\resizebox{1\columnwidth}{!}{
\begin{tabular}{llrrrr}
\toprule
\textbf{Dataset} &
\textbf{Domain} &
\textbf{\begin{tabular}[c]{@{}c@{}}Avg. \\ AR(\%)\end{tabular}} &
\textbf{\begin{tabular}[c]{@{}c@{}}Avg Total\\ Length\end{tabular}} &
\textbf{\begin{tabular}[c]{@{}c@{}}Avg Test\\ Length\end{tabular}} &
\textbf{\begin{tabular}[c]{@{}c@{}}Series\\ Count\end{tabular}} \\
\midrule
GAIA & AIOps & 1.26 & 9,776.8 & 8,799.3 & 184 \\
ECG~\cite{moody2001impact} & Health & 4.89 & 229,990.9 & 206,991.8 & 22 \\
IOPS~\cite{ren2019time} & Web & 2.15 & 101,430.7 & 91,288.3 & 11 \\
KDD21~\cite{KDD21} & Multiple & 0.58 & 65,902.6 & 47,294.0 & 243 \\
MGAB~\cite{thill2020markusthill} & Mackey-Glass & 0.20 & 100,000.0 & 90,000.0 & 6 \\
NAB~\cite{ahmad2017unsupervised} & Multiple & 9.84 & 7,114.8 & 3,557.0 & 45 \\
YAHOO~\cite{laptev2015s5} & Multiple & 0.63 & 1,570.0 & 784.8 & 346  \\\hline
NASA-MSL~\cite{benecki2021detecting} & Spacecraft & 3.85 & 5,166.6 & 2,896.8 & 22 \\
NASA-SMAP~\cite{benecki2021detecting} & Spacecraft & 2.20 & 10,538.5 & 7,929.7 & 35 \\
Daphnet~\cite{bachlin2009wearable} & Movement & 3.38 & 12,342.9 & 11,108.6 & 21 \\
GHL~\cite{filonov2016multivariate} & Machinery & 0.43 & 200,001.0 & 180,001.0 & 1 \\
Genesis~\cite{von2018anomaly} & Machinery & 0.31 & 16,220.0 & 14,598.0 & 1 \\
OPP~\cite{roggen2010collecting} & Movement & 4.12 & 31,358.6 & 28,223.2 & 462 \\
SMD~\cite{su2019robust} & Server Machine & 2.63 & 25,810.4 & 23,229.9 & 184 \\
SVDB~\cite{greenwald1990improved} & Health & 4.87 & 230,400.0 & 207,360.0 & 52 \\
\bottomrule
\multicolumn{6}{l}{AR: anomaly ratio.}
\end{tabular}
}
\end{table}

\subsubsection{Multivariate time series}
Table~\ref{Multivariate datasets} lists statistics of the 29 public multivariate datasets from 14 domains. We observe that the range of feature dimensions varies from 2 to 248 and the sequence length changes from 5,040 to \textcolor{black}{1,416,825}. Some datasets contain multiple multivariate time series. For instance, the Daphnet, SKAB, Exathlon, ASD, and TODS datasets contain 7, 34, 30, 12, and 12 multivariate time series, respectively. In the experiments, we report the average evaluation metrics across all time series for each dataset. 

\subsubsection{Univariate time series}
Table~\ref{Univariate datasets} summarizes statistical information of the univariate datasets. The first seven datasets~(split by horizontal line) are the most commonly used and are proposed for univariate time series originally. The remaining eight datasets are transformed from multivariate time series, which is consistent with TAB-UAD~\cite{paparrizos2022tsb}. \textcolor{black}{Specifically, for multivariate time series, we treat it as multiple univariate time series. We then run the AD method on each univariate time series separately and retain those univariate series where at least one method achieves AUC-ROC \textgreater~  0.85}. To ensure data quality, we proceed to filter according to the following two rules: 1)~have no anomaly; 2)~have anomaly ratio \textgreater~ 10\%~\cite{schmidl2022anomaly}, resulting in a total of 1,635 time series.
 
Due to legacy issues, we employ datasets used extensively in existing studies. However, we note that there are concerns regarding the ground-truth anomaly labels in these datasets~\cite{wu2021current}. It thus remains of interest to address such data-quality issues.

\begin{figure*}[t!]
    \centering
    \includegraphics[width=1\linewidth]{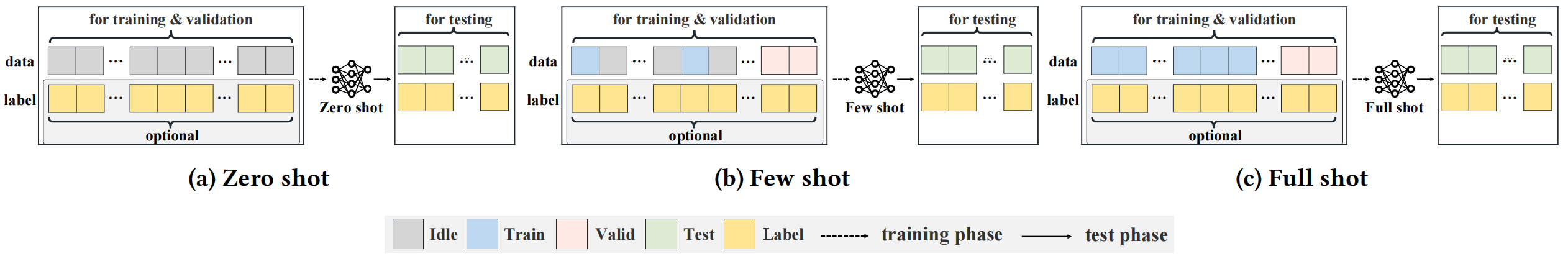}
    \caption{Evaluation strategies. }
    \label{Evaluation Settings}
\end{figure*}

\subsection{Comparison Methods}
\label{Comparison methods}
To investigate the strengths and limitations of different TSAD methods, we evaluate a large number of methods, which can be classified into non-learning~(NL), machine learning~(ML), deep learning~(DL), LLM-based, and Time-series pre-trained~(TS pre-trained) methods. Overall, \textbf{46 methods} are included for a comprehensive UTSAD comparison. S2G, LSTi, DWT, ARIMA, SARIMA, ZMS, Stat, and SR are removed from MTSAD; thus, \textbf{40 methods} are used for MTSAD comparison. For conciseness, we omit their detailed descriptions and categorize them briefly based on their distinct technical approaches, which are reported in Table~\ref{Overview of comparison methods.}. 

\begin{table}[t!]
\footnotesize
\caption{\textcolor{black}{Overview of comparison methods.}}
\label{Overview of comparison methods.}
\setlength{\tabcolsep}{2pt}
\resizebox{1\columnwidth}{!}{
\begin{tabular}{|c|c|c|c|c|c|}
\hline
\textbf{Method} &\textbf{Abbrev.}&\textbf{Area}& \textbf{Key Technology} & \textbf{UTAD} & \textbf{MTAD}\\ \hline
\resizebox{1.5mm}{1.5mm}{\myproximity}\hspace{0.5mm}\textbf{LOF}~\cite{breunig2000lof} & LOF &\multirow{11}{*}{\rotatebox{90}{Non Learning}} & Density & \scalebox{1.25}[1]{$\surd$} &  $\surd$  \\ \cline{1-2} \cline{4-6}

\resizebox{1.5mm}{1.5mm}{\myclustering}\hspace{0.5mm}\textbf{CBLOF}~\cite{he2003discovering} & CBLOF&  & Clustering & $\surd$ &  $\surd$\\ \cline{1-2} \cline{4-6}

\resizebox{1.5mm}{1.5mm}{\mydistribution}\hspace{0.5mm}\textbf{HBOS}~\cite{goldstein2012histogram} & HBOS& & Histogram & $\surd$ &  $\surd$\\ \cline{1-2} \cline{4-6}
\resizebox{1.5mm}{1.5mm}{\myforecasting}\hspace{0.5mm}\textbf{ARIMA}~\cite{hyndman2018forecasting} & ARIMA& & \begin{tabular}[c]{@{}c@{}}Autoregressive + Moving Average\end{tabular} & $\surd$ &  \scalebox{1.25}[1]{$\times$}\\ \cline{1-2} \cline{4-6}
\resizebox{1.5mm}{1.5mm}{\myforecasting}\hspace{0.5mm}\textbf{SARIMA}~\cite{greis2018comparing} &  SARIMA&&  Seasonal ARIMA& $\surd$ &  \scalebox{1.25}[1]{$\times$}\\ \cline{1-2} \cline{4-6}
\resizebox{1.5mm}{1.5mm}{\mydistribution}\hspace{0.5mm}\textbf{StatThreshold}~\cite{bhatnagar2021merlion} &  Stat& &  Static Thresholding& $\surd$ &  \scalebox{1.25}[1]{$\times$}\\ \cline{1-2} \cline{4-6} 

\resizebox{1.5mm}{1.5mm}{\mydistribution}\hspace{0.5mm}\textbf{DWT-MLEAD}~\cite{thill2017time} & DWT& &  Discrete Wavelet Transform& $\surd$ &  \scalebox{1.25}[1]{$\times$}\\ \cline{1-2} \cline{4-6}

\resizebox{1.5mm}{1.5mm}{\mydiscord}\hspace{0.5mm}\textbf{Left STAMPi}~\cite{yeh2016matrix} & LSTi& & Matrix  Profile& $\surd$ &  \scalebox{1.25}[1]{$\times$}\\ \cline{1-2} \cline{4-6}

\resizebox{1.5mm}{1.5mm}{\mydiscord}\hspace{0.5mm}\textbf{ZMS}~\cite{bhatnagar2021merlion} & ZMS& &  Multiple z-score& $\surd$ &  \scalebox{1.25}[1]{$\times$}\\ \cline{1-2} \cline{4-6}

\resizebox{1.5mm}{1.5mm}{\mygraph}\hspace{0.5mm}\textbf{Series2Graph}~\cite{BoniolP20} & S2G& & Graph representation of time series & $\surd$ &  \scalebox{1.25}[1]{$\times$}\\ \hline

\resizebox{1.5mm}{1.5mm}{\mydistribution}\hspace{0.5mm}\textbf{OCSVM}~\cite{scholkopf1999support}&Ocsvm & \multirow{9}{*}{\rotatebox{90}{Machine Learning}} & Classification & $\surd$ &  $\surd$\\ \cline{1-2} \cline{4-6}
\resizebox{1.5mm}{1.5mm}{\myreconstruction}\hspace{0.5mm}\textbf{DeepPoint}~\cite{bhatnagar2021merlion}& DP&  & Multilayer Perceptron& $\surd$ &  $\surd$\\ \cline{1-2} \cline{4-6}
\resizebox{1.5mm}{1.5mm}{\myproximity}\hspace{0.5mm}\textbf{KNN}~\cite{ramaswamy2000efficient} &KNN&  & K-nearest Neighbors & $\surd$ &  $\surd$\\ \cline{1-2} \cline{4-6}

\resizebox{1.5mm}{1.5mm}{\myclustering}\hspace{0.5mm}\textbf{KMeans}~\cite{yairi2001fault} &KMeans&  & K-Means & $\surd$ &  $\surd$\\ \cline{1-2} \cline{4-6}

\resizebox{1.5mm}{1.5mm}{\mytree}\hspace{0.5mm}\textbf{Isolation Forest}~\cite{liu2008isolation} &IF&  & Isolation Tree & $\surd$ &  $\surd$\\ \cline{1-2} \cline{4-6} 

\resizebox{1.5mm}{1.5mm}{\mytree}\hspace{0.5mm}\textbf{EIF}~\cite{hariri2019extended} &EIF&  & Isolation Tree & $\surd$ &  $\surd$\\ \cline{1-2} \cline{4-6} 

\resizebox{1.5mm}{1.5mm}{\myreconstruction}\hspace{0.5mm}\textbf{Spectral Residual}~\cite{ren2019time} &SR&  & \begin{tabular}[c]{@{}c@{}}Frequency Spectrum + \\Fourier Transform\end{tabular} & $\surd$ &  \scalebox{1.25}[1]{$\times$}\\ \cline{1-2} \cline{4-6}
\resizebox{1.5mm}{1.5mm}{\mydistribution}\hspace{0.5mm}\textbf{LODA}~\cite{pevny2016loda} &LODA&  & Ensemble Learning & $\surd$ &  $\surd$\\ \cline{1-2} \cline{4-6}
\resizebox{1.5mm}{1.5mm}{\myencoding}\hspace{0.5mm}\textbf{PCA}~\cite{shyu2003novel} &PCA&  & \begin{tabular}[c]{@{}c@{}} Dimensionality reduction + Restruction\end{tabular} & $\surd$ &  $\surd$\\  \hline
\resizebox{1.5mm}{1.5mm}{\myreconstruction}\hspace{0.5mm}\textbf{DAGMM}~\cite{zong2018deep} &Dagmm&  & \begin{tabular}[c]{@{}c@{}} Deep autoencoder + Gaussian Mixture\end{tabular} & $\surd$ &  $\surd$\\ \cline{1-2} \cline{4-6}

\resizebox{1.5mm}{1.5mm}{\myforecasting}\hspace{0.5mm}\textbf{Torsk}~\cite{heim2019adaptive} &Torsk&  & Echo State Networks & $\surd$ &  $\surd$\\ \cline{1-2} \cline{4-6}

\resizebox{1.5mm}{1.5mm}{\myreconstruction}\hspace{0.5mm}\textbf{iTransformer}~\cite{liu2023itransformer} &iTrans& \multirow{17}{*}{\rotatebox{90}{{Deep Learning}}} &  Inverted Transformer & $\surd$ &  $\surd$\\ \cline{1-2} \cline{4-6}
\resizebox{1.5mm}{1.5mm}{\myreconstruction}\hspace{0.5mm}\textbf{TimesNet}~\cite{wu2022timesnet} &TsNet&  & \begin{tabular}[c]{@{}c@{}}Temporal 2D-variations +\\ Multi-periodicity\end{tabular} & $\surd$ &  $\surd$\\ \cline{1-2} \cline{4-6}
\resizebox{1.5mm}{1.5mm}{\myreconstruction}\hspace{0.5mm}\textbf{DUET} ~\cite{qiu2025duet} &DUET&  & \begin{tabular}[c]{@{}c@{}}Temporal Clustering +\\Channel Soft Clustering \end{tabular} & $\times$ &  $\surd$\\ \cline{1-2} \cline{4-6}
\resizebox{1.5mm}{1.5mm}{\myreconstruction}\hspace{0.5mm}\begin{tabular}[c]{@{}c@{}}\textbf{Anomaly}\\ \textbf{Transformer}~\cite{xu2021anomaly}\end{tabular} &ATrans&  &  \begin{tabular}[c]{@{}c@{}}Anomaly Attention +\\ Association Discrepancy\end{tabular} & $\surd$ &  $\surd$\\ \cline{1-2} \cline{4-6} 

\resizebox{1.5mm}{1.5mm}{\myreconstruction}\hspace{0.5mm}\textbf{PatchTST}~\cite{nie2022time}&Patch &  & Channel Independent + Patchify & $\surd$ &  $\surd$\\ \cline{1-2} \cline{4-6} 
\resizebox{1.5mm}{1.5mm}{\myreconstruction}\hspace{0.5mm}\textbf{ModernTCN}~\cite{luo2024moderntcn} &Modern&  & TCN with bigger receptive field & $\surd$ &  $\surd$\\ \cline{1-2} \cline{4-6} 
\resizebox{1.5mm}{1.5mm}{\myreconstruction}\hspace{0.5mm}\textbf{TranAD}~\cite{tuli2022tranad} & TranAD & & \begin{tabular}[c]{@{}c@{}}Self-conditioning + Meta Learning +\\ Two-Phase Adversarial Training\end{tabular}& $\surd$ &  $\surd$\\ \cline{1-2} \cline{4-6}
\resizebox{1.5mm}{1.5mm}{\myreconstruction}\hspace{0.5mm}\textbf{DualTF}~\cite{dualtf}& DualTF &  & \begin{tabular}[c]{@{}c@{}}Time-Frequency combination +\\ Anomaly Transformer \end{tabular} & $\surd$ &  $\surd$\\ \cline{1-2} \cline{4-6}

\resizebox{1.5mm}{1.5mm}{\myreconstruction}\hspace{0.5mm}\textbf{AE}~\cite{sakurada2014anomaly} &AE&  & AutoEncoder & $\surd$ &  $\surd$\\ \cline{1-2} \cline{4-6}
\resizebox{1.5mm}{1.5mm}{\myreconstruction}\hspace{0.5mm}\textbf{VAE}~\cite{kingma2013auto} & VAE &  & AutoEncoder & $\surd$ &  $\surd$\\ \cline{1-2} \cline{4-6}

\resizebox{1.5mm}{1.5mm}{\myreconstruction}\hspace{0.5mm}\textbf{NLinear}~\cite{zeng2023transformers}&NLin &  & \begin{tabular}[c]{@{}c@{}}Linear + Simple normalization\end{tabular} & $\surd$ &  $\surd$\\ \cline{1-2} \cline{4-6}
\resizebox{1.5mm}{1.5mm}{\myreconstruction}\hspace{0.5mm}\textbf{DLinear}~\cite{zeng2023transformers} &DLin&  & \begin{tabular}[c]{@{}c@{}}Linear + Decomposition\end{tabular} & $\surd$ &  $\surd$\\ \cline{1-2} \cline{4-6} 
\resizebox{1.5mm}{1.5mm}{\myforecasting}\hspace{0.5mm}\textbf{LSTMED}~\cite{bhatnagar2021merlion} & LSTM &  & LSTM + Encoder-Decoder-based & $\surd$ &  $\surd$\\ \cline{1-2} \cline{4-6}
\resizebox{1.5mm}{1.5mm}{\mycontrast}\hspace{0.5mm}\textbf{DCdetector}~\cite{yang2023dcdetector} &DC&  & \begin{tabular}[c]{@{}c@{}}Multi-scale Dual Attention +\\ Contrastive Learning\end{tabular} & $\surd$ &  $\surd$\\ \cline{1-2} \cline{4-6}
\resizebox{1.5mm}{1.5mm}{\mycontrast}\hspace{0.5mm}\textbf{ContraAD} ~\cite{zhuang2025ContraAD} &ConAD&  & \begin{tabular}[c]{@{}c@{}}Contrastive Learning\end{tabular} & $\surd$ &  $\surd$\\ \cline{1-2} \cline{4-6}
\resizebox{1.5mm}{1.5mm}{\myreconstruction}\hspace{0.5mm}\textbf{CATCH} ~\cite{wu2024catch} &CATCH&  &  \begin{tabular}[c]{@{}c@{}}Frequency Reconstruction + \\Channel Fusion \end{tabular}& $\times$ &  $\surd$\\ \cline{1-3} \cline{4-6}

\resizebox{1.5mm}{1.5mm}{\myforecasting}\hspace{0.5mm}\textbf{GPT4TS}~\cite{gpt4ts}& GPT4TS& \multirow{5}{*}{\rotatebox{90}{{LLM-based}}} & \begin{tabular}[c]{@{}c@{}}Parameter-Efficient Fine-Tuning\end{tabular} & $\surd$ &  $\surd$ \\ \cline{1-2}\cline{4-6}
\resizebox{1.5mm}{1.5mm}{\myreconstruction}\hspace{0.5mm}\textbf{UniTime}~\cite{unitime} &UniTime&  & \begin{tabular}[c]{@{}c@{}}Domain Instructions +\\ Masking Technique +\\ Cross-Domain Training Strategy\end{tabular} & $\surd$ &  $\surd$ \\ \cline{1-2}\cline{4-6}
\resizebox{1.5mm}{1.5mm}{\myforecasting}\hspace{0.5mm}\textbf{CALF}~\cite{CALF} &CALF&  & \begin{tabular}[c]{@{}c@{}}cross-modal
Fine-Tuning + \\Parameter-Efficient Fine-Tuning\end{tabular} & $\surd$ &  $\surd$ \\ \cline{1-2}\cline{4-6}
\resizebox{1.5mm}{1.5mm}{\myreconstruction}\hspace{0.5mm}\textbf{LLMMixer}~\cite{LLMMixer} &LLMMixer&  & \begin{tabular}[c]{@{}c@{}}Multi-scale time series decomposition +\\ Prompt Embedding \end{tabular} & $\surd$ &  $\surd$ \\ \hline

\resizebox{1.5mm}{1.5mm}{\myforecasting}\hspace{0.5mm}\textbf{Timer}~\cite{timer} &Timer & \multirow{10}{*}{\rotatebox{90}{{Pre-trained}}} & \begin{tabular}[c]{@{}c@{}}Single-Series Sequence + Autoregressive\end{tabular}  & $\surd$ &  $\surd$ \\ \cline{1-2}\cline{4-6} 
\resizebox{1.5mm}{1.5mm}{\myforecasting}\hspace{0.5mm}\textbf{TinyTimeMixer}~\cite{ekambaram2024ttms} &TTM&  & \begin{tabular}[c]{@{}c@{}}Adaptive Patching + Exogenous Mixer + \\ Resolution Prefix Tuning \end{tabular} & $\surd$ &  $\surd$ \\ \cline{1-2}\cline{4-6} 
\resizebox{1.5mm}{1.5mm}
{\myforecasting}\hspace{0.5mm}\textbf{TimesFM}~\cite{TimesFM} &TimesFM & & \begin{tabular}[c]{@{}c@{}}Patching + Autoregressive\end{tabular} & $\surd$ &  $\surd$ \\ \cline{1-2}\cline{4-6} 
\resizebox{1.5mm}{1.5mm}{\myreconstruction}\hspace{0.5mm}\textbf{UniTS}~\cite{units} &UniTS & & \begin{tabular}[c]{@{}c@{}}Dynamic Linear + Prompt Learning +\\ Unified Masked Reconstruction\end{tabular} & $\surd$ &  $\surd$ \\ \cline{1-2}\cline{4-6} 
\resizebox{1.5mm}{1.5mm}{\myreconstruction}\hspace{0.5mm}\textbf{Moment} ~\cite{moment} &Moment&  & \begin{tabular}[c]{@{}c@{}}Masking Technique +\\ Lightweight Reconstruction Head\end{tabular} & $\surd$ &  $\surd$ \\  \cline{1-2}\cline{4-6} 
\resizebox{1.5mm}{1.5mm}{\myreconstruction}\hspace{0.5mm}\textbf{DADA}~\cite{DADA} &DADA & & \begin{tabular}[c]{@{}c@{}}Adaptive Bottlenecks + \\ Dual Adversarial Decoders\end{tabular} & $\surd$ &  $\surd$ \\ \cline{1-2}\cline{4-6} 
\resizebox{1.5mm}{1.5mm}{\myforecasting}\hspace{0.5mm}\textbf{Chronos}~\cite{ansari2024chronos} &Chronos & & \begin{tabular}[c]{@{}c@{}}Time Series Tokenization + \\ Data Augmentation + Probability Prediction\end{tabular} & $\surd$ &  $\surd$ \\
\hline

\multicolumn{6}{l}{
\begin{tabular}[l]{@{}l@{}}
distance-based (\textcolor{orange}{\textbf{orange}}), density-based (\textcolor{magenta}{\textbf{red}}), prediction-based (\textcolor{cyan}{\textbf{blue}}), contrast-based (\textcolor[rgb]{0,0.5,0}{\textbf{green}})
\\
\resizebox{1.5mm}{1.5mm}{\myproximity}\hspace{0.5mm}proximity-based, 
\resizebox{1.5mm}{1.5mm}{\myclustering}\hspace{0.5mm}clustering-based, 
\resizebox{1.5mm}{1.5mm}{\mydiscord}\hspace{0.5mm}discord-based, 
\resizebox{1.5mm}{1.5mm}{\mydistribution}\hspace{0.5mm}distribution-based, 
\resizebox{1.5mm}{1.5mm}{\mygraph}\hspace{0.5mm}graph-based,
\resizebox{1.5mm}{1.5mm}{\mytree}\hspace{0.5mm}tree-based,
\\ 
\resizebox{1.5mm}{1.5mm}{\myencoding}\hspace{0.5mm}encoding-based, 
\resizebox{1.5mm}{1.5mm}{\myforecasting}\hspace{0.5mm}forecasting-based, 
\resizebox{1.5mm}{1.5mm}{\myreconstruction}\hspace{0.5mm}reconstruction-based, 
\resizebox{1.5mm}{1.5mm}{\mycontrast}\hspace{0.5mm}contrast-based
\end{tabular}
}
\end{tabular}}
\end{table}

\subsection{Evaluation Settings}
\label{sec:Evaluation Settings}






\begin{table*}[t]
\caption{Overview of evaluation metrics supported in TAB.}
\label{Overview of evaluation metrics supported in TAB}
\centering
\resizebox{2\columnwidth}{!}{
\begin{tabular}{@{}c|l|l|l@{}}
\toprule
\multicolumn{1}{l|}{Category} & Metric & Abbreviation & Short summary \\ \midrule
\multirow{10}{*}{Label\_based} & Accuracy & Acc & Measures the proportion of correct predictions among the total number of predictions.\\\cline{2-4}
 & Precision & P & Measures the proportion of true positive predictions among all positive predictions. Important when false positives are costly. \\\cline{2-4}
 & Recall & R & Masures the proportion of true positive predictions among all actual positives. Important when false negatives are costly. \\\cline{2-4}
 & F1-score & F1 & A balanced metric when you need to account for both precision and recall. \\\cline{2-4}
 & Range-Precision & R-P~\cite{tatbul2018precision} & These metrics' definitions consider several factors: the ratio of detected anomaly subsequences to the total number of anomalies,  \\
 & Range-Recall & R-R~\cite{tatbul2018precision} &  the ratio of detected point outliers to total point outliers, the relative position of true positives within each anomaly subsequence,\\
 & Range-F1-score & R-F1~\cite{tatbul2018precision} & and the number of fragmented prediction regions corresponding to one real anomaly subsequence.\\ \cline{2-4}
 & Affiliated-Precision & Aff-P~\cite{huet2022local} & These metrics' definitions are the extension of the classical precision/recall/f1-score for time series anomaly detection that is  \\
 & Affiliated-Recall & Aff-R~\cite{huet2022local} & local (each ground truth event is considered separately), parameter-free, and applicable generically on both point and subsequence  \\
 & Affiliated-F1-score & Aff-F1~\cite{huet2022local} & anomalies. Besides the construction of these metrics makes them both theoretically principled and practically useful.\\\cline{1-4}
\multirow{6}{*}{Score\_based} & AUC-PR & A-P~\cite{davis2006relationship} & Measures the area under the curve corresponding to Recall on the x-axis and Precision on the y-axis at various threshold settings. \\\cline{2-4}
 & AUC-ROC & A-R~\cite{fawcett2006introduction} & Measures the area under the curve
corresponding to FPR on the x-axis and TPR on the y-axis at various threshold settings.  \\  \cline{2-4}
 & R-AUC-PR & R-A-P~\cite{paparrizos2022volume} & They mitigate the issue that AUC-PR and AUC-ROC are designed for point-based anomaly detection, where each point is\\ 
 & R-AUC-ROC & R-A-R~\cite{paparrizos2022volume} &    assigned equal weight in calculating the overall AUC. This makes them unsuitable for evaluating subsequence anomalies.\\ \cline{2-4}
 & VUS-PR & V-PR~\cite{paparrizos2022volume} &  VUS is an extension of the ROC and PR curves. It introduces a buffer region at the outliers’ boundaries, thereby accommodating \\ 
 & VUS-ROC & V-ROC~\cite{paparrizos2022volume} &  the false tolerance of labeling in the ground  truth and assigning higher anomaly scores near the outlier boundaries.\\ \bottomrule
\end{tabular}}
\end{table*}

\subsubsection{Evaluation strategies}
TAB supports multiple evaluation strategies, including \textit{zero-shot}, \textit{few-shot}, and \textit{full-shot}---see Figure~\ref{Evaluation Settings}, which support the output of either anomaly scores or anomaly labels.\\


(1) The zero-shot evaluation only uses the test data to evaluate the generalization ability of foundation methods to new datasets, assessing whether the method has truly learned general knowledge from vast amounts of pre-training data. 

(2) The few-shot evaluation only utilizes a subset of training data and full validation data for fine-tuning, reflecting the prediction performance in low-data learning scenarios. This approach assesses whether models can effectively generalize and reason with minimal data support. 

(3) The full-shot evaluation utilizes full train data and validation data for fine-tuning. It evaluates the performance when utilizing all available data, revealing its upper-bound performance.


\subsubsection{Evaluation metrics}

\label{sec:Evaluation metrics}
The metrics supported by TAB can be divided into two categories: Label-based and Score-based. Label-based metrics depend on threshold parameters to convert anomaly scores into anomaly labels. Score-based metrics, on the other hand, evaluate the performance of the model based on the raw anomaly scores it generates. Please refer to Table~\ref{Overview of evaluation metrics supported in TAB} for an overview of these metrics. Please note that TAB computes all metrics to provide a complete picture of each method. 

Utilizing a diverse set of evaluation metrics is crucial for a comprehensive assessment of model performance. Different metrics offer insights from various perspectives, allowing for a nuanced understanding of model capabilities. In addition to the evaluation metrics provided in the table, TAB offers excellent extensibility in its evaluation metrics, allowing users to easily and conveniently customize additional metrics to better meet specific needs.

\begin{figure}[t!]
    \centering
    \includegraphics[width=1\linewidth]{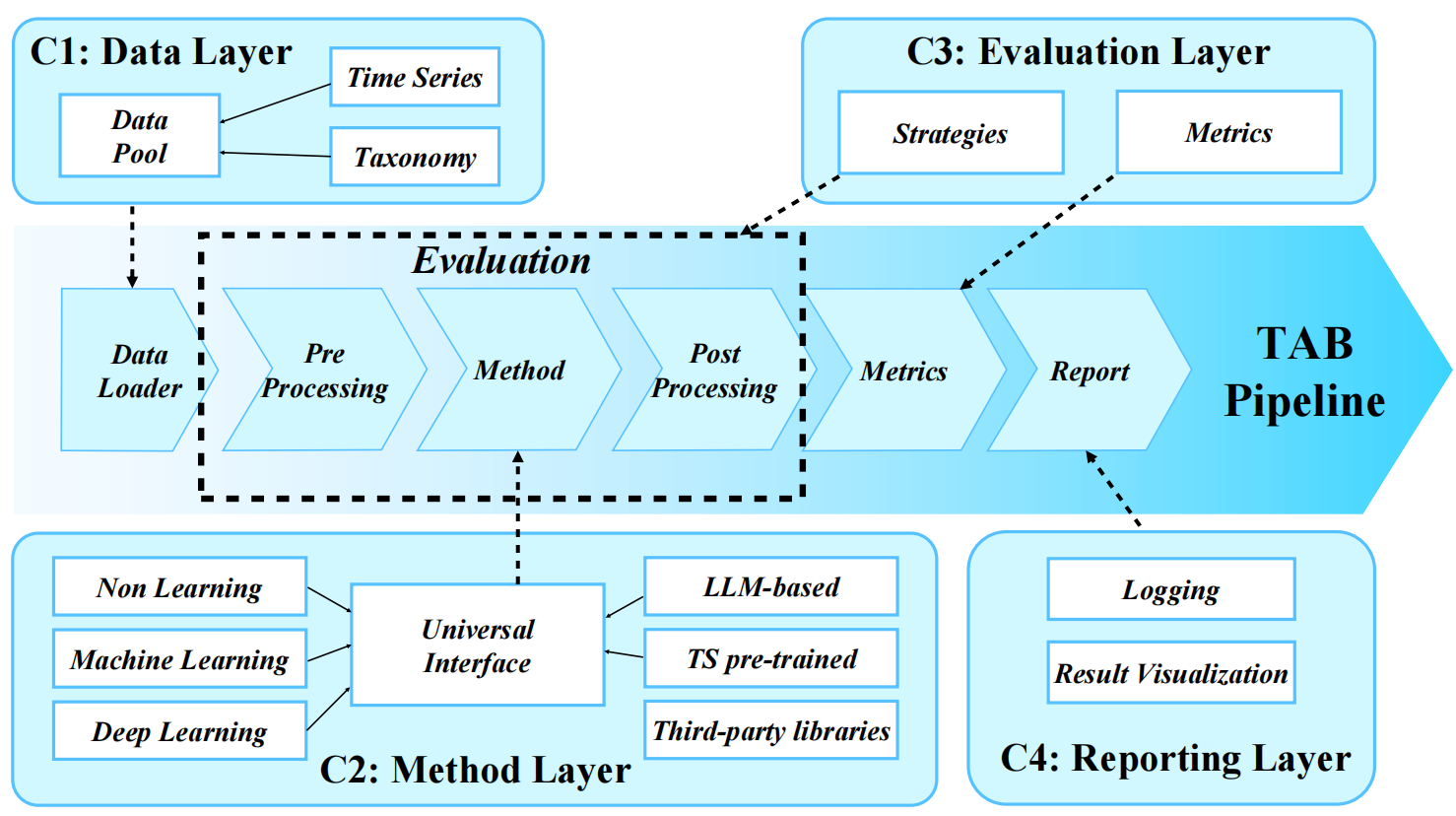}
    \caption{Pipeline of the TAB benchmark. }
    \label{fig:pipeline}
\end{figure}
\subsection{Unified Pipeline}
\label{sec:Unified Pipeline}
\textcolor{black}{To facilitate fair and comprehensive comparisons of methods, TAB proposes a unified pipeline---see Figure~\ref{fig:pipeline} where we first have a data layer that ensures all the input time series are in standardized formats. Thus, we can ensure the data pre-processing and splitting is the same for all the compared AD methods. Then, we have a method layer that can embed both self-implemented and third-party AD methods with a universal interface. Thus, all the compared AD methods are trained and tested in the same workflow. Finally, we have a evaluation layer that has a fixed post-processing such as changing anomaly score to anomaly label. With this layer, the results from all the AD methods are evaluated in the same way. With those three layers, we claim that TAB can ensure a fair comparison among different AD methods. Each component is detailed as follows.}

(1)~\underline{Data Layer.} The data layer comprises both univariate and multivariate time series data from diverse domains. Its organization is carefully designed to categorize time series according to their distinctive characteristics, anomaly ratios, and anomaly types. To improve usability and comparability, time series data is stored in a unified format.

(2)~\underline{Method Layer.} The method layer is constructed elaborately to support different time series AD methods, i.e., non-learning, machine learning, and deep learning methods. 
To comply with other third-party time series AD libraries, e.g., TODS~\cite{tods}, Merlion~\cite{bhatnagar2021merlion}, we design a simple universal interface, such that users can easily integrate third-party libraries into TAB to facilitate fast comparison.


(3)~\underline{Evaluation Layer.} The evaluation layer implements different evaluation strategies and metrics that are introduced in existing studies. Moreover, it also provides customized metrics for a more comprehensive assessment of method performance.

(4)~\underline{Reporting Layer.} The reporting layer encompasses a logging system for tracking information to enable traceability of the experimental settings. Further, it encompasses a visualization module to facilitate a clear understanding of method performance.


\textcolor{black}{Besides the pipeline is flexible in the following ways. 1) It can embed third-party libraries. 2) It supports diverse evaluation strategies that can cover existing common evaluation strategies. 3) It allows users to customize evaluation metrics and evaluation strategies. 4) It can select the dataset to be evaluated based on specified dataset characteristics or anomaly type, such as only evaluating datasets with trends. Besides the pipeline is scalable and we support large-scale parallel operations.}

Moreover, TAB can comply with both CPU and GPU hardware to facilitate the evaluation in various computing environments. Additionally, it supports different modes of program execution (sequential and parallel), providing users with flexible choices.

To sum up, TAB serves as a benchmarking tool tailored for TSAD methods. 
We can swiftly evaluate a new method by integrating it into the model layer. 
Therefore, it can enhance users' understanding and assist in choosing TSAD methods suitable for specific application scenarios.

\begin{table*}[t]
\caption{\textcolor{black}{Performance overview for all 48 methods: Box plots show the score distribution (mean in \textcolor{green}{green} and median in \textcolor{orange}{orange}) for V-PR (VUS-PR) and Aff-F1 (Affiliated-F1) across all univariate (U) and multivariate (M) datasets, and along with training and inference time (seconds), CPU and GPU memory usage (GB). \protect\includegraphics[width=0.03\textwidth]{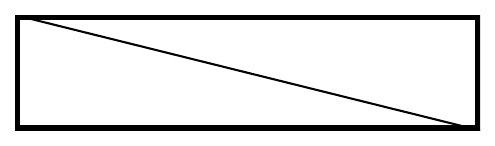}} indicates that the operation is not executable.}
\label{main table}
\centering
\huge
\resizebox{2.01\columnwidth}{!}{
\renewcommand{\arraystretch}{1.02}
\begin{tabular}{c|c|l|c|c|c|c|cc|cc|cc|cc}
\toprule
\multirow{2}{*}{\textbf{Mode}} & \multirow{2}{*}{\textbf{Area}} & \multirow{2}{*}{\textbf{Method}} & \multirow{2}{*}{\textbf{U Aff-F1}} & \multirow{2}{*}{\textbf{U V-PR}} & \multirow{2}{*}{\textbf{M Aff-F1}} & \multirow{2}{*}{\textbf{M V-PR}} & \multicolumn{2}{c|}{\textbf{Train. Time}} & \multicolumn{2}{c|}{\textbf{Infer. Time}} & \multicolumn{2}{c|}{\textbf{CPU}} & \multicolumn{2}{c}{\textbf{GPU}} \\  
& & & & & & & \textbf{U} & \textbf{M} & \textbf{U} & \textbf{M} & \textbf{U} & \textbf{M} & \textbf{U} & \textbf{M} \\
\midrule
\multirow{48}{*}{\rotatebox{90}{\textbf{Full Shot}}} & \multirow{10}{*}{\rotatebox{90}{\textbf{Non Learning}}} &
\resizebox{3.0mm}{3.0mm}{\mygraph}\hspace{0.5mm}S2G & \includegraphics[width=0.25\textwidth]{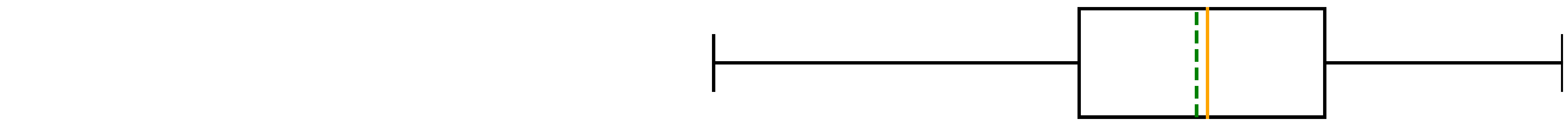} & \includegraphics[width=0.25\textwidth]{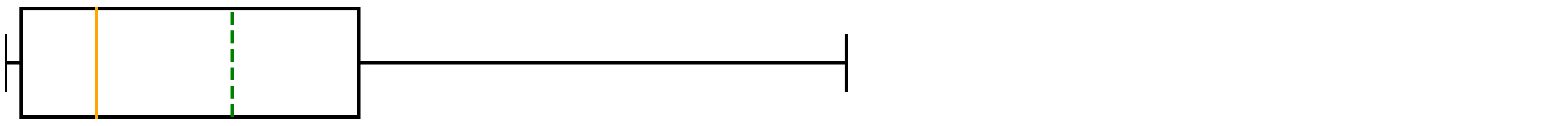} & \includegraphics[width=0.25\textwidth]{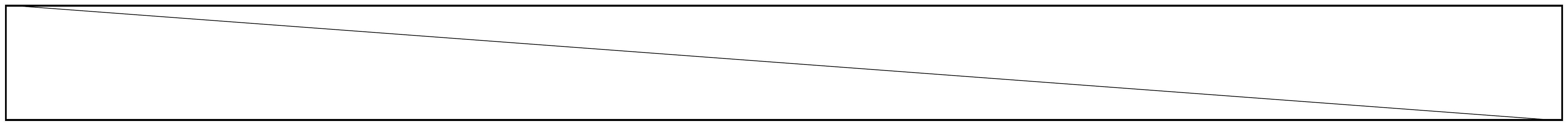}  & \includegraphics[width=0.25\textwidth]{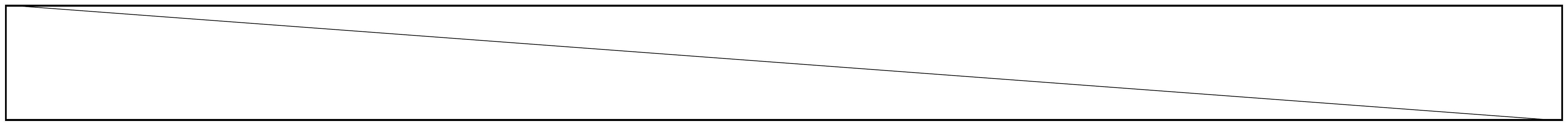} & 0.000 & \includegraphics[width=0.06\textwidth]{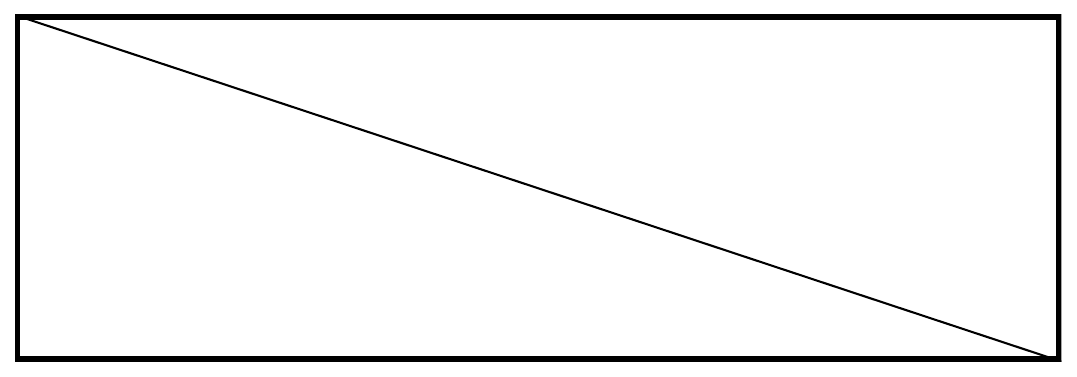} & 1.588 & \includegraphics[width=0.06\textwidth]{figures/main-table/no-data.pdf} & 0.720 & \includegraphics[width=0.06\textwidth]{figures/main-table/no-data.pdf} & 0.000 & \includegraphics[width=0.06\textwidth]{figures/main-table/no-data.pdf} \\
 & & \resizebox{3.0mm}{3.0mm}{\mydiscord}\hspace{0.5mm}LSTi & \includegraphics[width=0.25\textwidth]{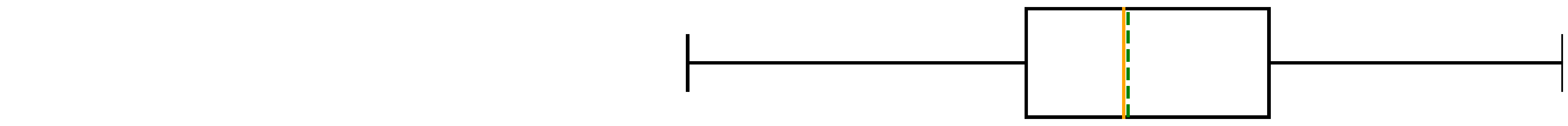} & \includegraphics[width=0.25\textwidth]{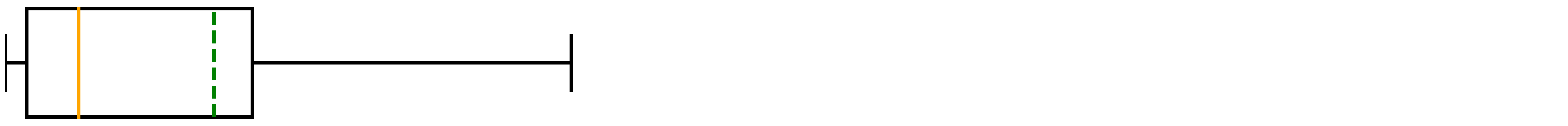} & \includegraphics[width=0.25\textwidth]{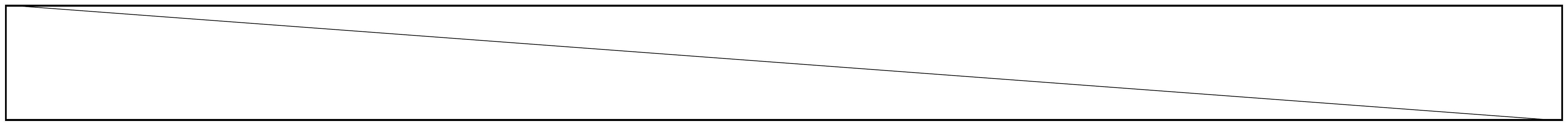}  & \includegraphics[width=0.25\textwidth]{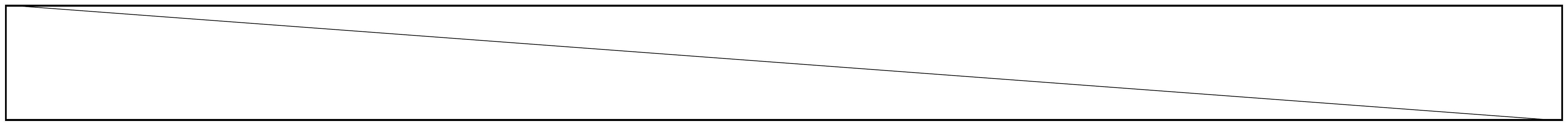} & 0.000 & \includegraphics[width=0.06\textwidth]{figures/main-table/no-data.pdf} & 31.191 & \includegraphics[width=0.06\textwidth]{figures/main-table/no-data.pdf} & 0.786 & \includegraphics[width=0.06\textwidth]{figures/main-table/no-data.pdf} & 0.000 & \includegraphics[width=0.06\textwidth]{figures/main-table/no-data.pdf} \\
 & & \resizebox{3.0mm}{3.0mm}{\mydistribution}\hspace{0.5mm}DWT & \includegraphics[width=0.25\textwidth]{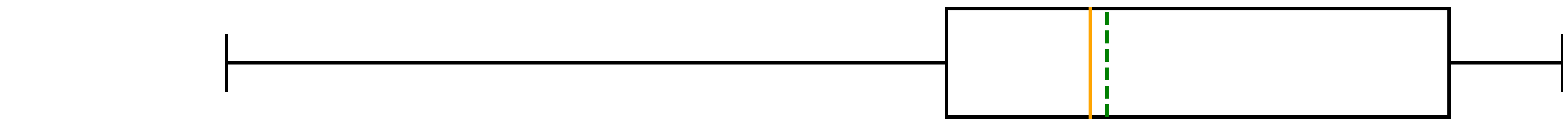} & \includegraphics[width=0.25\textwidth]{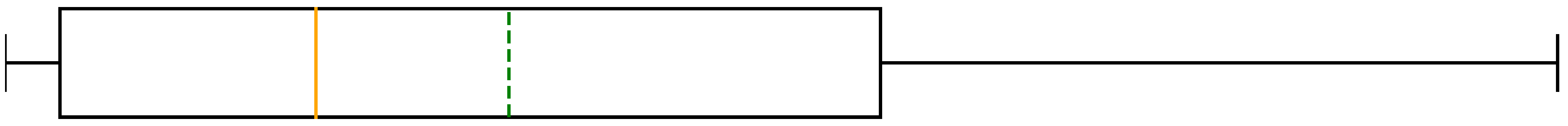} & \includegraphics[width=0.25\textwidth]{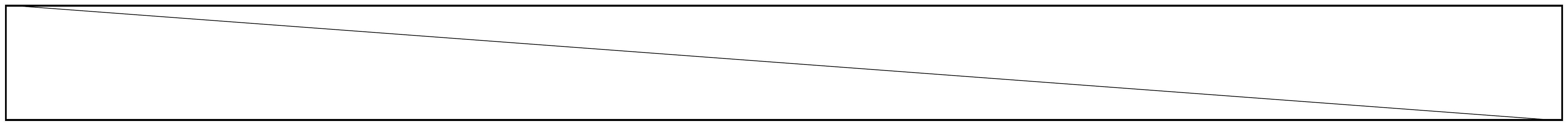}  & \includegraphics[width=0.25\textwidth]{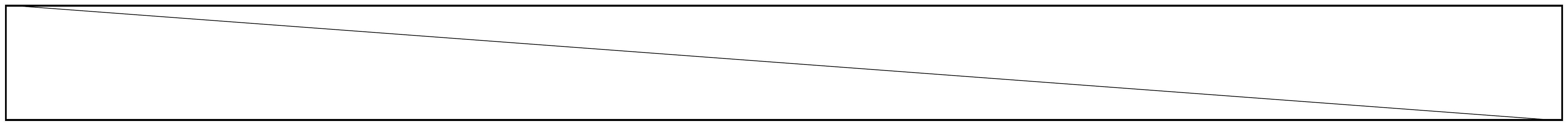} & 0.000 & \includegraphics[width=0.06\textwidth]{figures/main-table/no-data.pdf} & 1.788 & \includegraphics[width=0.06\textwidth]{figures/main-table/no-data.pdf} & 0.721 & \includegraphics[width=0.06\textwidth]{figures/main-table/no-data.pdf} & 0.000 & \includegraphics[width=0.06\textwidth]{figures/main-table/no-data.pdf} \\
 & & \resizebox{3.0mm}{3.0mm}{\myproximity}\hspace{0.5mm}LOF & \includegraphics[width=0.25\textwidth]{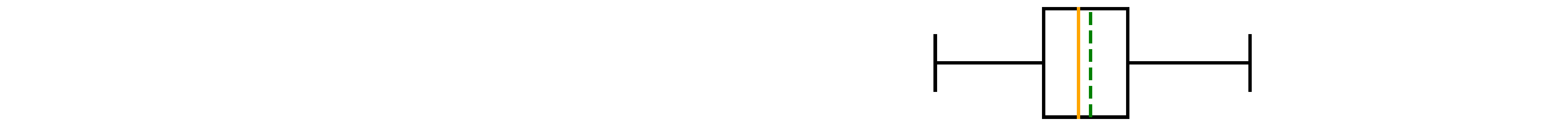} & \includegraphics[width=0.25\textwidth]{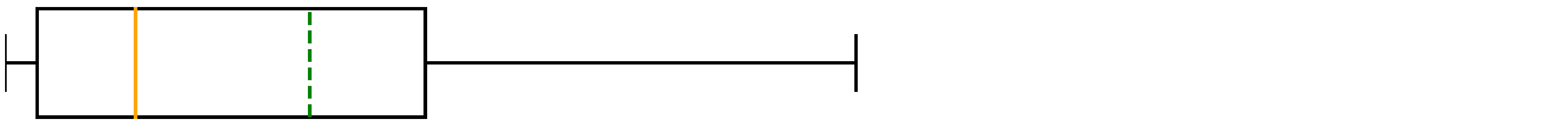} & \includegraphics[width=0.25\textwidth]{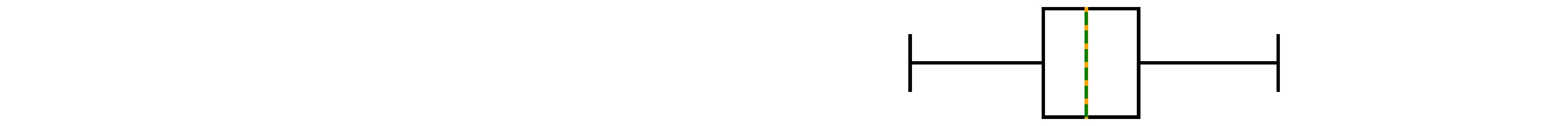}  & \includegraphics[width=0.25\textwidth]{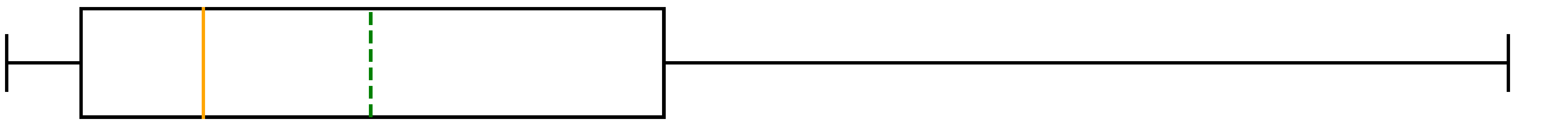} & 0.019 & 0.125 & 0.046 & 0.036 & 0.830 & 0.841 & 0.000 & 0.000 \\
 & & \resizebox{3.0mm}{3.0mm}{\myforecasting}\hspace{0.5mm}SARIMA & \includegraphics[width=0.25\textwidth]{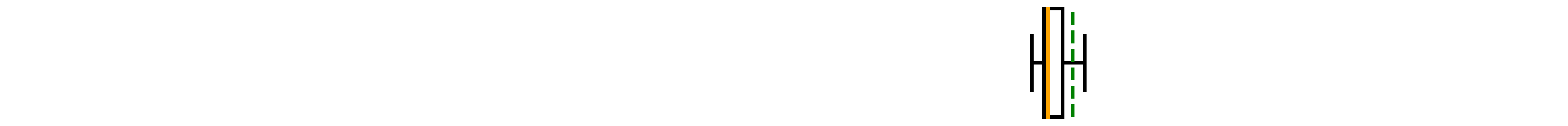} & \includegraphics[width=0.25\textwidth]{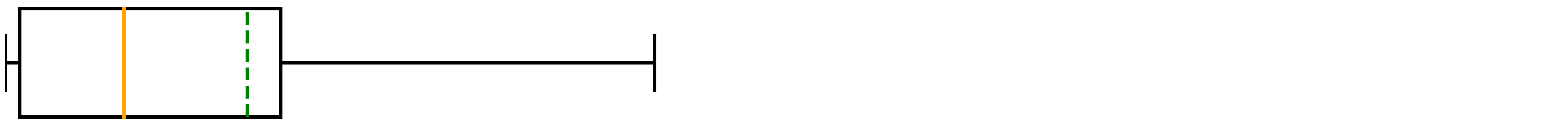} & \includegraphics[width=0.25\textwidth]{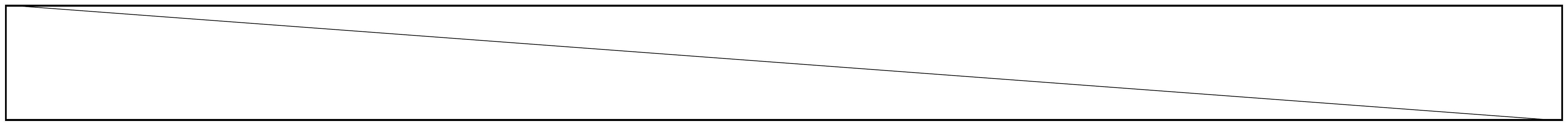}  & \includegraphics[width=0.25\textwidth]{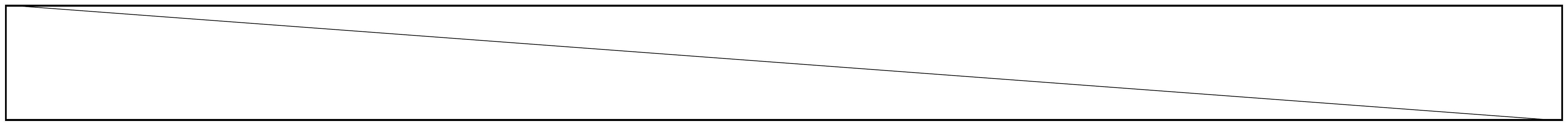} & 44.489 & \includegraphics[width=0.06\textwidth]{figures/main-table/no-data.pdf} & 1.929 & \includegraphics[width=0.06\textwidth]{figures/main-table/no-data.pdf} & 0.534 & \includegraphics[width=0.06\textwidth]{figures/main-table/no-data.pdf} & 0.000 & \includegraphics[width=0.06\textwidth]{figures/main-table/no-data.pdf} \\
 & & \resizebox{3.0mm}{3.0mm}{\mydistribution}\hspace{0.5mm}HBOS & \includegraphics[width=0.25\textwidth]{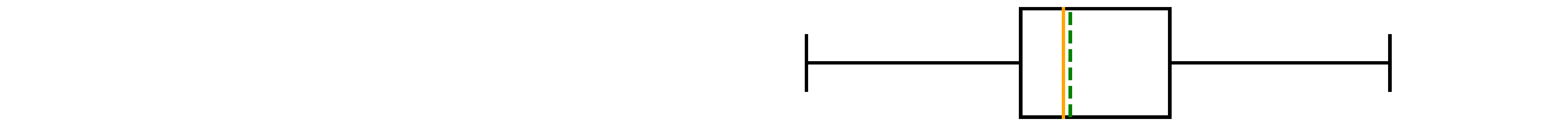} & \includegraphics[width=0.25\textwidth]{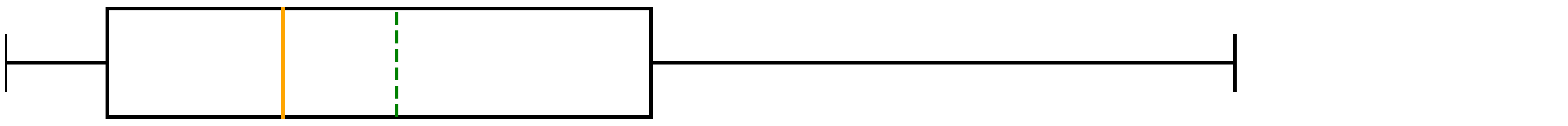} & \includegraphics[width=0.25\textwidth]{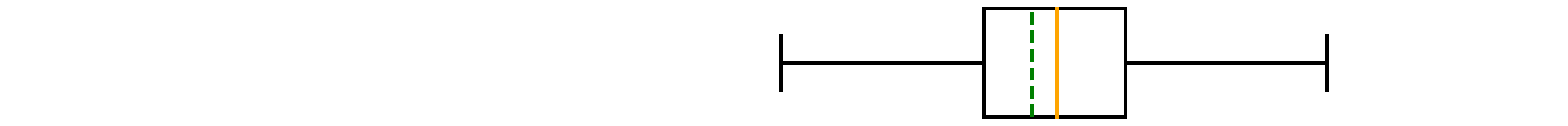}  & \includegraphics[width=0.25\textwidth]{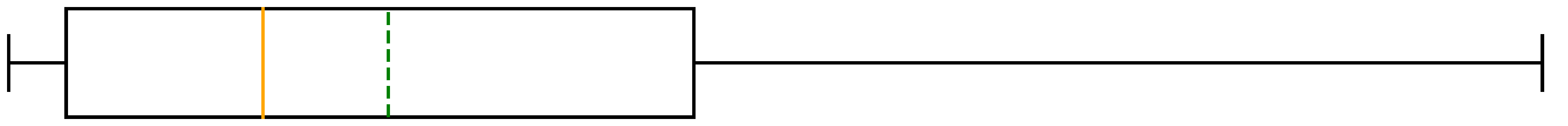} & 3.526 & 3.580 & 0.007 & 0.009 & 0.853 & 0.864 & 0.000 & 0.000 \\
 & & \resizebox{3.0mm}{3.0mm}{\myclustering}\hspace{0.5mm}CBLOF & \includegraphics[width=0.25\textwidth]{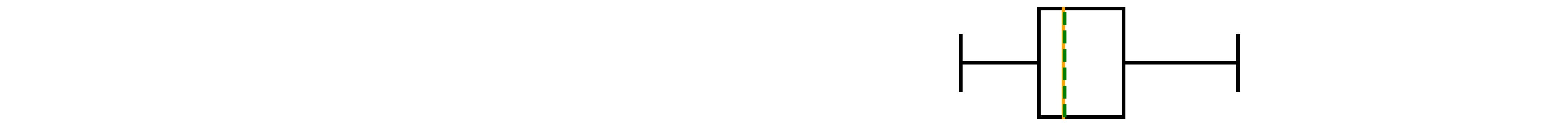} & \includegraphics[width=0.25\textwidth]{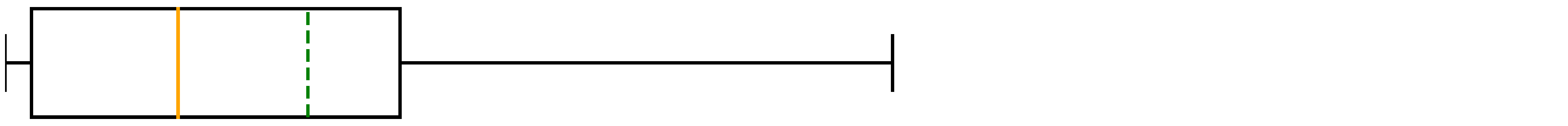} & \includegraphics[width=0.25\textwidth]{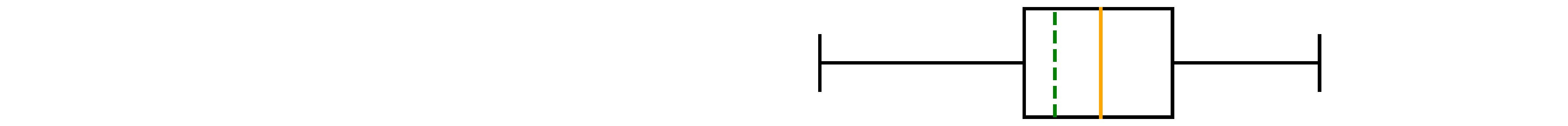}  & \includegraphics[width=0.25\textwidth]{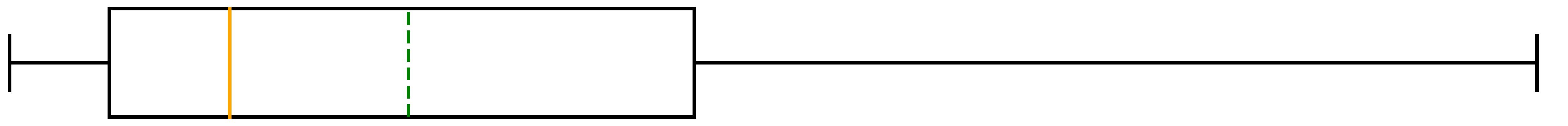} & 3.325 & 4.432 & 0.010 & 0.018 & 0.849 & 0.860 & 0.000 & 0.000 \\
 & & \resizebox{3.0mm}{3.0mm}{\mydiscord}\hspace{0.5mm}ZMS & \includegraphics[width=0.25\textwidth]{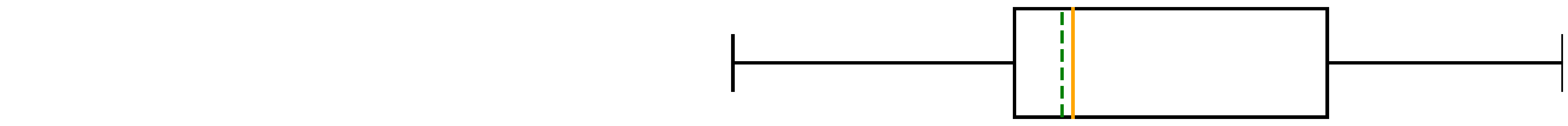} & \includegraphics[width=0.25\textwidth]{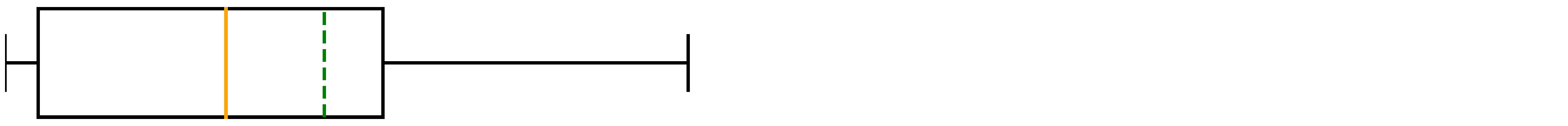} & \includegraphics[width=0.25\textwidth]{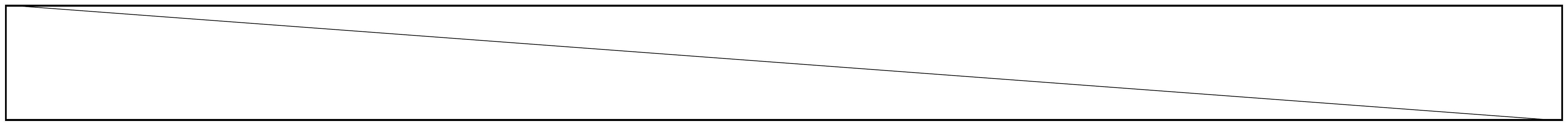}  & \includegraphics[width=0.25\textwidth]{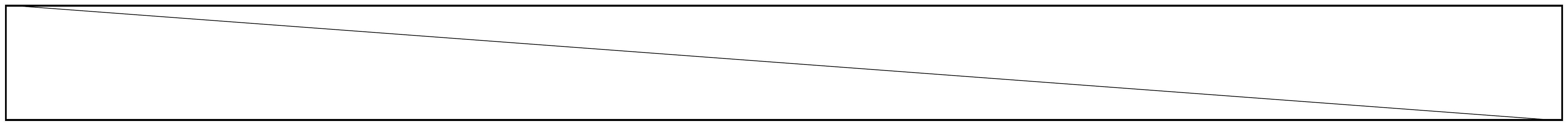} & 0.110 & \includegraphics[width=0.06\textwidth]{figures/main-table/no-data.pdf} & 0.063 & \includegraphics[width=0.06\textwidth]{figures/main-table/no-data.pdf} & 0.479 & \includegraphics[width=0.06\textwidth]{figures/main-table/no-data.pdf} & 0.000 & \includegraphics[width=0.06\textwidth]{figures/main-table/no-data.pdf} \\
 & & \resizebox{3.0mm}{3.0mm}{\myforecasting}\hspace{0.5mm}ARIMA & \includegraphics[width=0.25\textwidth]{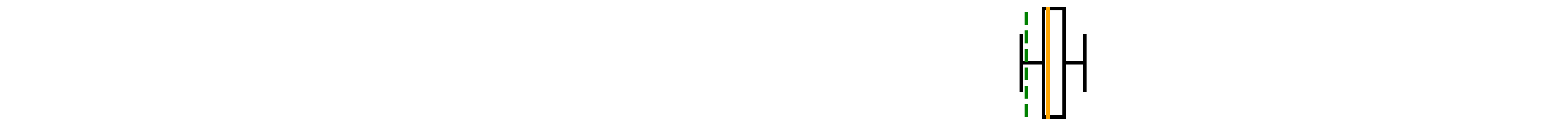} & \includegraphics[width=0.25\textwidth]{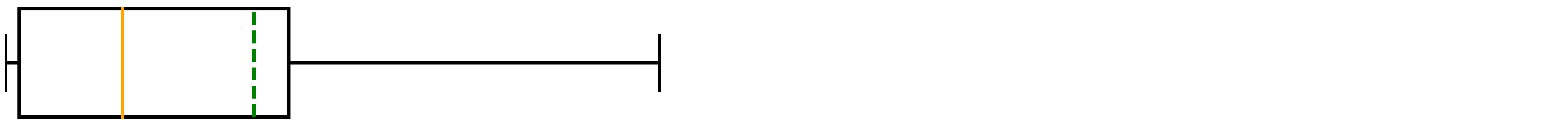} & \includegraphics[width=0.25\textwidth]{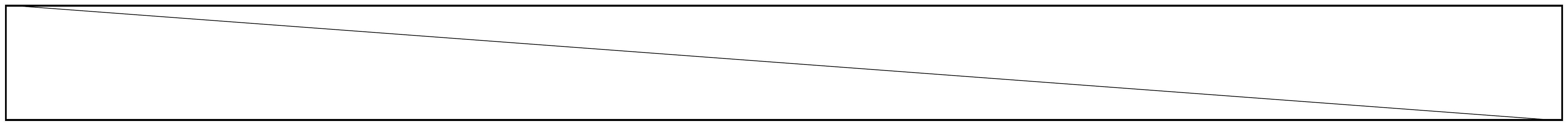}  & \includegraphics[width=0.25\textwidth]{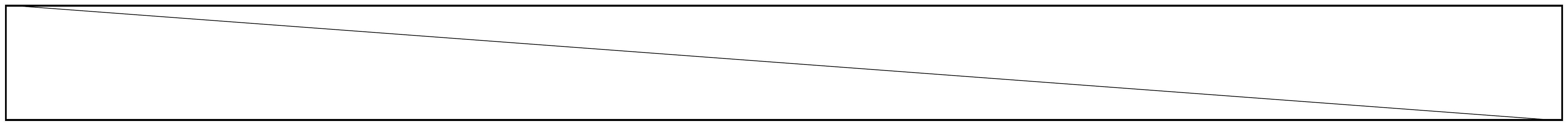} & 0.993 & \includegraphics[width=0.06\textwidth]{figures/main-table/no-data.pdf} & 0.710 & \includegraphics[width=0.06\textwidth]{figures/main-table/no-data.pdf} & 0.484 & \includegraphics[width=0.06\textwidth]{figures/main-table/no-data.pdf} & 0.000 & \includegraphics[width=0.06\textwidth]{figures/main-table/no-data.pdf} \\
 & & \resizebox{3.0mm}{3.0mm}{\mydistribution}\hspace{0.5mm}Stat & \includegraphics[width=0.25\textwidth]{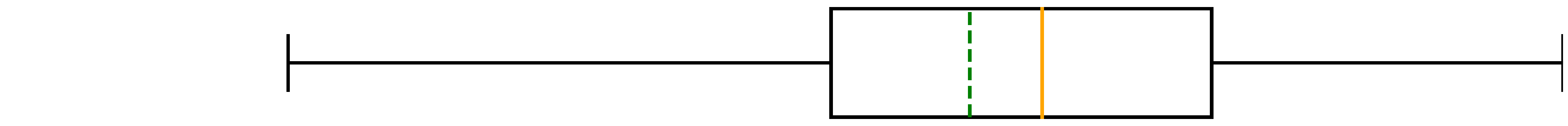} & \includegraphics[width=0.25\textwidth]{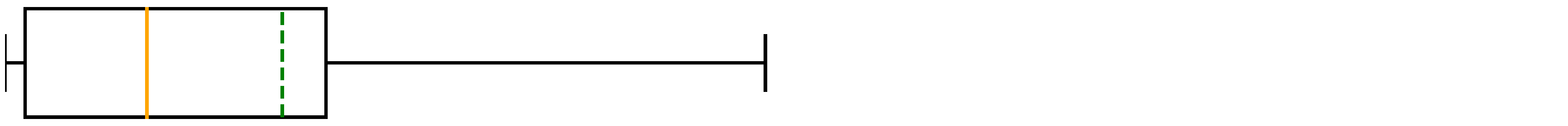} & \includegraphics[width=0.25\textwidth]{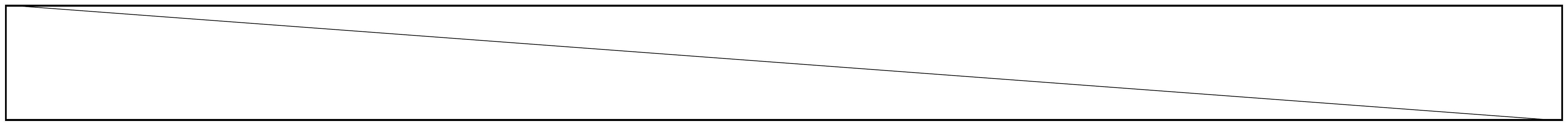}  & \includegraphics[width=0.25\textwidth]{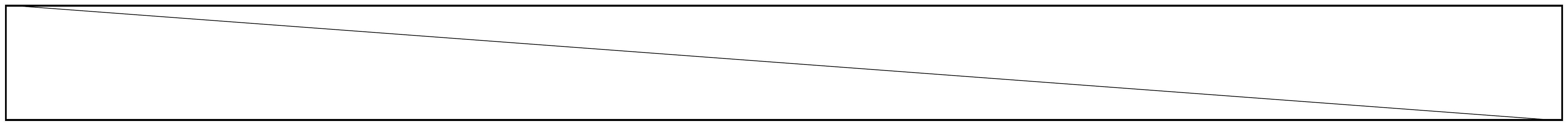} & 0.045 & \includegraphics[width=0.06\textwidth]{figures/main-table/no-data.pdf} & 0.015 & \includegraphics[width=0.06\textwidth]{figures/main-table/no-data.pdf} & 0.479 & \includegraphics[width=0.06\textwidth]{figures/main-table/no-data.pdf} & 0.000 & \includegraphics[width=0.06\textwidth]{figures/main-table/no-data.pdf} \\
\cmidrule(lr){2-15}
& \multirow{9}{*}{\rotatebox{90}{\textbf{Machine Learning}}} &
\resizebox{3.0mm}{3.0mm}{\mytree}\hspace{0.5mm}EIF & \includegraphics[width=0.25\textwidth]{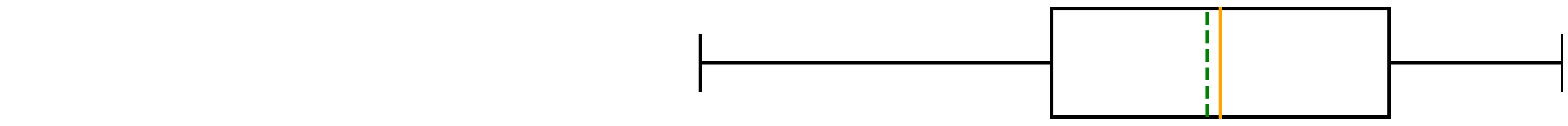} & \includegraphics[width=0.25\textwidth]{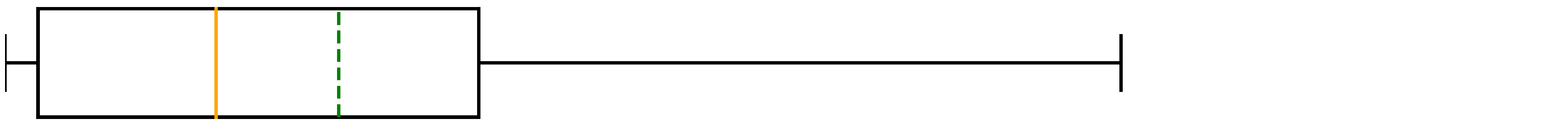} & \includegraphics[width=0.25\textwidth]{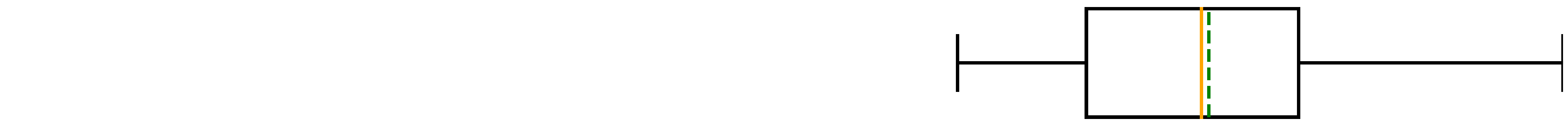}  & \includegraphics[width=0.25\textwidth]{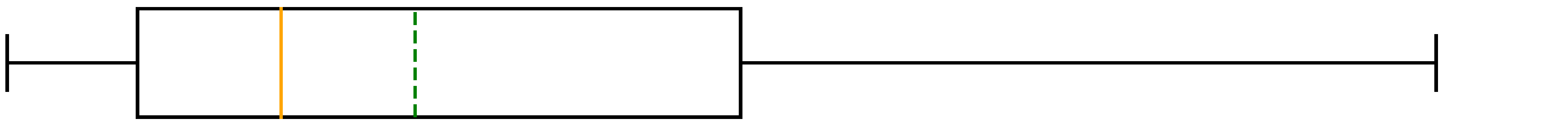} & 0.000 & 23.730 & 52.428 & 8.231 & 0.727 & 0.743 & 0.000 & 0.000 \\
 & & \resizebox{3.0mm}{3.0mm}{\myclustering}\hspace{0.5mm}KMeans & \includegraphics[width=0.25\textwidth]{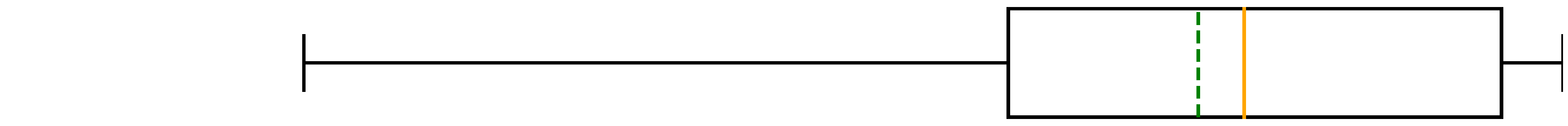} & \includegraphics[width=0.25\textwidth]{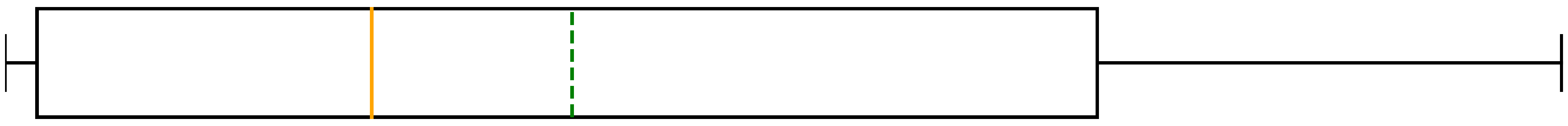} & \includegraphics[width=0.25\textwidth]{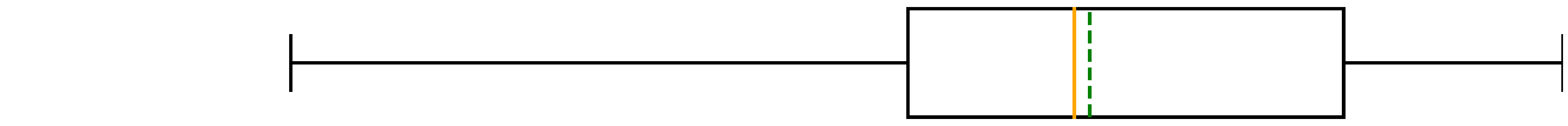}  & \includegraphics[width=0.25\textwidth]{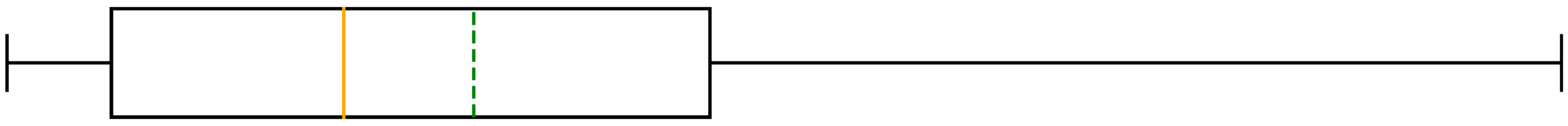} & 1.286 & 3.614 & 0.715 & 0.344 & 0.722 & 0.753 & 0.000 & 0.000 \\
 & & \resizebox{3.0mm}{3.0mm}{\mydistribution}\hspace{0.5mm}OCSVM & \includegraphics[width=0.25\textwidth]{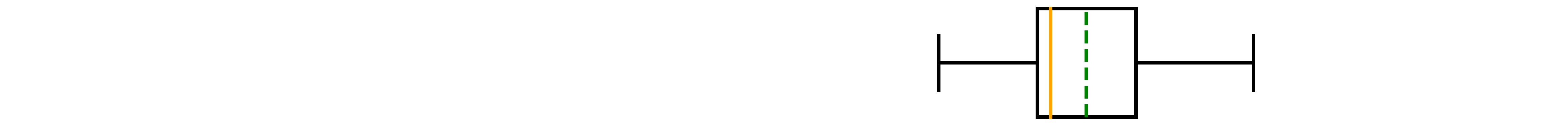} & \includegraphics[width=0.25\textwidth]{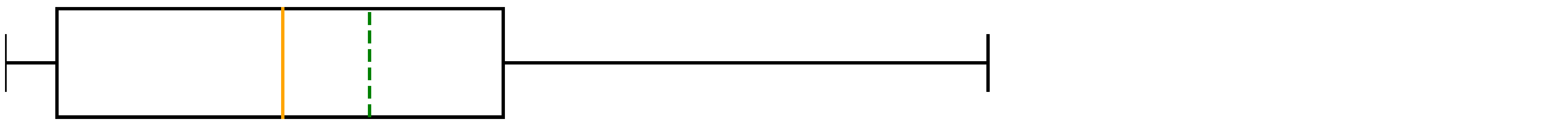} & \includegraphics[width=0.25\textwidth]{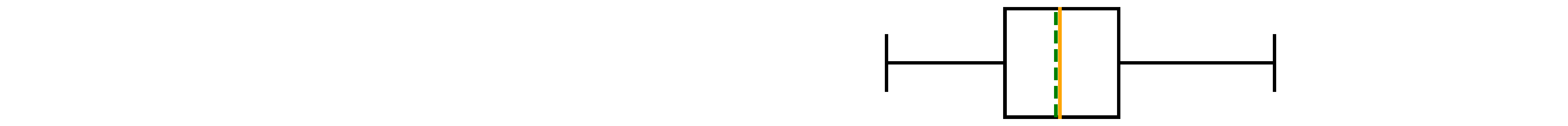}  & \includegraphics[width=0.25\textwidth]{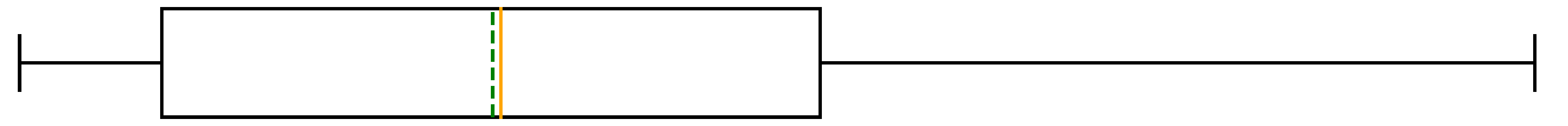} & 0.062 & 52.483 & 0.228 & 4.230 & 0.828 & 1.014 & 0.000 & 0.000 \\
 & & \resizebox{3.0mm}{3.0mm}{\myproximity}\hspace{0.5mm}KNN & \includegraphics[width=0.25\textwidth]{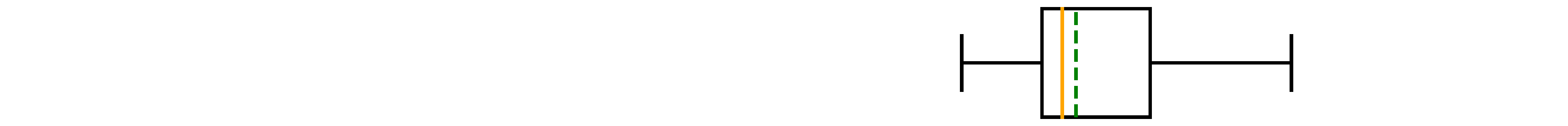} & \includegraphics[width=0.25\textwidth]{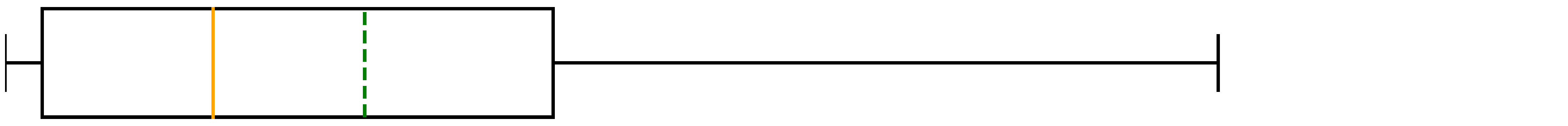} & \includegraphics[width=0.25\textwidth]{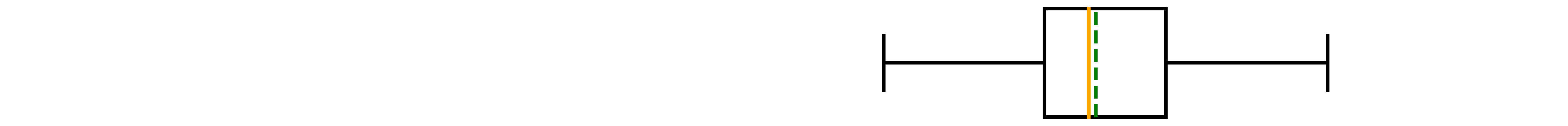}  & \includegraphics[width=0.25\textwidth]{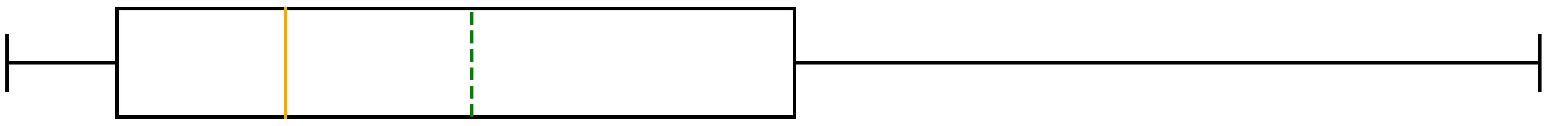} & 0.017 & 0.077 & 1.151 & 0.678 & 0.827 & 0.834 & 0.000 & 0.000 \\
 & & \resizebox{3.0mm}{3.0mm}{\mytree}\hspace{0.5mm}IF & \includegraphics[width=0.25\textwidth]{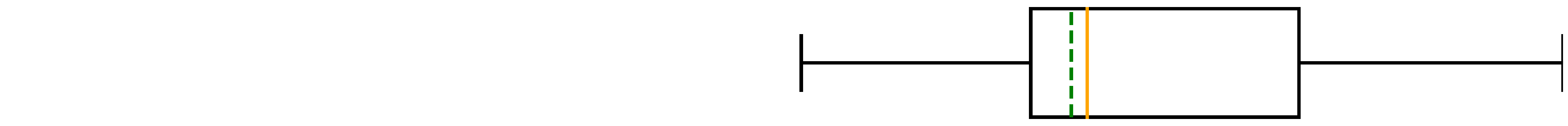} & \includegraphics[width=0.25\textwidth]{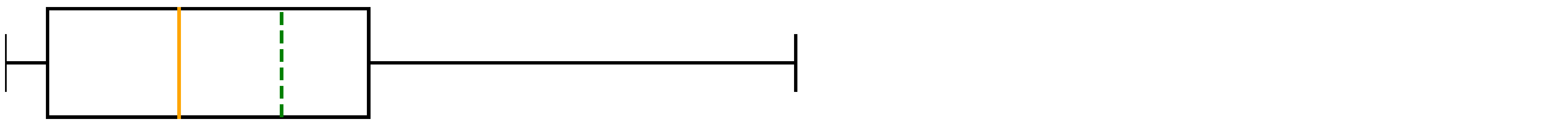} & \includegraphics[width=0.25\textwidth]{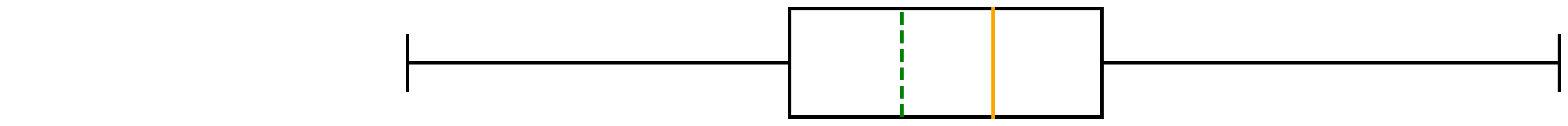}  & \includegraphics[width=0.25\textwidth]{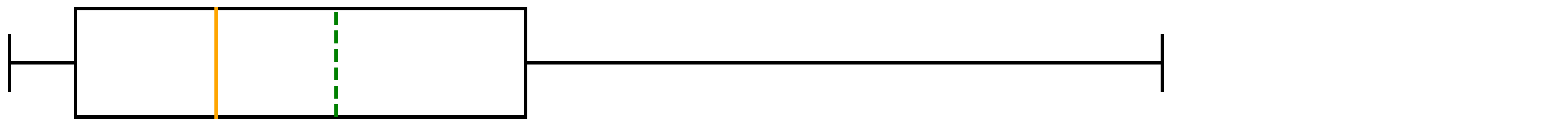} & 0.556 & 0.222 & 0.080 & 0.038 & 0.481 & 0.490 & 0.000 & 0.000 \\
 & & \resizebox{3.0mm}{3.0mm}{\myreconstruction}\hspace{0.5mm}DP & \includegraphics[width=0.25\textwidth]{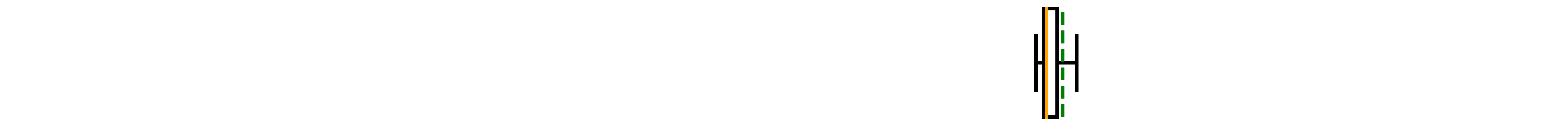} & \includegraphics[width=0.25\textwidth]{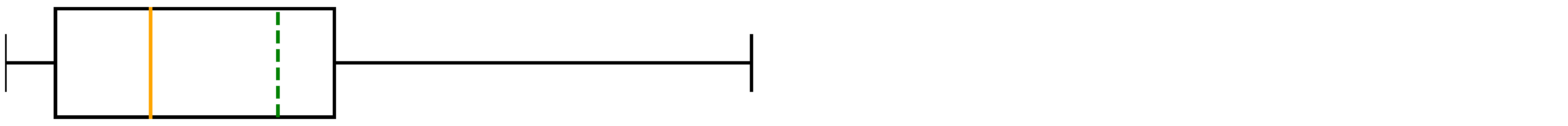} & \includegraphics[width=0.25\textwidth]{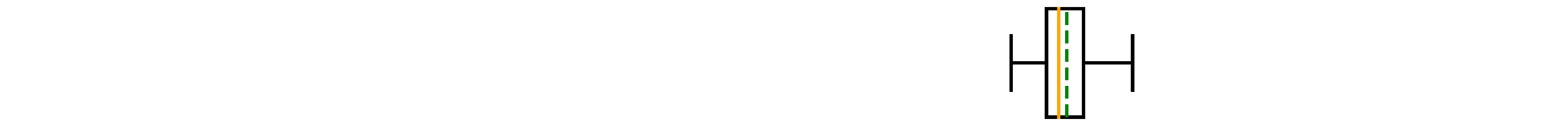}  & \includegraphics[width=0.25\textwidth]{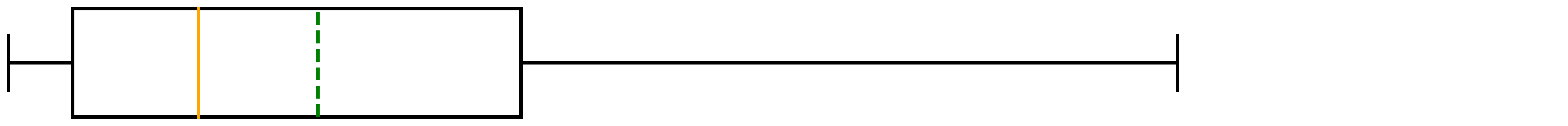} & 1.061 & 6.222 & 3.238 & 1.640 & 0.689 & 0.724 & 0.154 & 0.275 \\
 & & \resizebox{3.0mm}{3.0mm}{\myreconstruction}\hspace{0.5mm}SR & \includegraphics[width=0.25\textwidth]{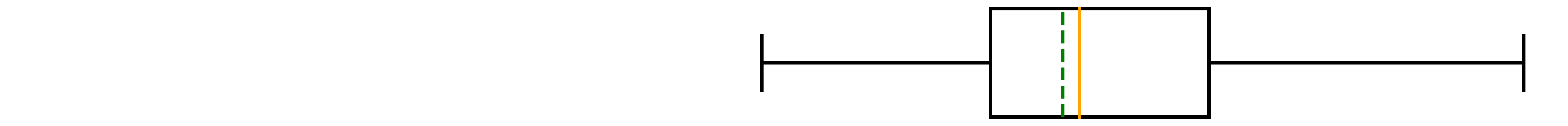} & \includegraphics[width=0.25\textwidth]{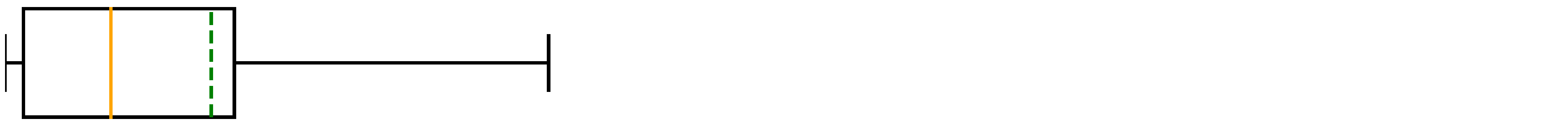} & \includegraphics[width=0.25\textwidth]{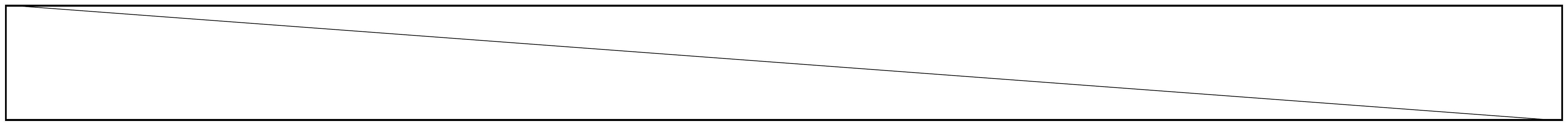}  & \includegraphics[width=0.25\textwidth]{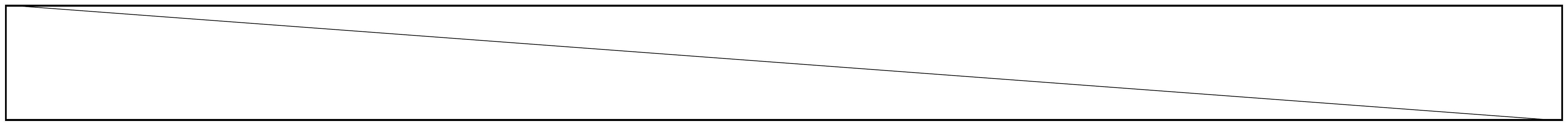} & 0.056 & \includegraphics[width=0.06\textwidth]{figures/main-table/no-data.pdf} & 0.028 & \includegraphics[width=0.06\textwidth]{figures/main-table/no-data.pdf} & 0.480 & \includegraphics[width=0.06\textwidth]{figures/main-table/no-data.pdf} & 0.000 & \includegraphics[width=0.06\textwidth]{figures/main-table/no-data.pdf} \\
 & & \resizebox{3.0mm}{3.0mm}{\myencoding}\hspace{0.5mm}PCA & \includegraphics[width=0.25\textwidth]{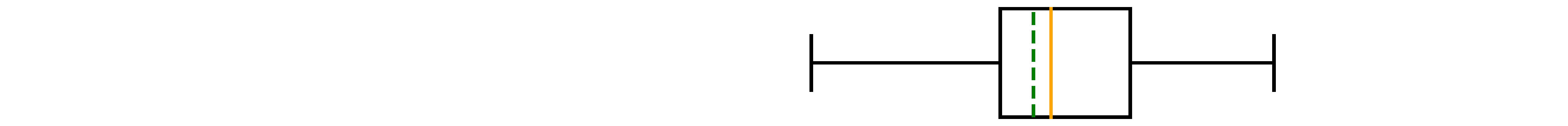} & \includegraphics[width=0.25\textwidth]{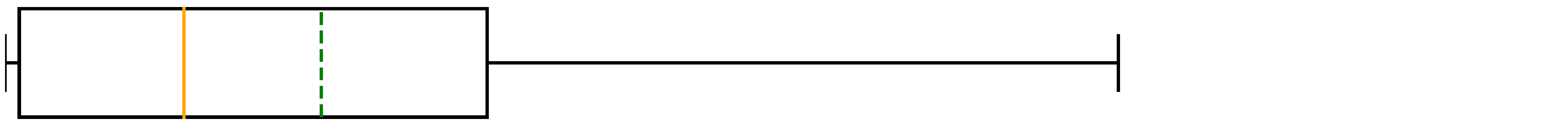} & \includegraphics[width=0.25\textwidth]{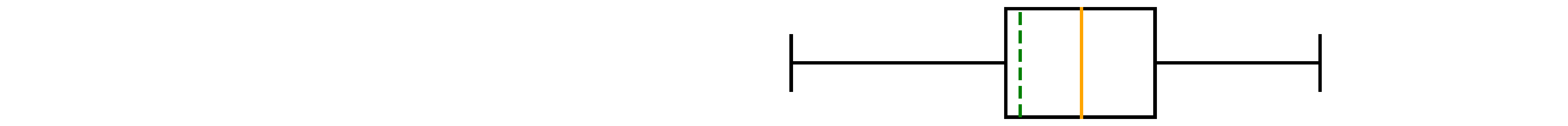}  & \includegraphics[width=0.25\textwidth]{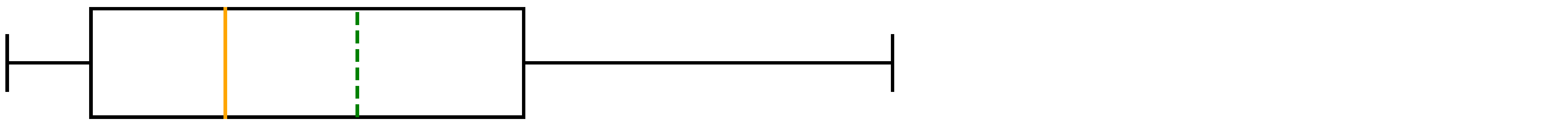} & 0.020 & 0.127 & 0.044 & 0.031 & 0.829 & 0.836 & 0.000 & 0.000 \\
 & & \resizebox{3.0mm}{3.0mm}{\mydistribution}\hspace{0.5mm}LODA & \includegraphics[width=0.25\textwidth]{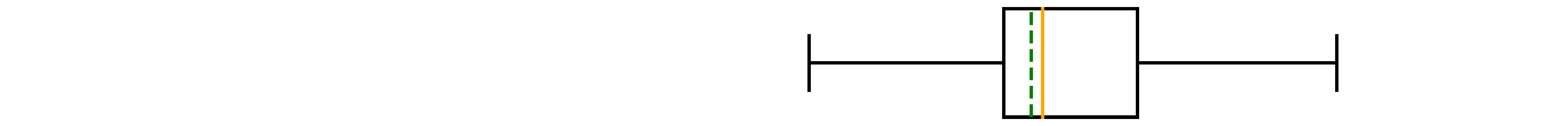} & \includegraphics[width=0.25\textwidth]{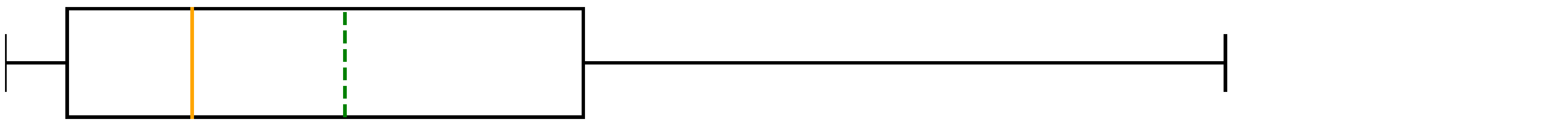} & \includegraphics[width=0.25\textwidth]{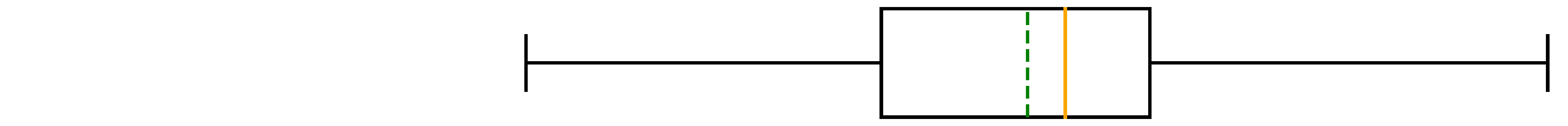}  & \includegraphics[width=0.25\textwidth]{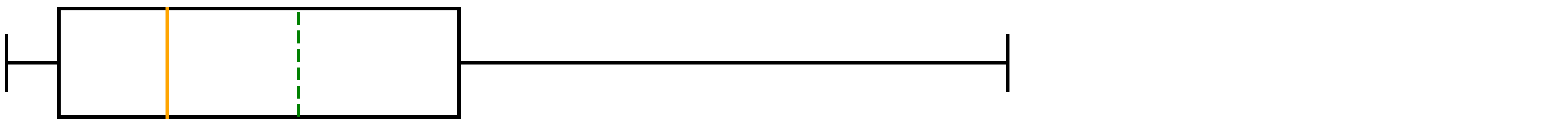} & 0.047 & 0.133 & 0.035 & 0.022 & 0.827 & 0.832 & 0.000 & 0.000 \\
\cmidrule(lr){2-15}
& \multirow{18}{*}{\rotatebox{90}{\textbf{Deep Learning}}} &
\resizebox{3.0mm}{3.0mm}{\myreconstruction}\hspace{0.5mm}iTrans & \includegraphics[width=0.25\textwidth]{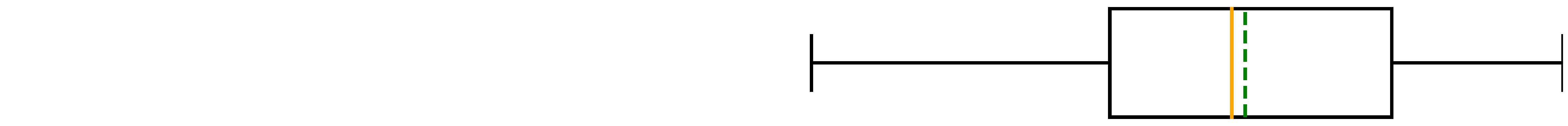} & \includegraphics[width=0.25\textwidth]{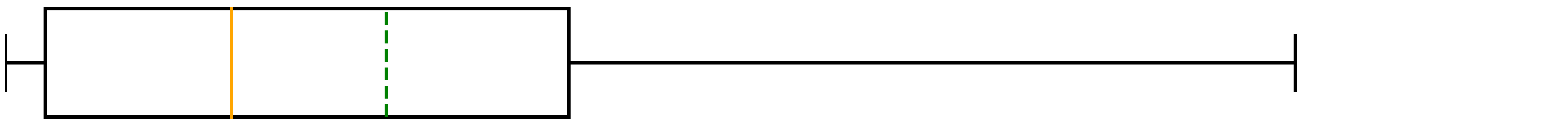} & \includegraphics[width=0.25\textwidth]{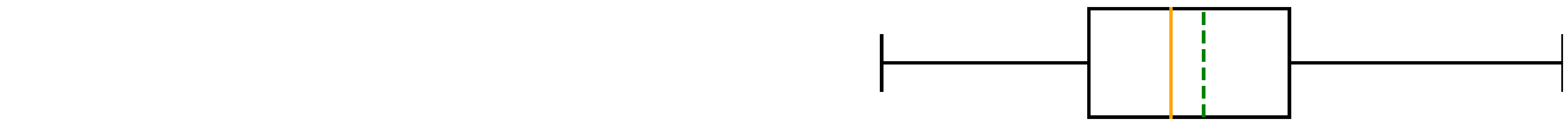}  & \includegraphics[width=0.25\textwidth]{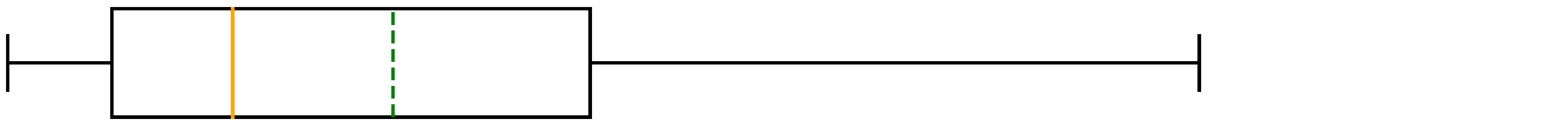} & 4.476 & 39.587 & 0.267 & 0.025 & 0.747 & 0.807 & 0.219 & 0.273 \\
 & & \resizebox{3.0mm}{3.0mm}{\myreconstruction}\hspace{0.5mm}DLin & \includegraphics[width=0.25\textwidth]{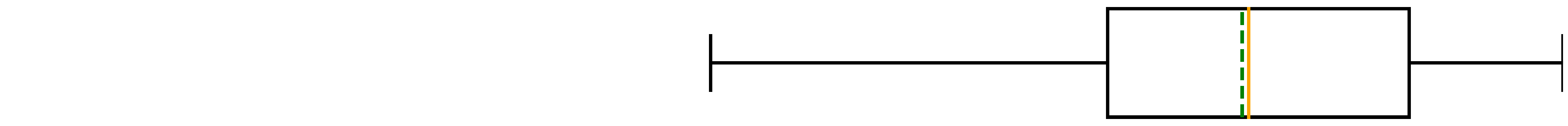} & \includegraphics[width=0.25\textwidth]{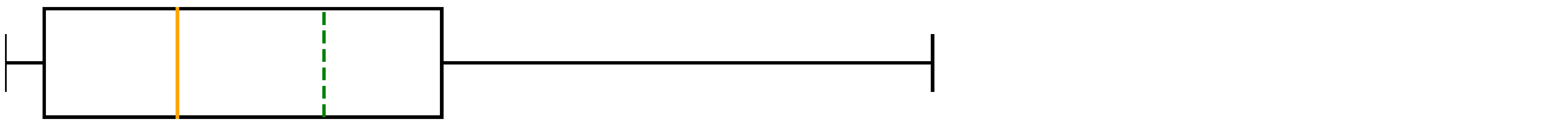} & \includegraphics[width=0.25\textwidth]{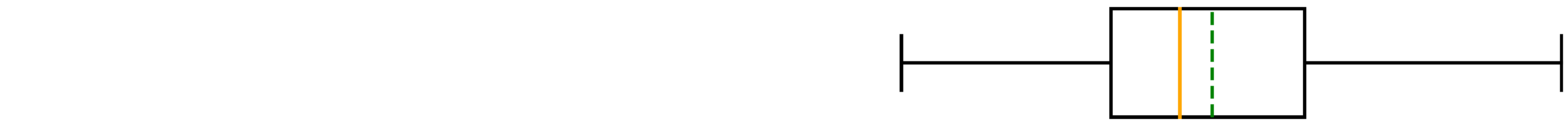}  & \includegraphics[width=0.25\textwidth]{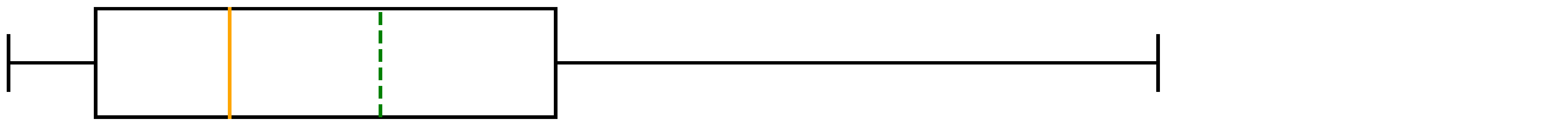} & 0.597 & 22.487 & 0.013 & 0.011 & 0.572 & 0.665 & 0.007 & 0.133 \\
 & & \resizebox{3.0mm}{3.0mm}{\myreconstruction}\hspace{0.5mm}TsNet & \includegraphics[width=0.25\textwidth]{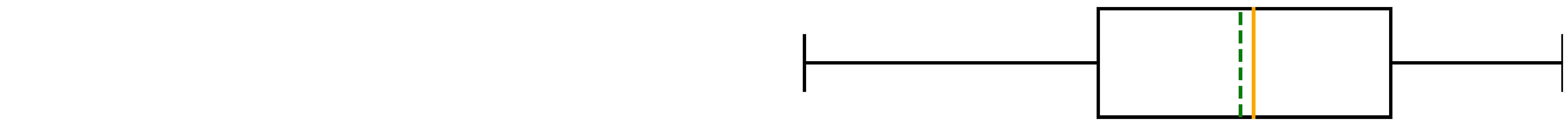} & \includegraphics[width=0.25\textwidth]{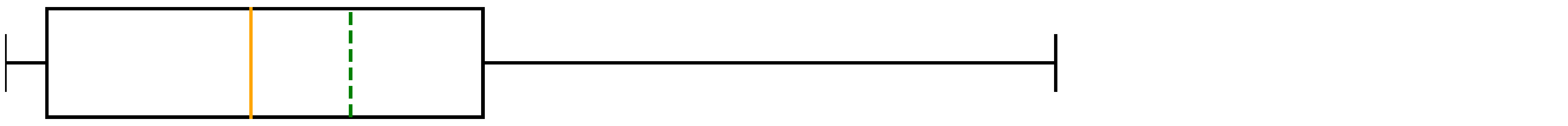} & \includegraphics[width=0.25\textwidth]{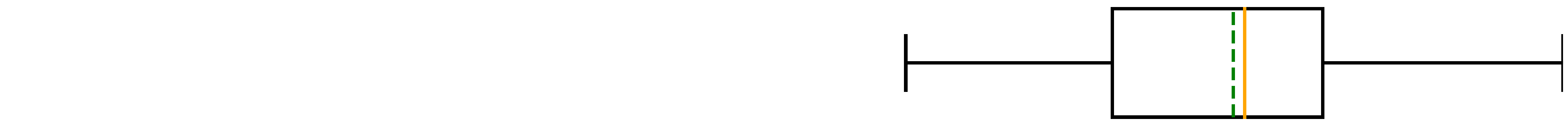}  & \includegraphics[width=0.25\textwidth]{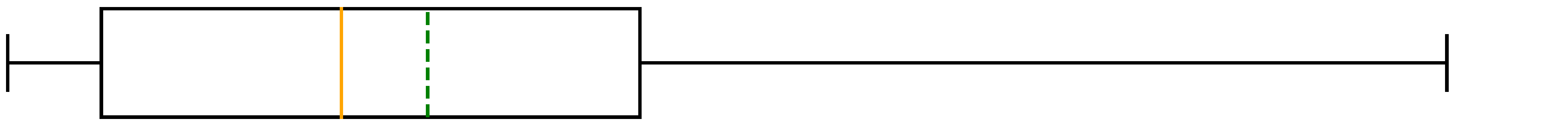} & 4.089 & 6.343 & 0.252 & 0.126 & 0.918 & 0.989 & 0.426 & 0.505 \\
 & & \resizebox{3.0mm}{3.0mm}{\myreconstruction}\hspace{0.5mm}Modern & \includegraphics[width=0.25\textwidth]{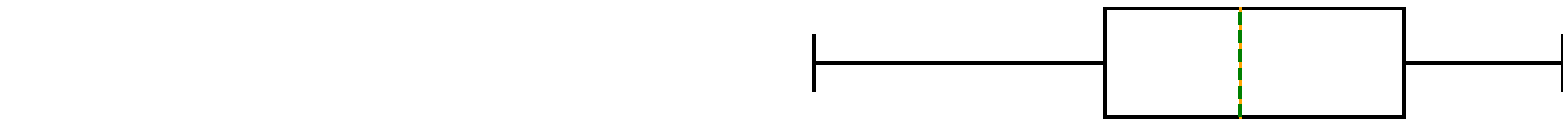} & \includegraphics[width=0.25\textwidth]{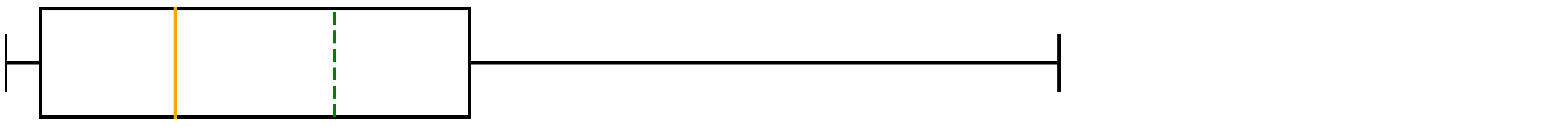} & \includegraphics[width=0.25\textwidth]{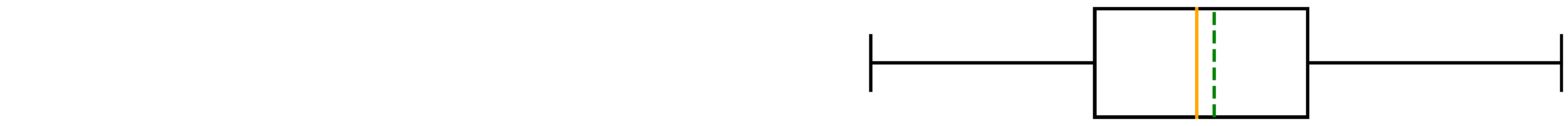}  & \includegraphics[width=0.25\textwidth]{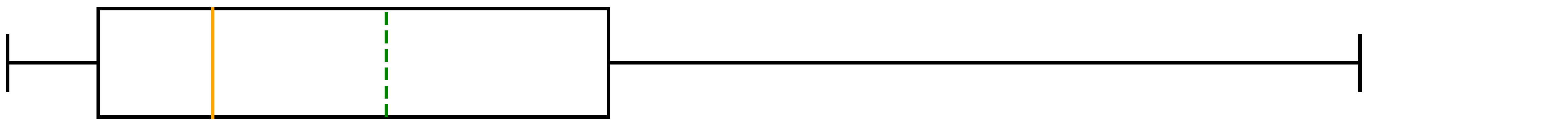} & 2.705 & 3.892 & 0.218 & 0.017 & 0.947 & 0.981 & 0.235 & 0.130 \\
 & & \resizebox{3.0mm}{3.0mm}{\myreconstruction}\hspace{0.5mm}Patch & \includegraphics[width=0.25\textwidth]{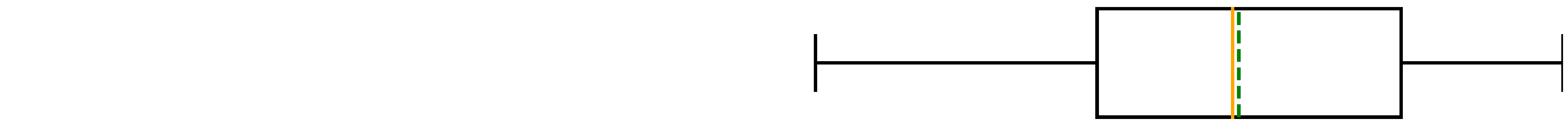} & \includegraphics[width=0.25\textwidth]{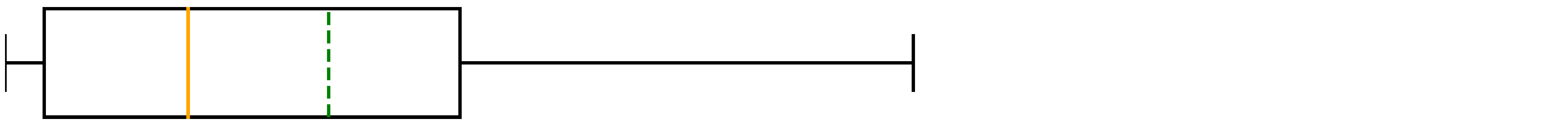} & \includegraphics[width=0.25\textwidth]{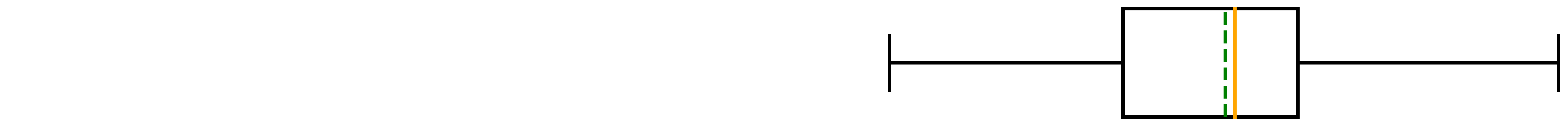}  & \includegraphics[width=0.25\textwidth]{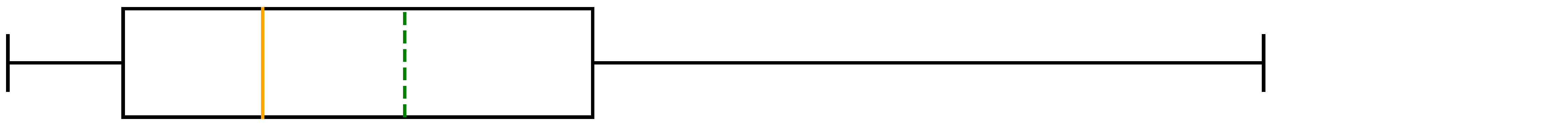} & 1.449 & 12.001 & 0.027 & 0.027 & 0.675 & 0.800 & 0.206 & 0.523 \\
 & & \resizebox{3.0mm}{3.0mm}{\myreconstruction}\hspace{0.5mm}NLin & \includegraphics[width=0.25\textwidth]{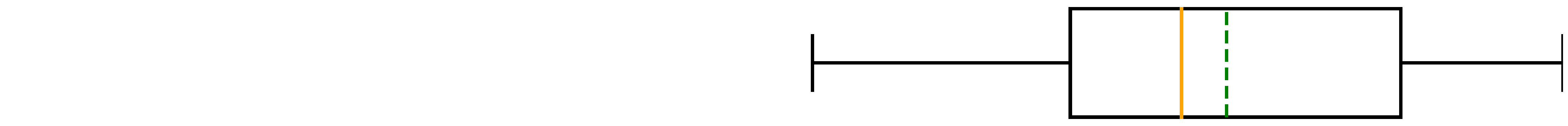} & \includegraphics[width=0.25\textwidth]{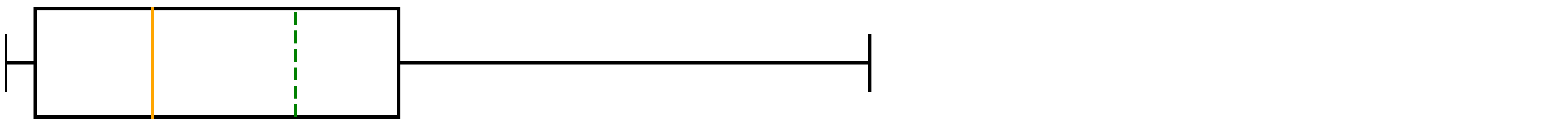} & \includegraphics[width=0.25\textwidth]{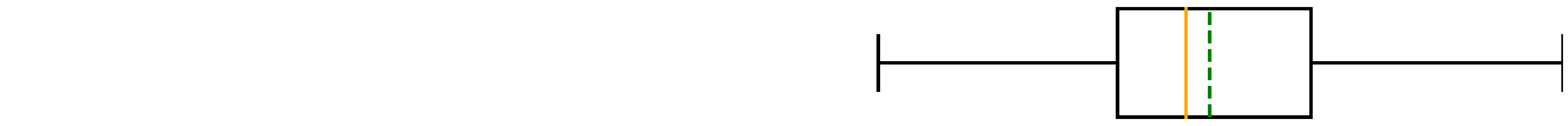}  & \includegraphics[width=0.25\textwidth]{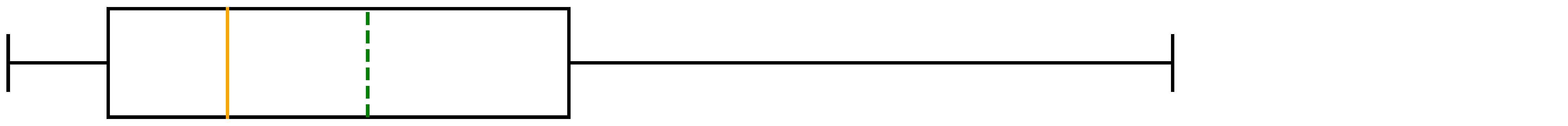} & 0.752 & 21.245 & 0.012 & 0.010 & 0.587 & 0.661 & 0.033 & 0.131 \\
 & & \resizebox{3.0mm}{3.0mm}{\myreconstruction}\hspace{0.5mm}TranAD & \includegraphics[width=0.25\textwidth]{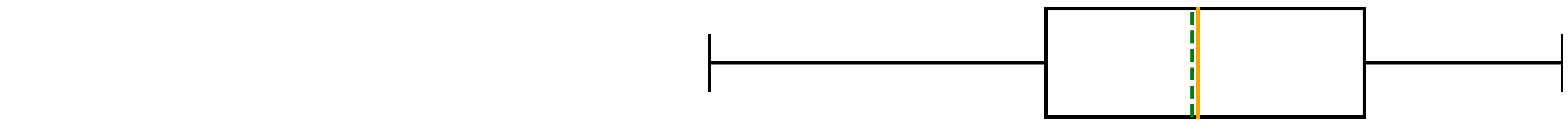} & \includegraphics[width=0.25\textwidth]{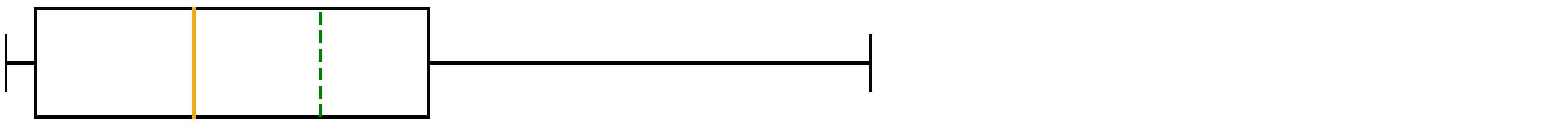} & \includegraphics[width=0.25\textwidth]{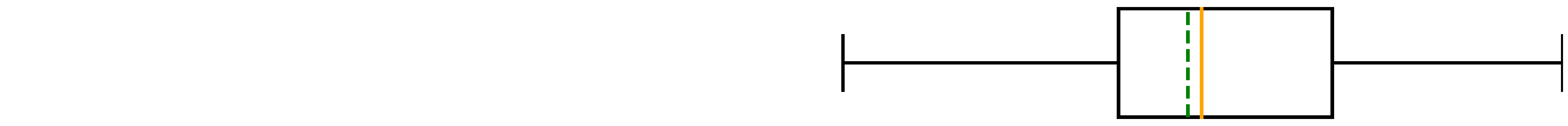}  & \includegraphics[width=0.25\textwidth]{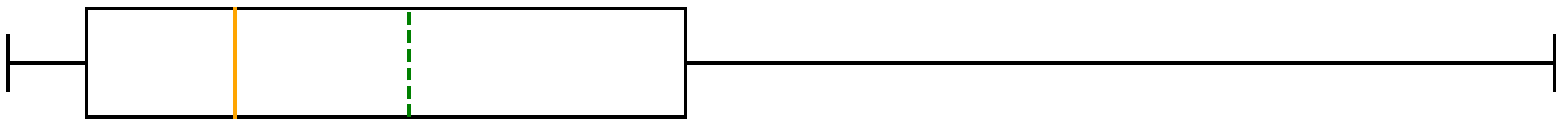} & 1.673 & 20.314 & 0.172 & 0.108 & 0.849 & 0.885 & 0.102 & 0.111 \\
 & & \resizebox{3.0mm}{3.0mm}{\myreconstruction}\hspace{0.5mm}DualTF & \includegraphics[width=0.25\textwidth]{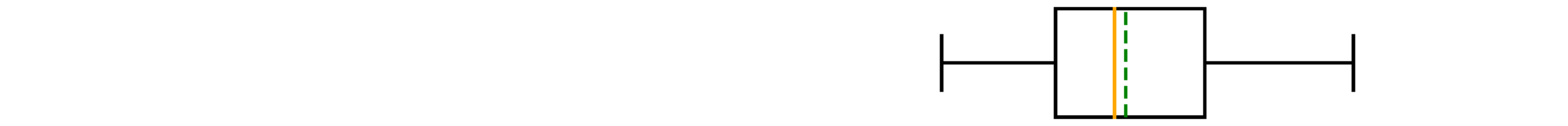} & \includegraphics[width=0.25\textwidth]{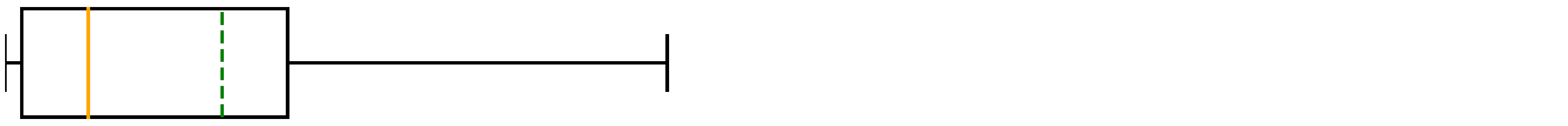} & \includegraphics[width=0.25\textwidth]{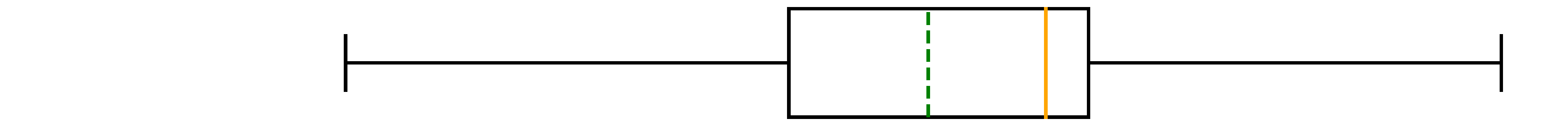}  & \includegraphics[width=0.25\textwidth]{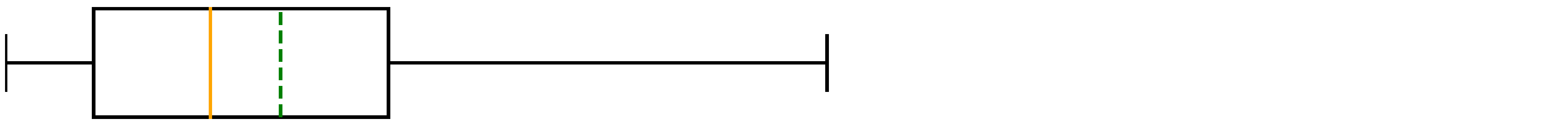} & 163.936 & 381.367 & 155.892 & 35.221 & 0.981 & 1.176 & 18.564 & 2.797 \\
 & & \resizebox{3.0mm}{3.0mm}{\myreconstruction}\hspace{0.5mm}ATrans & \includegraphics[width=0.25\textwidth]{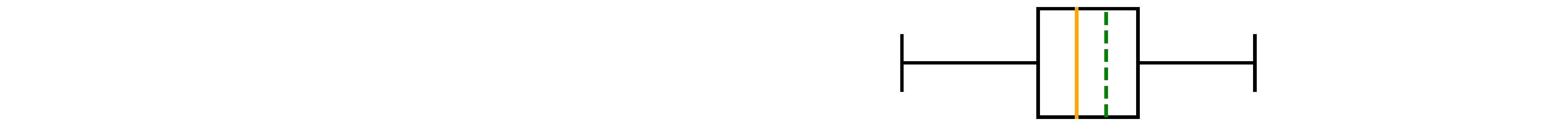} & \includegraphics[width=0.25\textwidth]{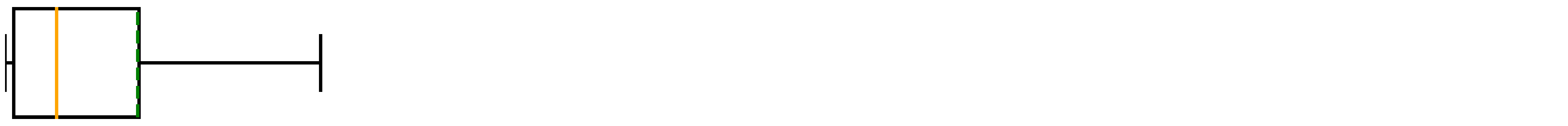} & \includegraphics[width=0.25\textwidth]{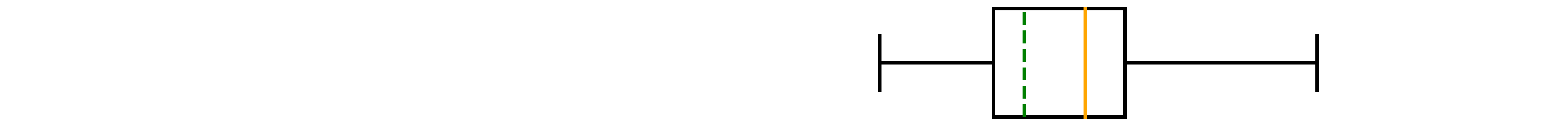}  & \includegraphics[width=0.25\textwidth]{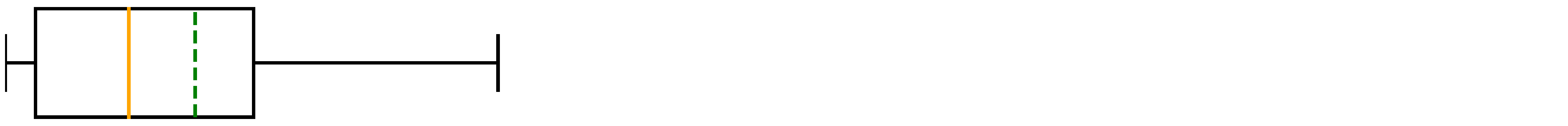} & 2.234 & 23.122 & 0.054 & 0.037 & 0.940 & 1.021 & 1.171 & 5.380 \\
 & & \resizebox{3.0mm}{3.0mm}{\mycontrast}\hspace{0.5mm}ConAD & \includegraphics[width=0.25\textwidth]{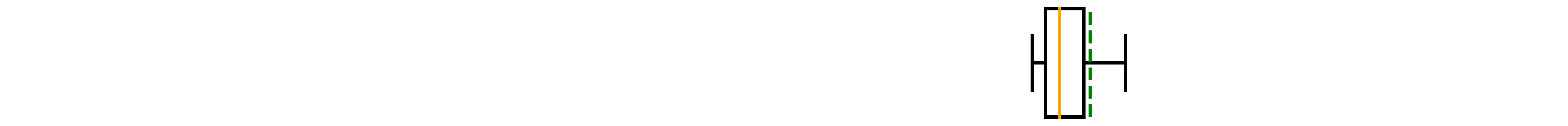} & \includegraphics[width=0.25\textwidth]{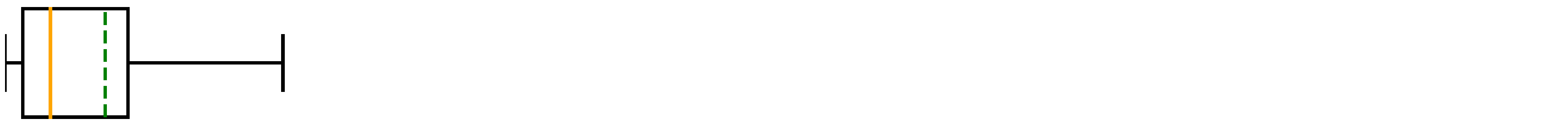} & \includegraphics[width=0.25\textwidth]{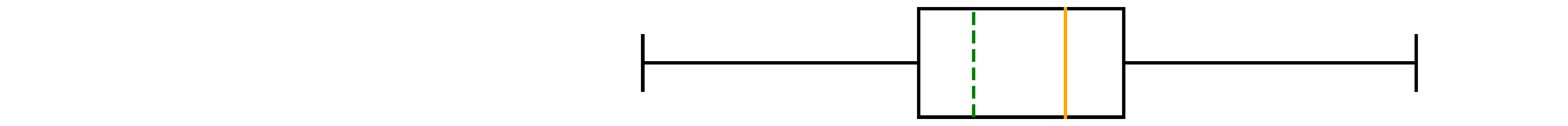}  & \includegraphics[width=0.25\textwidth]{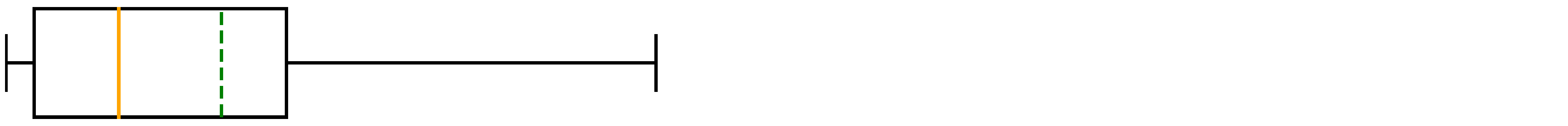} & 24.001 & 271.416 & 0.216 & 0.133 & 0.987 & 0.994 & 16.199 & 19.917 \\
 & & \resizebox{3.0mm}{3.0mm}{\myforecasting}\hspace{0.5mm}Torsk & \includegraphics[width=0.25\textwidth]{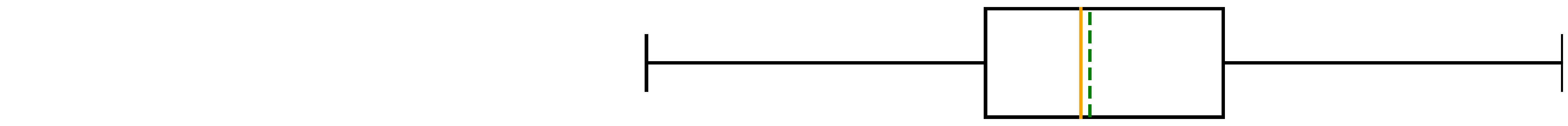} & \includegraphics[width=0.25\textwidth]{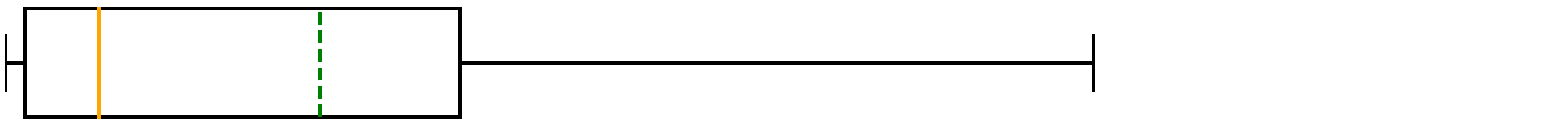} & \includegraphics[width=0.25\textwidth]{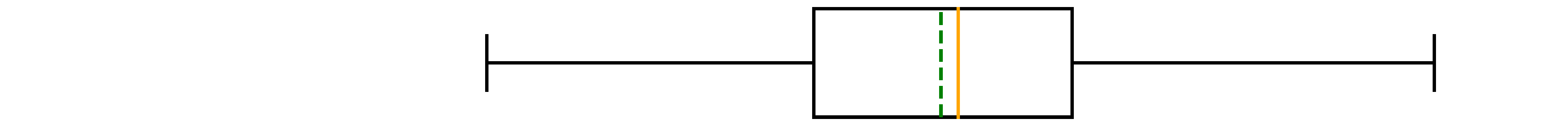}  & \includegraphics[width=0.25\textwidth]{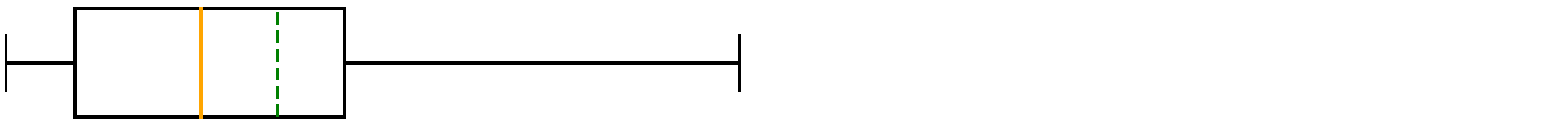} & 13.280 & 72.011 & 119.179 & 24.267 & 0.721 & 0.726 & 0.000 & 0.000 \\
 & & \resizebox{3.0mm}{3.0mm}{\mycontrast}\hspace{0.5mm}DC & \includegraphics[width=0.25\textwidth]{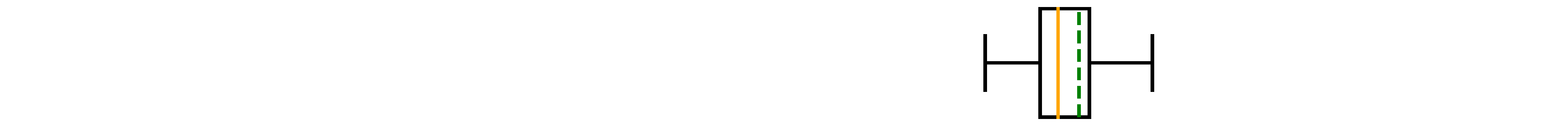} & \includegraphics[width=0.25\textwidth]{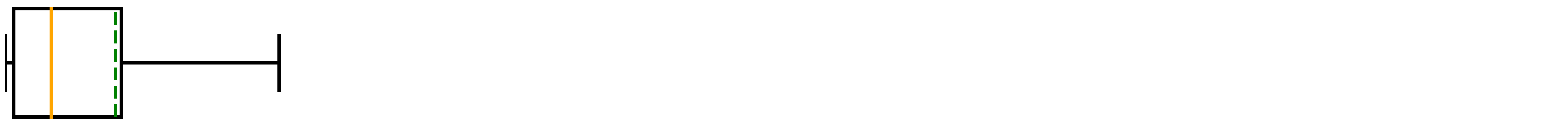} & \includegraphics[width=0.25\textwidth]{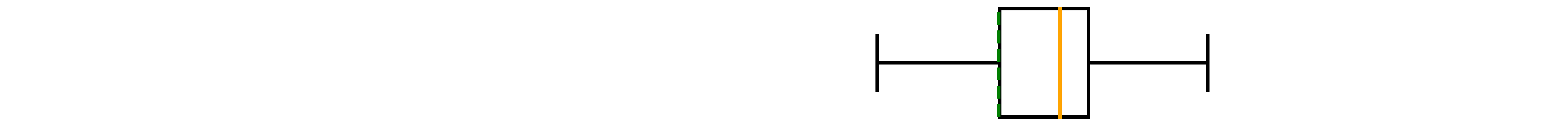}  & \includegraphics[width=0.25\textwidth]{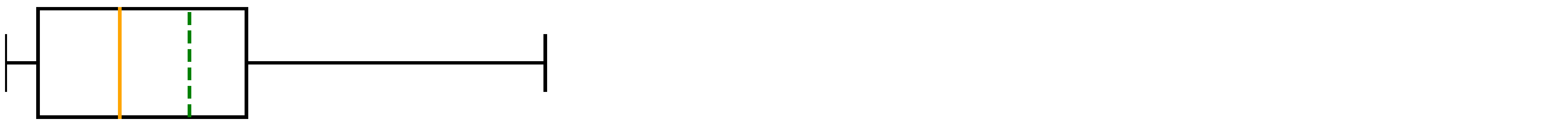} & 3.854 & 13.655 & 0.044 & 0.029 & 0.957 & 0.949 & 0.274 & 0.821 \\
 & & \resizebox{3.0mm}{3.0mm}{\myforecasting}\hspace{0.5mm}LSTM & \includegraphics[width=0.25\textwidth]{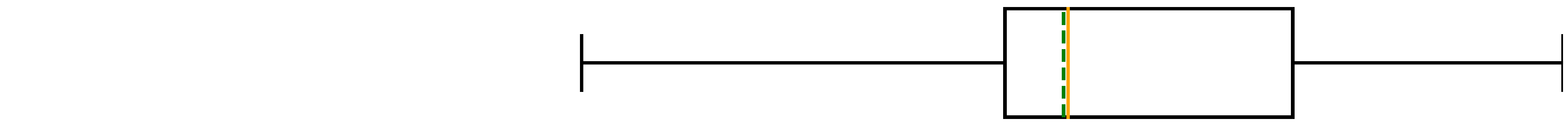} & \includegraphics[width=0.25\textwidth]{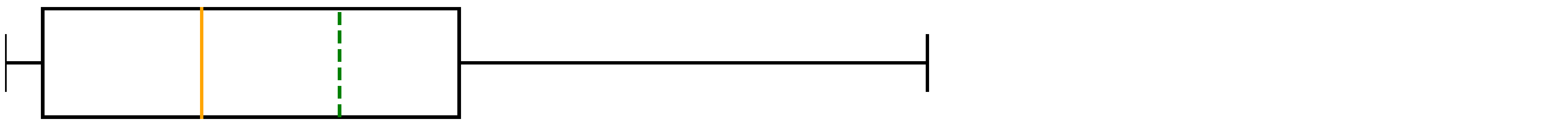} & \includegraphics[width=0.25\textwidth]{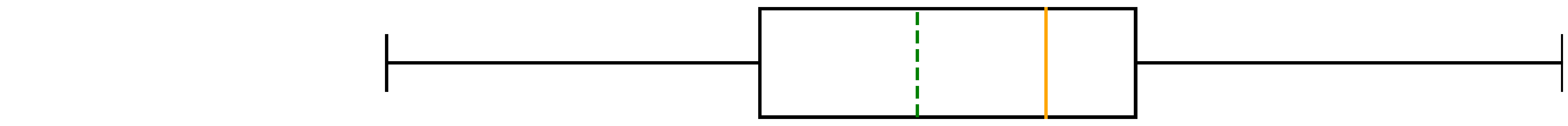}  & \includegraphics[width=0.25\textwidth]{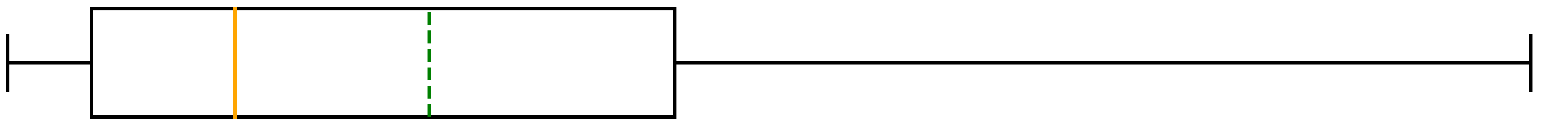} & 3.636 & 34.668 & 0.962 & 0.567 & 0.697 & 0.741 & 0.129 & 0.162 \\
 & & \resizebox{3.0mm}{3.0mm}{\myreconstruction}\hspace{0.5mm}AE & \includegraphics[width=0.25\textwidth]{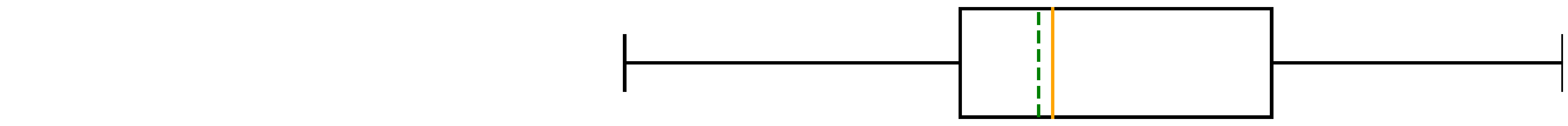} & \includegraphics[width=0.25\textwidth]{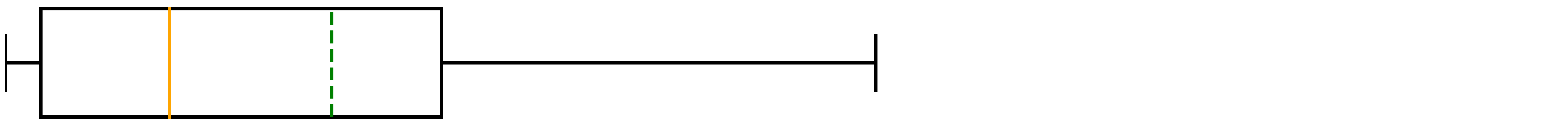} & \includegraphics[width=0.25\textwidth]{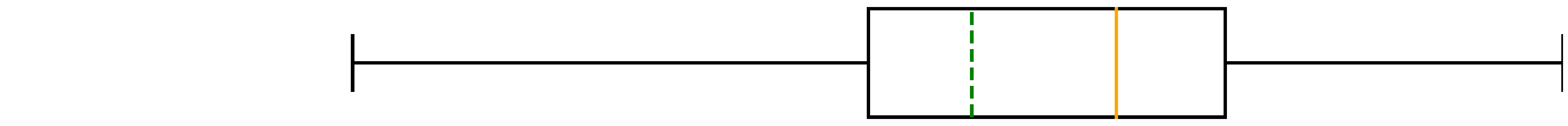}  & \includegraphics[width=0.25\textwidth]{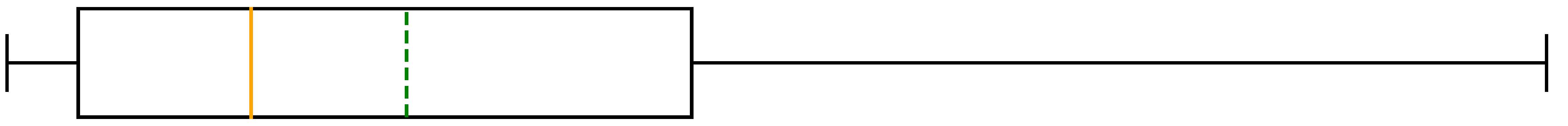} & 1.886 & 70.892 & 0.629 & 0.362 & 0.667 & 0.751 & 0.072 & 0.111 \\
 & & \resizebox{3.0mm}{3.0mm}{\myreconstruction}\hspace{0.5mm}VAE & \includegraphics[width=0.25\textwidth]{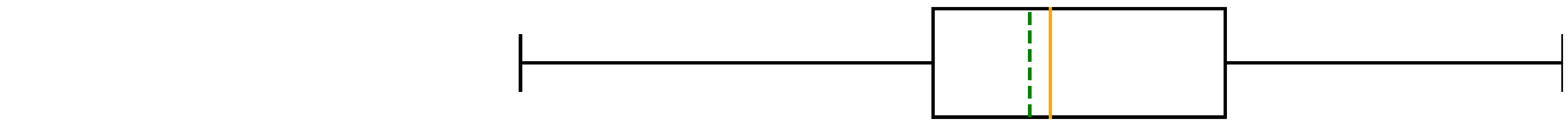} & \includegraphics[width=0.25\textwidth]{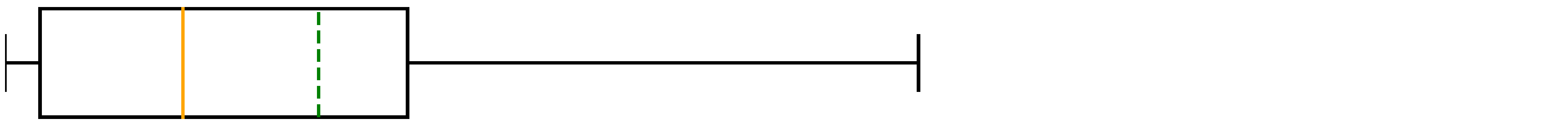} & \includegraphics[width=0.25\textwidth]{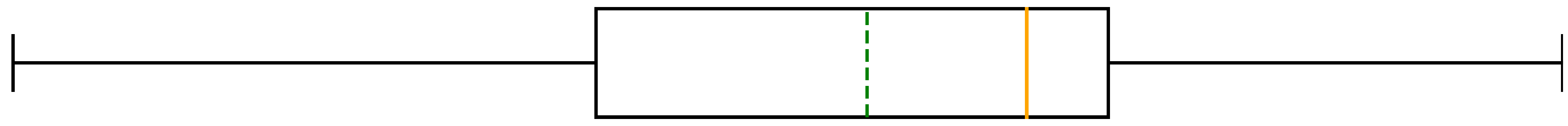}  & \includegraphics[width=0.25\textwidth]{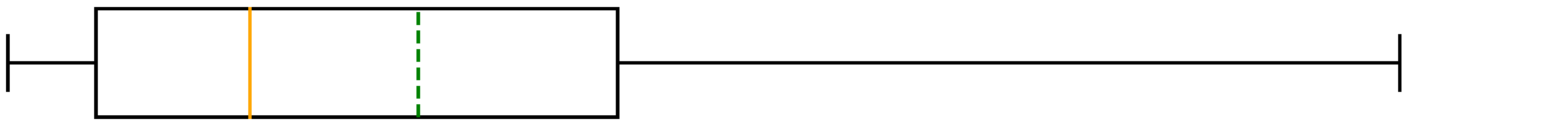} & 1.814 & 17.637 & 0.941 & 0.452 & 0.683 & 0.743 & 0.065 & 0.112 \\
 & & \resizebox{3.0mm}{3.0mm}{\myreconstruction}\hspace{0.5mm}DAGMM & \includegraphics[width=0.25\textwidth]{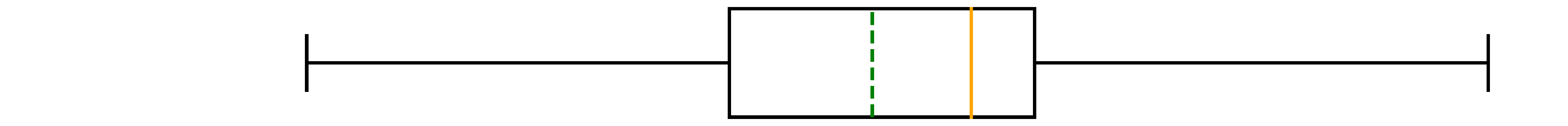} & \includegraphics[width=0.25\textwidth]{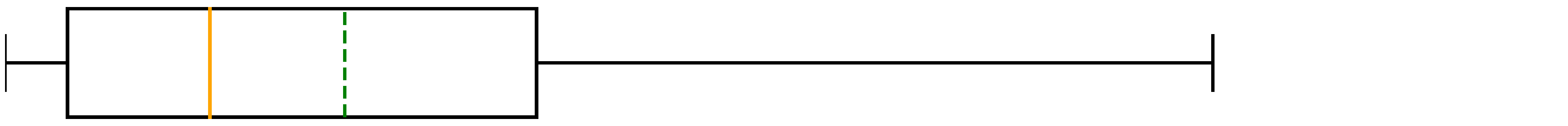} & \includegraphics[width=0.25\textwidth]{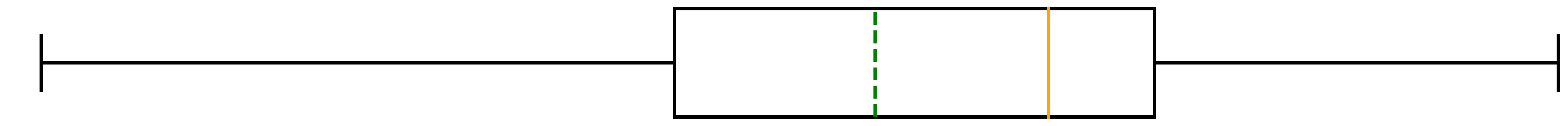}  & \includegraphics[width=0.25\textwidth]{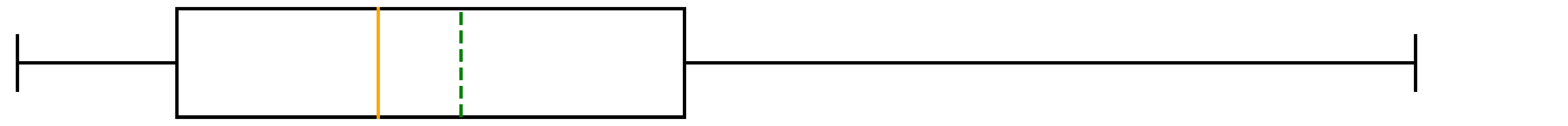} & 6.838 & 74.846 & 33.441 & 19.729 & 0.828 & 0.847 & 0.109 & 0.111 \\
 & & \resizebox{3.0mm}{3.0mm}{\myreconstruction}\hspace{0.5mm}CATCH & \includegraphics[width=0.25\textwidth]{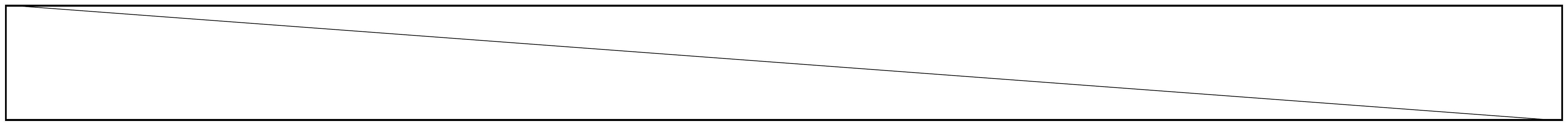} & \includegraphics[width=0.25\textwidth]{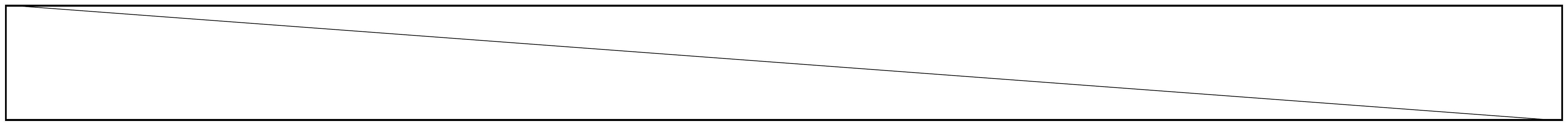} & \includegraphics[width=0.25\textwidth]{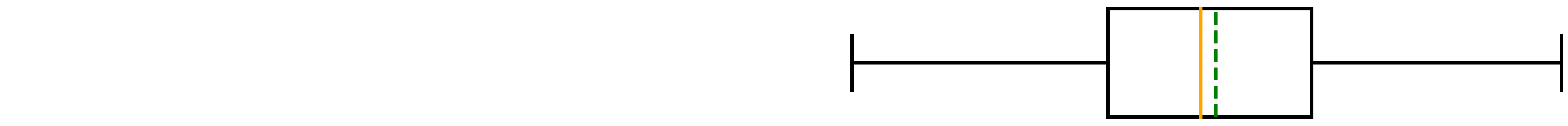}  & \includegraphics[width=0.25\textwidth]{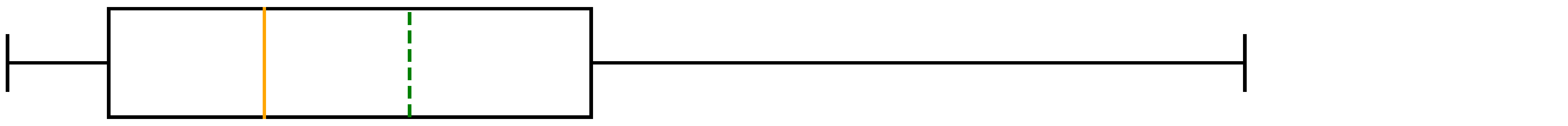} & \includegraphics[width=0.06\textwidth]{figures/main-table/no-data.pdf} & 51.000 & \includegraphics[width=0.06\textwidth]{figures/main-table/no-data.pdf} & 0.049 & \includegraphics[width=0.06\textwidth]{figures/main-table/no-data.pdf} & 1.373 & \includegraphics[width=0.06\textwidth]{figures/main-table/no-data.pdf} & 5.154 \\
 & & \resizebox{3.0mm}{3.0mm}{\myreconstruction}\hspace{0.5mm}DUET & \includegraphics[width=0.25\textwidth]{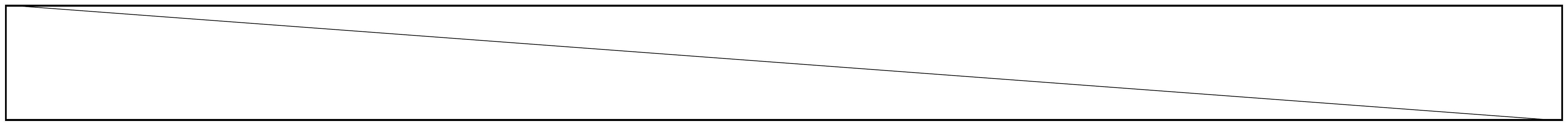} & \includegraphics[width=0.25\textwidth]{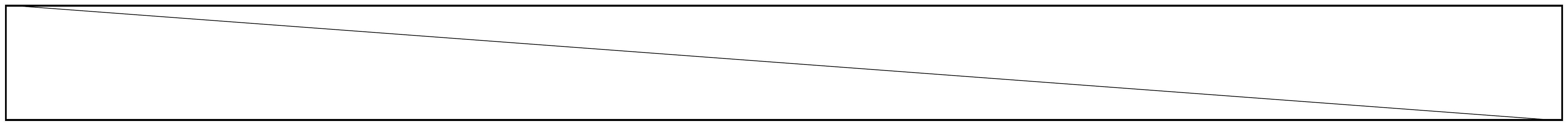} & \includegraphics[width=0.25\textwidth]{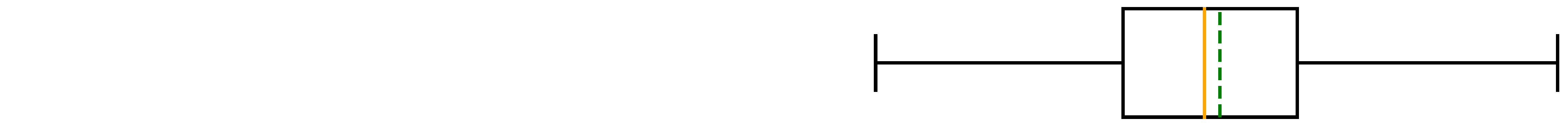}  & \includegraphics[width=0.25\textwidth]{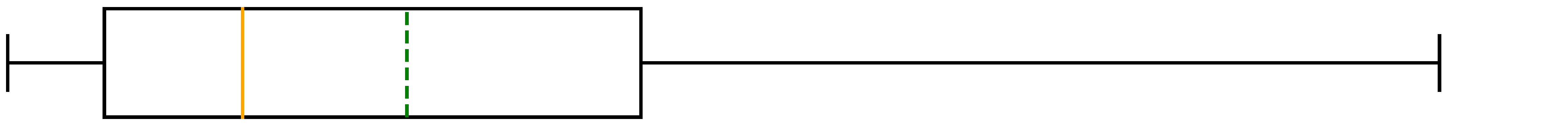} & \includegraphics[width=0.06\textwidth]{figures/main-table/no-data.pdf} & 38.778 & \includegraphics[width=0.06\textwidth]{figures/main-table/no-data.pdf} & 0.032 & \includegraphics[width=0.06\textwidth]{figures/main-table/no-data.pdf} & 1.049 & \includegraphics[width=0.06\textwidth]{figures/main-table/no-data.pdf} & 0.500 \\
\cmidrule(lr){2-15}
& \multirow{4}{*}{\rotatebox{90}{\textbf{LLM}}} &
\resizebox{3.0mm}{3.0mm}{\myforecasting}\hspace{0.5mm}CALF & \includegraphics[width=0.25\textwidth]{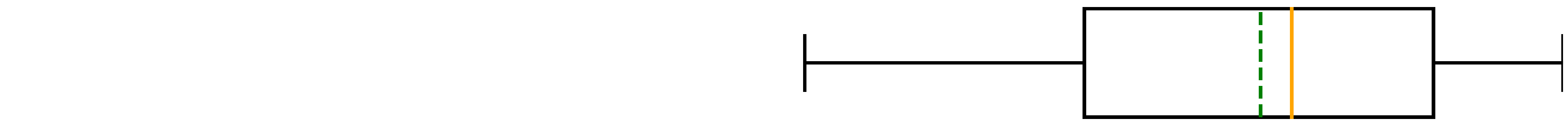} & \includegraphics[width=0.25\textwidth]{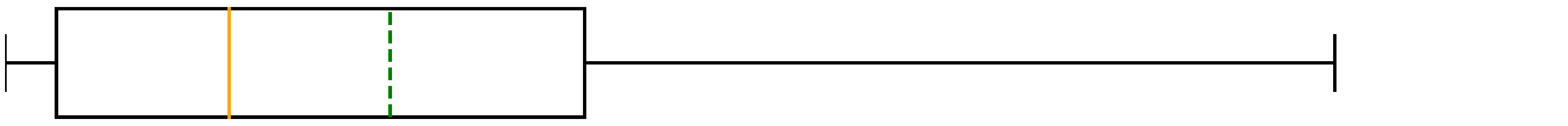} & \includegraphics[width=0.25\textwidth]{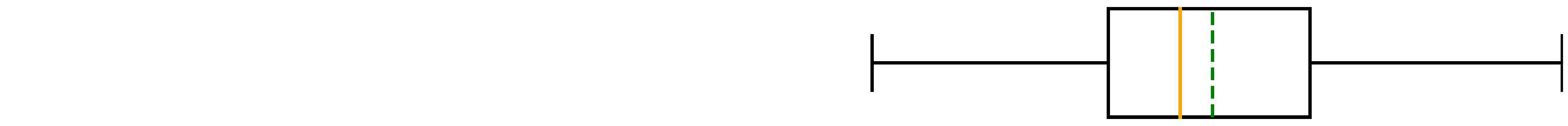}  & \includegraphics[width=0.25\textwidth]{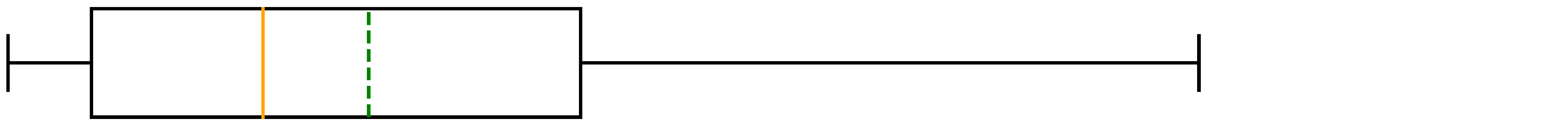} & 3.930 & 34.475 & 0.056 & 0.035 & 1.063 & 1.073 & 1.871 & 1.877 \\
 & & \resizebox{3.0mm}{3.0mm}{\myforecasting}\hspace{0.5mm}LLMMixer & \includegraphics[width=0.25\textwidth]{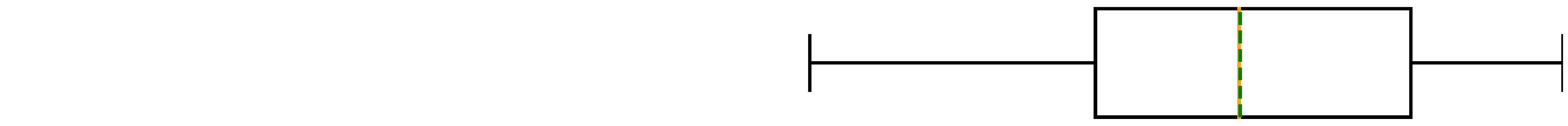} & \includegraphics[width=0.25\textwidth]{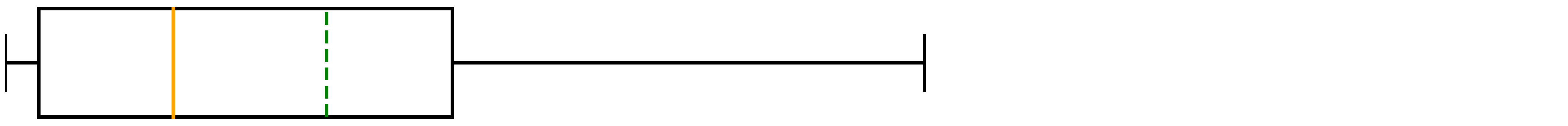} & \includegraphics[width=0.25\textwidth]{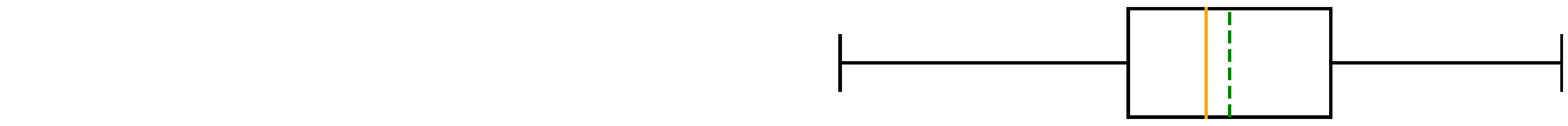}  & \includegraphics[width=0.25\textwidth]{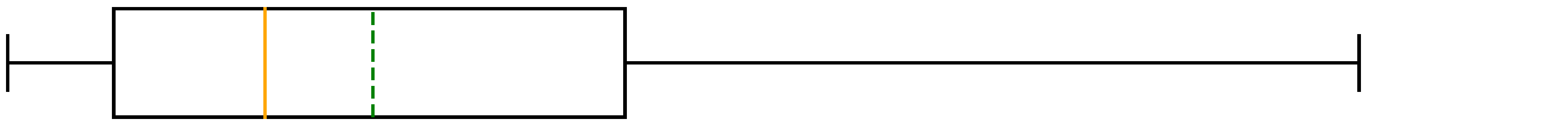} & 6.698 & 45.499 & 0.079 & 0.060 & 1.069 & 1.080 & 1.213 & 1.211 \\
 & & \resizebox{3.0mm}{3.0mm}{\myforecasting}\hspace{0.5mm}GPT4TS & \includegraphics[width=0.25\textwidth]{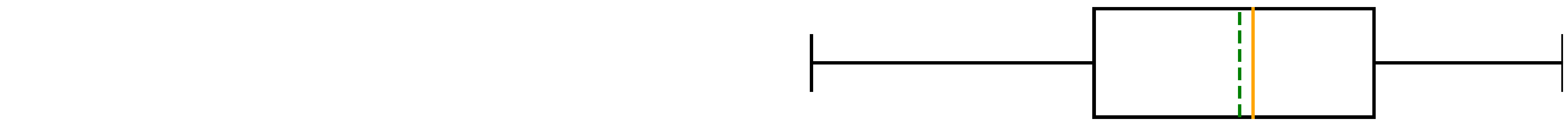} & \includegraphics[width=0.25\textwidth]{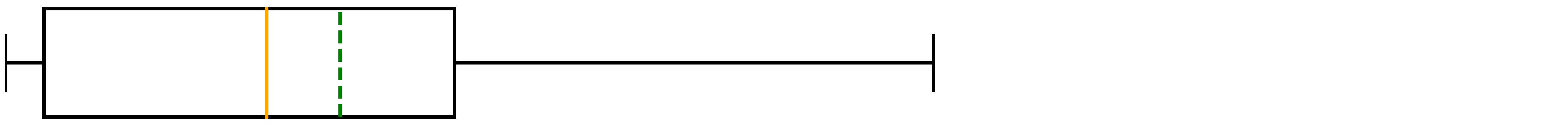} & \includegraphics[width=0.25\textwidth]{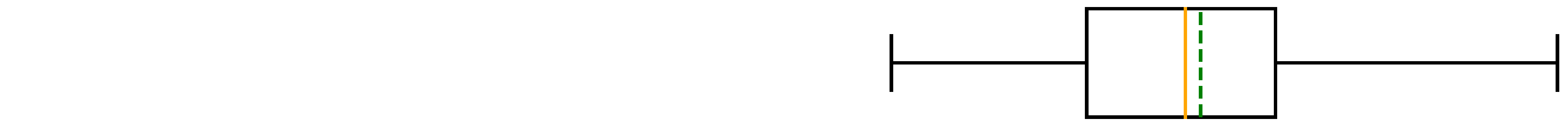}  & \includegraphics[width=0.25\textwidth]{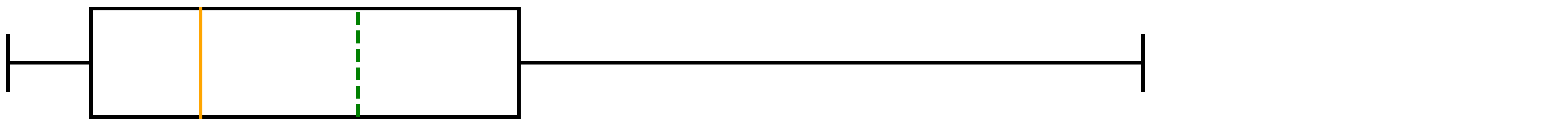} & 4.249 & 37.492 & 0.085 & 0.029 & 0.851 & 0.856 & 2.605 & 2.461 \\
 & & \resizebox{3.0mm}{3.0mm}{\myreconstruction}\hspace{0.5mm}UniTime & \includegraphics[width=0.25\textwidth]{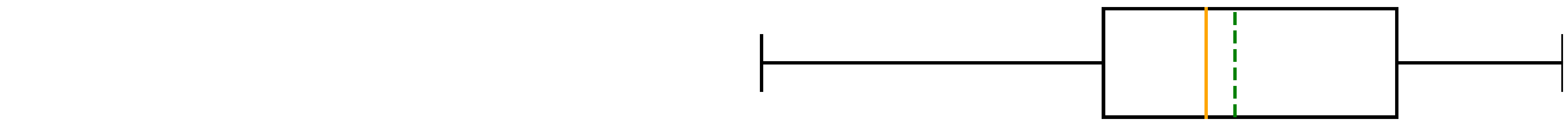} & \includegraphics[width=0.25\textwidth]{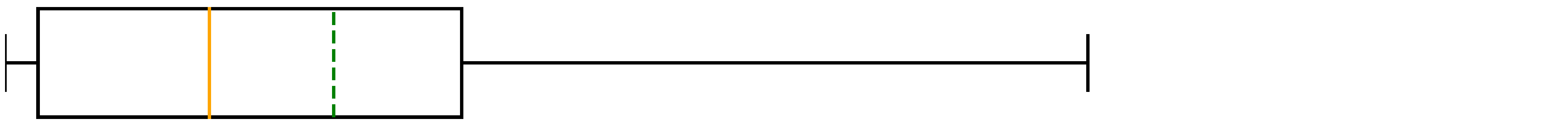} & \includegraphics[width=0.25\textwidth]{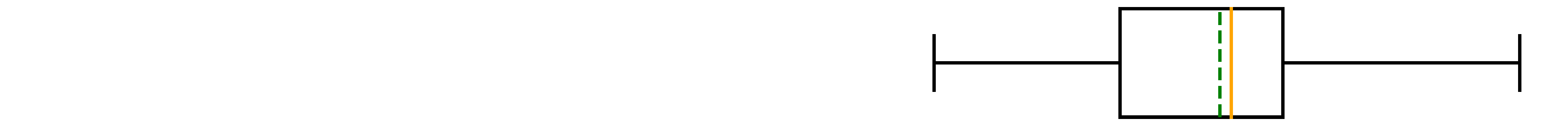}  & \includegraphics[width=0.25\textwidth]{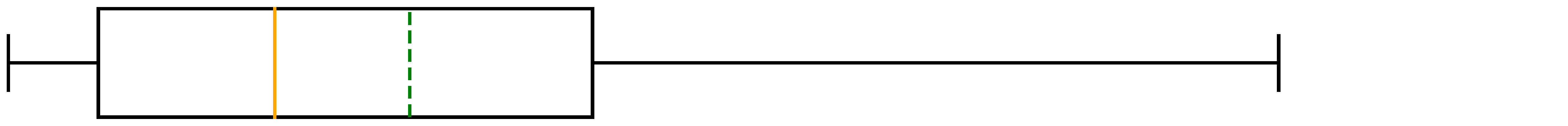} & 6.343 & 393.426 & 0.059 & 0.040 & 0.971 & 1.006 & 2.873 & 8.285 \\
\cmidrule(lr){2-15}
& \multirow{7}{*}{\rotatebox{90}{\textbf{Pre-trained}}} &
\resizebox{3.0mm}{3.0mm}{\myforecasting}\hspace{0.5mm}TimesFM & \includegraphics[width=0.25\textwidth]{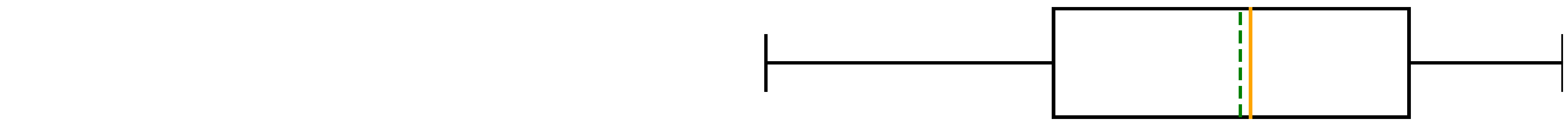} & \includegraphics[width=0.25\textwidth]{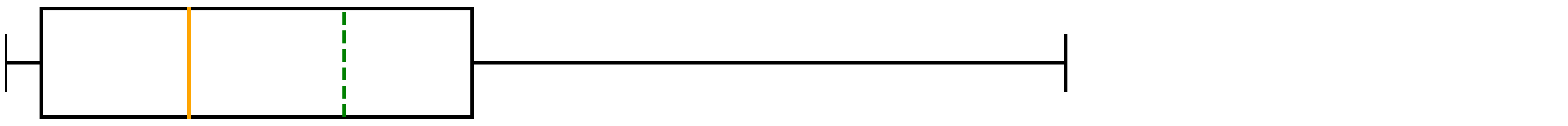} & \includegraphics[width=0.25\textwidth]{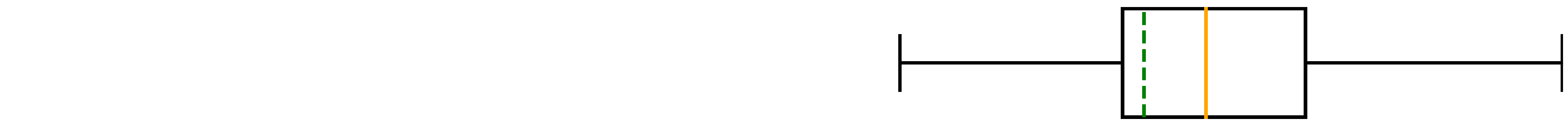}  & \includegraphics[width=0.25\textwidth]{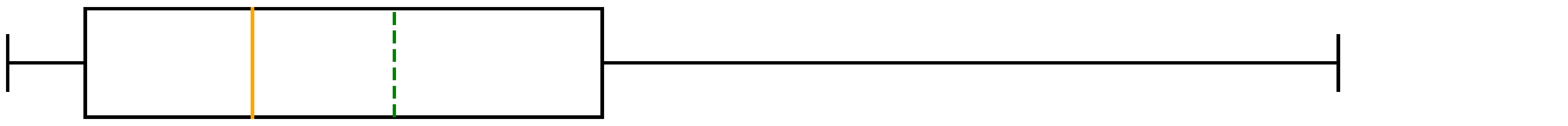} & 24.809 & 441.420 & 0.151 & 0.211 & 0.902 & 0.941 & 5.420 & 7.123 \\
 & & \resizebox{3.0mm}{3.0mm}{\myreconstruction}\hspace{0.5mm}Dada & \includegraphics[width=0.25\textwidth]{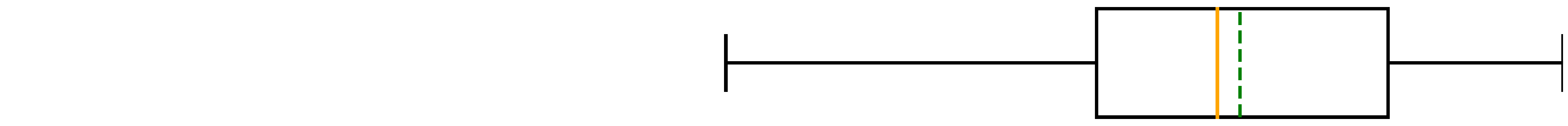} & \includegraphics[width=0.25\textwidth]{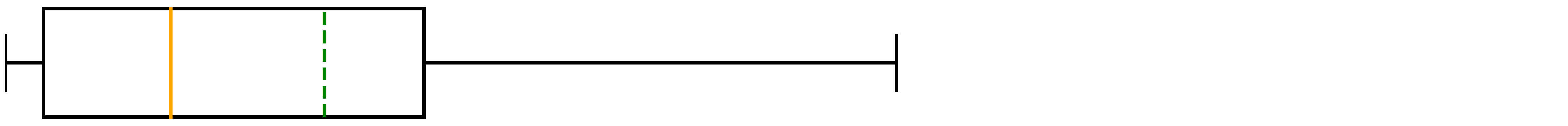} & \includegraphics[width=0.25\textwidth]{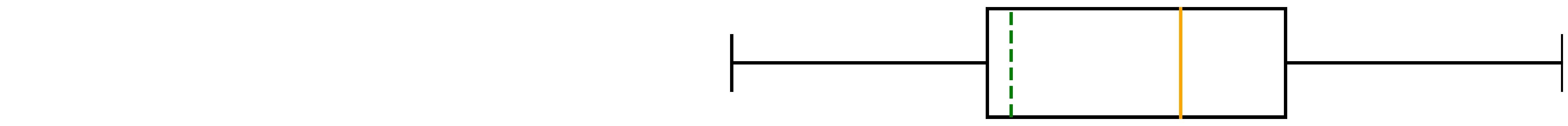}  & \includegraphics[width=0.25\textwidth]{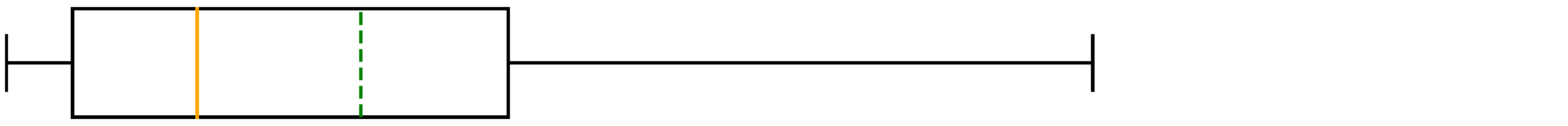} & 4.144 & 47.752 & 0.049 & 0.095 & 1.061 & 1.098 & 1.561 & 4.311 \\
 & & \resizebox{3.0mm}{3.0mm}{\myreconstruction}\hspace{0.5mm}Moment & \includegraphics[width=0.25\textwidth]{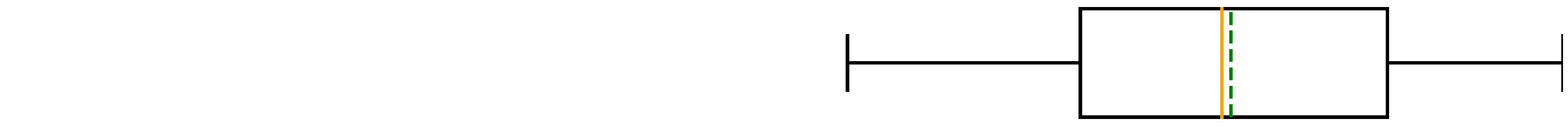} & \includegraphics[width=0.25\textwidth]{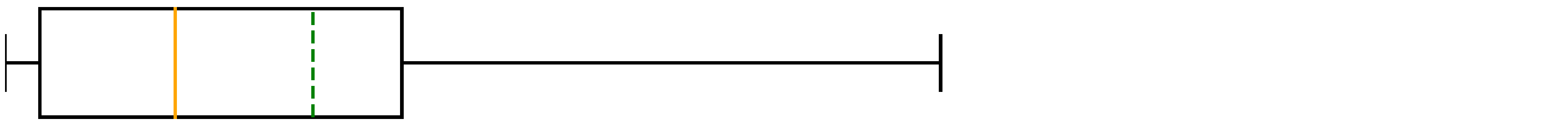} & \includegraphics[width=0.25\textwidth]{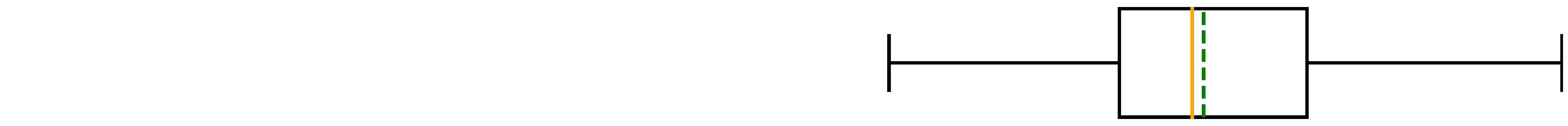}  & \includegraphics[width=0.25\textwidth]{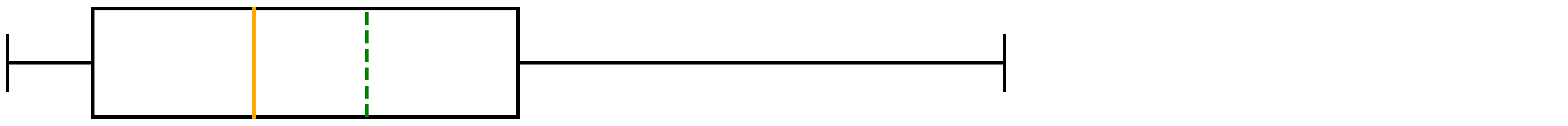} & 17.115 & 1195.658 & 0.094 & 0.102 & 1.144 & 1.165 & 4.291 & 5.201 \\
 & & \resizebox{3.0mm}{3.0mm}{\myreconstruction}\hspace{0.5mm}UniTS & \includegraphics[width=0.25\textwidth]{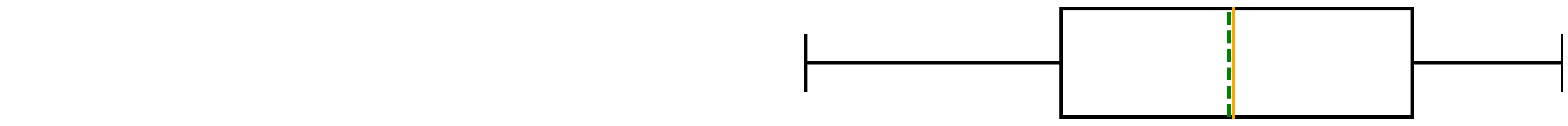} & \includegraphics[width=0.25\textwidth]{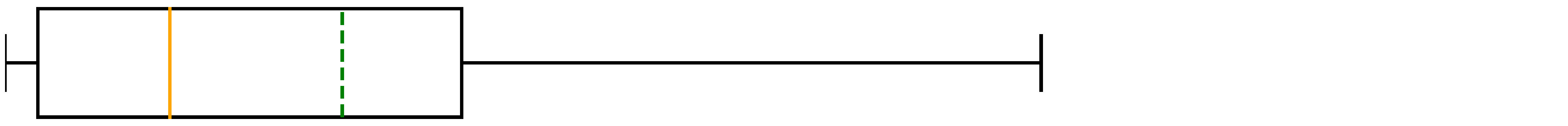} & \includegraphics[width=0.25\textwidth]{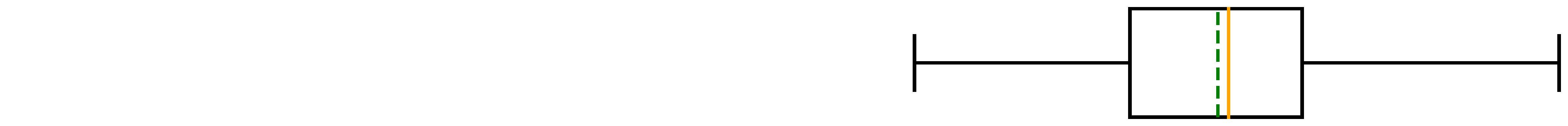}  & \includegraphics[width=0.25\textwidth]{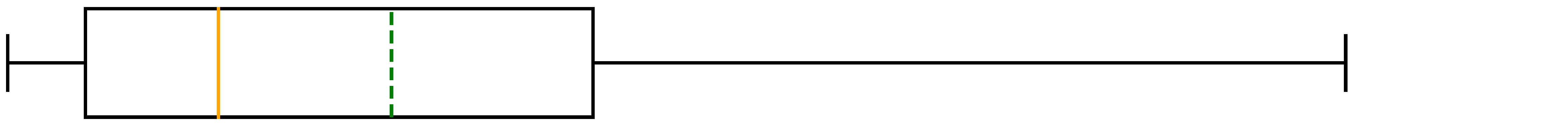} & 4.505 & 51.667 & 0.038 & 0.035 & 0.860 & 0.882 & 0.182 & 0.293 \\
 & & \resizebox{3.0mm}{3.0mm}{\myforecasting}\hspace{0.5mm}Chronos & \includegraphics[width=0.25\textwidth]{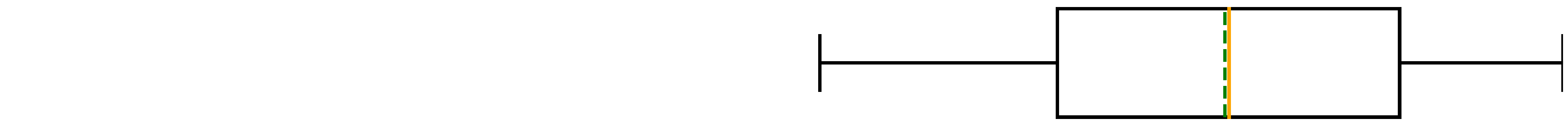} & \includegraphics[width=0.25\textwidth]{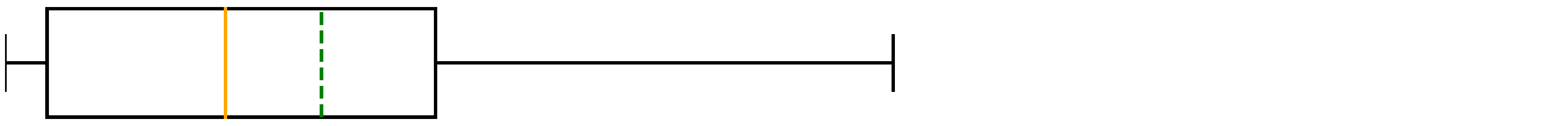} & \includegraphics[width=0.25\textwidth]{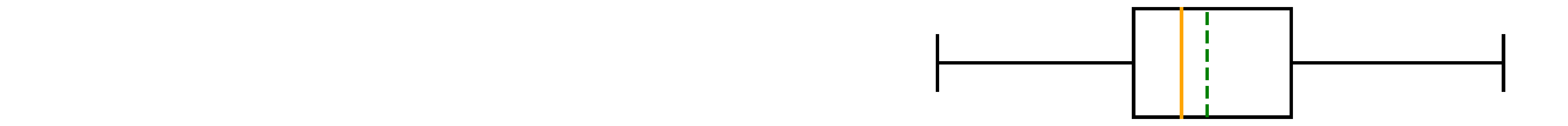}  & \includegraphics[width=0.25\textwidth]{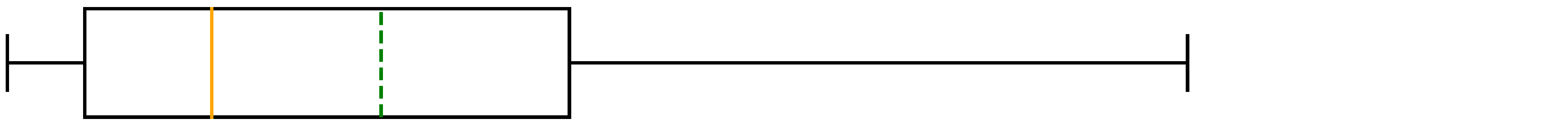} & 17.714 & 1053.727 & 0.296 & 0.515 & 1.047 & 1.066 & 0.506 & 0.506 \\
 & & \resizebox{3.0mm}{3.0mm}{\myforecasting}\hspace{0.5mm}Timer & \includegraphics[width=0.25\textwidth]{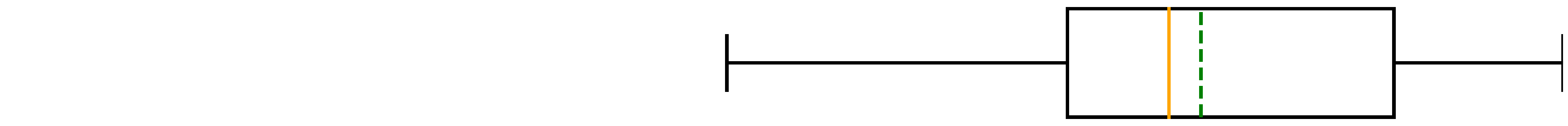} & \includegraphics[width=0.25\textwidth]{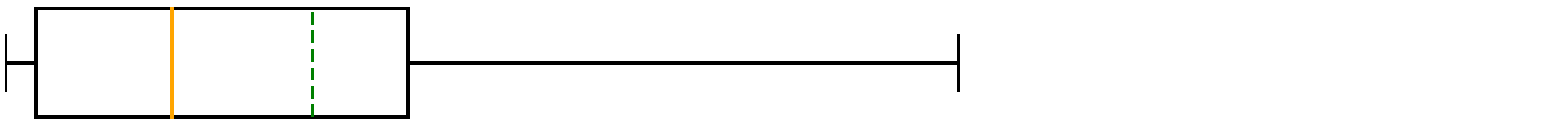} & \includegraphics[width=0.25\textwidth]{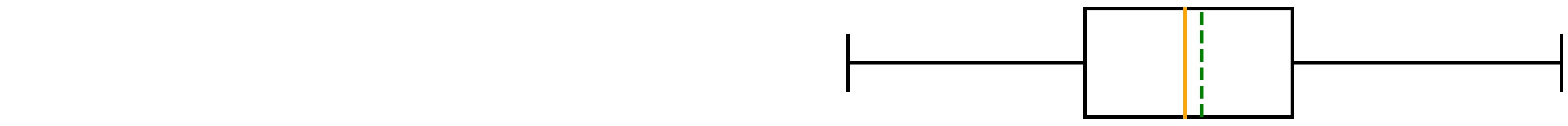}  & \includegraphics[width=0.25\textwidth]{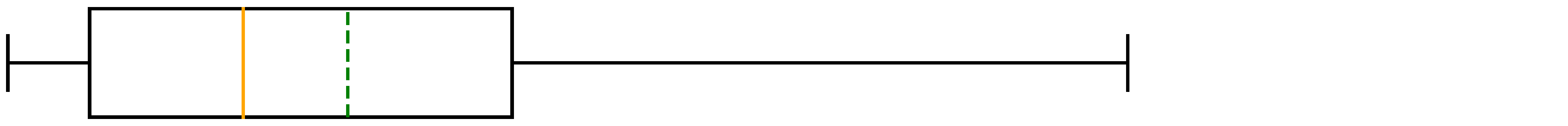} & 3.672 & 39.233 & 0.042 & 0.040 & 0.900 & 0.966 & 0.438 & 0.566 \\
 & & \resizebox{3.0mm}{3.0mm}{\myforecasting}\hspace{0.5mm}TTM & \includegraphics[width=0.25\textwidth]{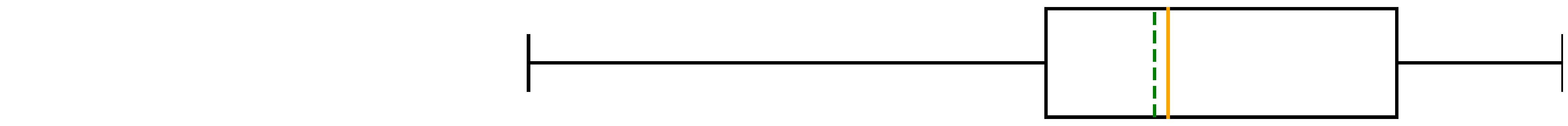} & \includegraphics[width=0.25\textwidth]{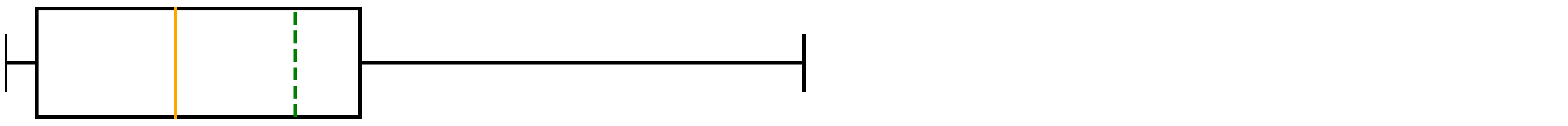} & \includegraphics[width=0.25\textwidth]{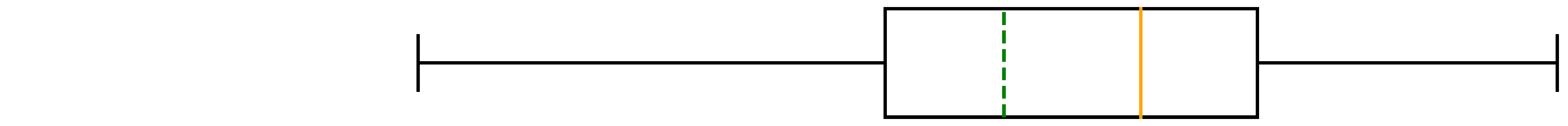}  & \includegraphics[width=0.25\textwidth]{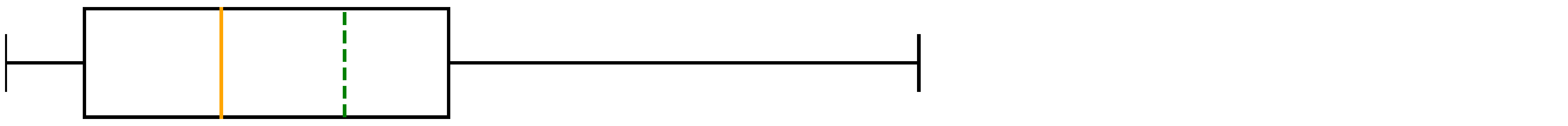} & 5.460 & 46.190 & 0.036 & 0.028 & 0.786 & 0.811 & 0.246 & 0.457 \\
\cmidrule(lr){1-15}
\multirow{11}{*}{\rotatebox{90}{\textbf{Few Shot}}} & \multirow{4}{*}{\rotatebox{90}{\textbf{LLM}}} &
\resizebox{3.0mm}{3.0mm}{\myforecasting}\hspace{0.5mm}GPT4TS & \includegraphics[width=0.25\textwidth]{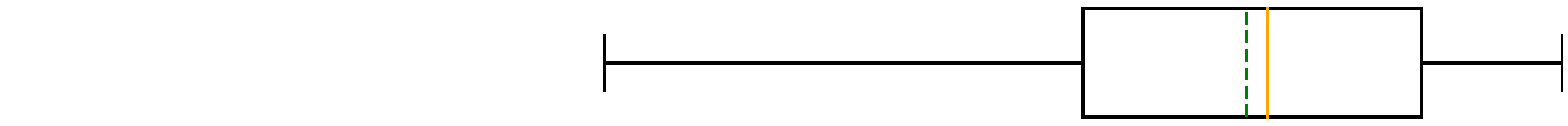} & \includegraphics[width=0.25\textwidth]{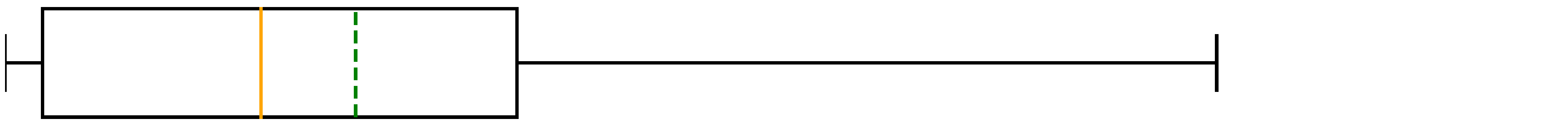} & \includegraphics[width=0.25\textwidth]{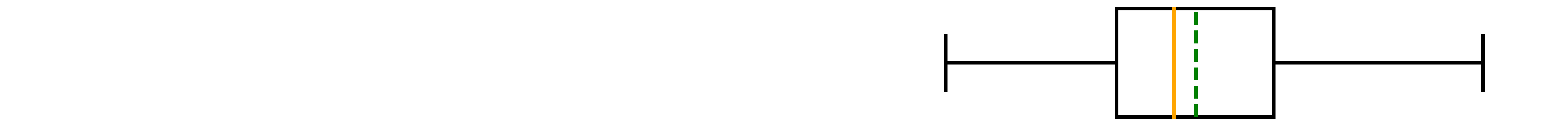}  & \includegraphics[width=0.25\textwidth]{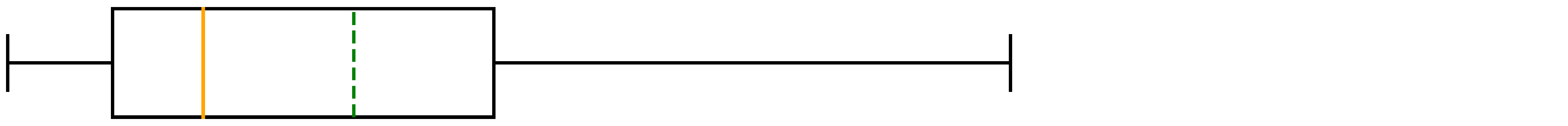} & 2.256 & 24.054 & 0.043 & 0.031 & 0.843 & 0.854 & 2.645 & 2.611 \\
 & & \resizebox{3.0mm}{3.0mm}{\myreconstruction}\hspace{0.5mm}UniTime & \includegraphics[width=0.25\textwidth]{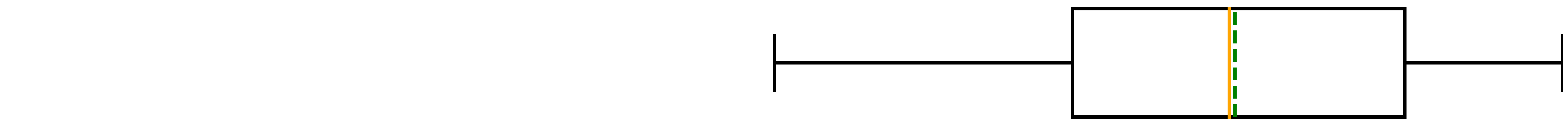} & \includegraphics[width=0.25\textwidth]{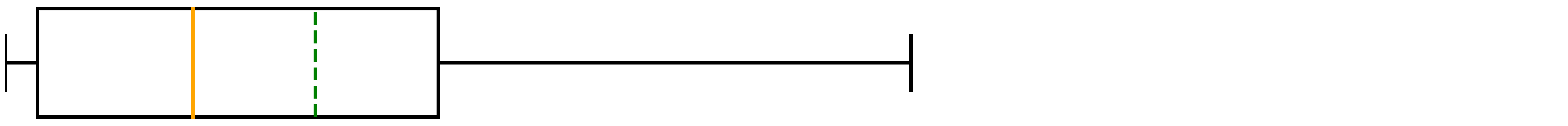} & \includegraphics[width=0.25\textwidth]{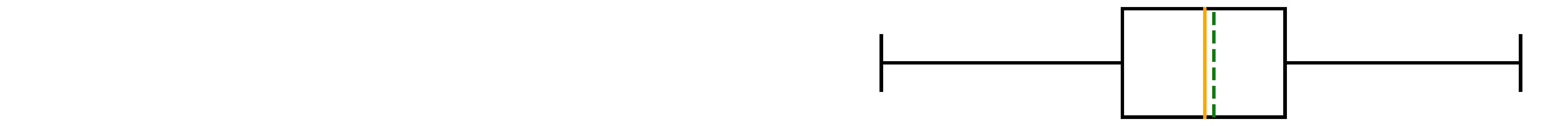}  & \includegraphics[width=0.25\textwidth]{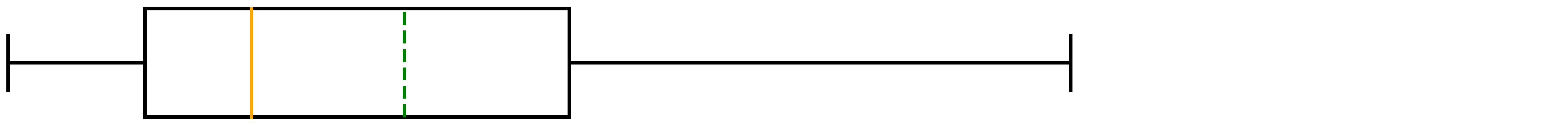} & 3.511 & 27.959 & 0.060 & 0.033 & 0.968 & 0.995 & 2.537 & 8.197 \\
 & & \resizebox{3.0mm}{3.0mm}{\myforecasting}\hspace{0.5mm}CALF & \includegraphics[width=0.25\textwidth]{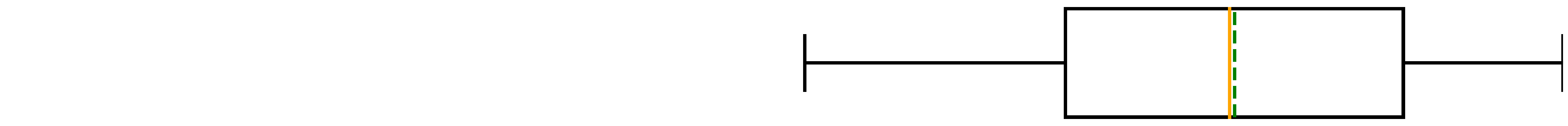} & \includegraphics[width=0.25\textwidth]{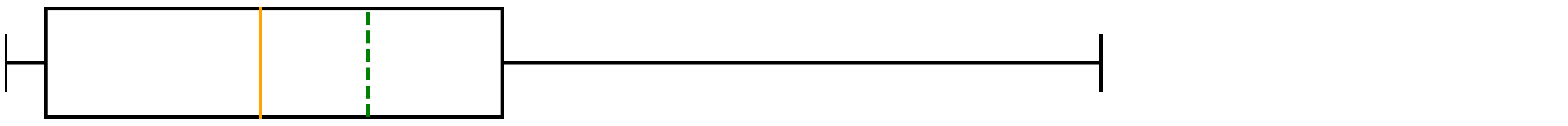} & \includegraphics[width=0.25\textwidth]{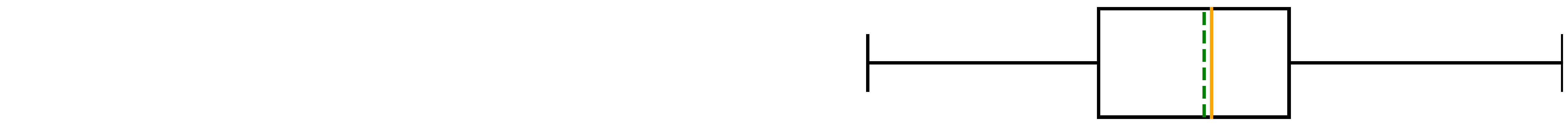}  & \includegraphics[width=0.25\textwidth]{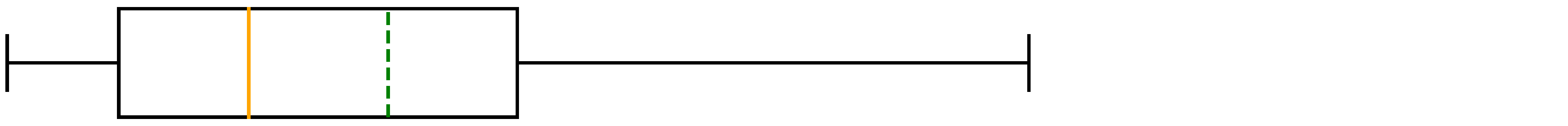} & 3.022 & 10.602 & 0.052 & 0.040 & 1.008 & 1.064 & 0.604 & 1.416 \\
 & & \resizebox{3.0mm}{3.0mm}{\myforecasting}\hspace{0.5mm}LLMMixer & \includegraphics[width=0.25\textwidth]{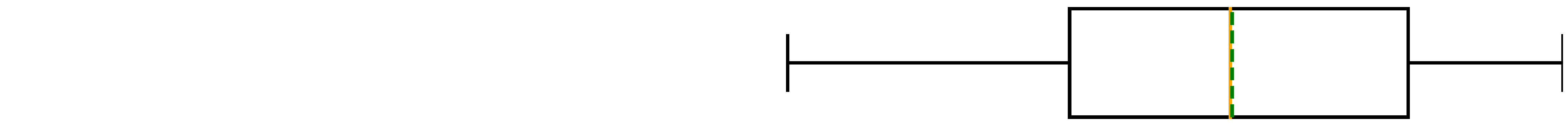} & \includegraphics[width=0.25\textwidth]{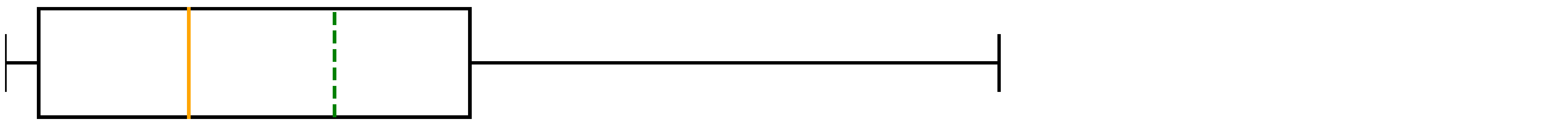} & \includegraphics[width=0.25\textwidth]{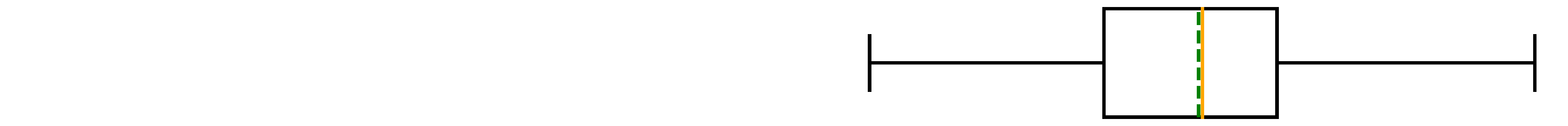}  & \includegraphics[width=0.25\textwidth]{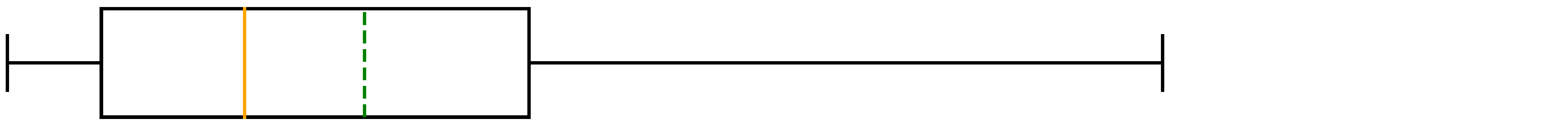} & 4.517 & 9.588 & 0.087 & 0.065 & 1.069 & 1.074 & 1.721 & 1.219 \\
\cmidrule(lr){2-15}
& \multirow{7}{*}{\rotatebox{90}{\textbf{Pre-trained}}} &
\resizebox{3.0mm}{3.0mm}{\myreconstruction}\hspace{0.5mm}UniTS & \includegraphics[width=0.25\textwidth]{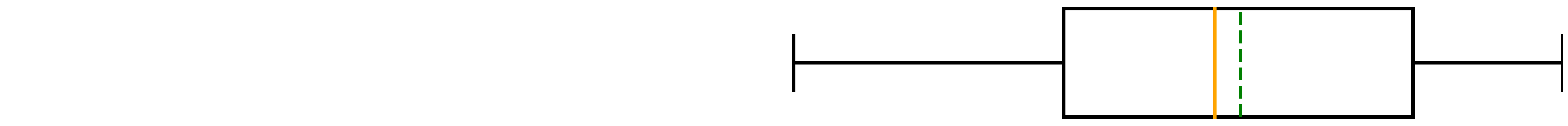} & \includegraphics[width=0.25\textwidth]{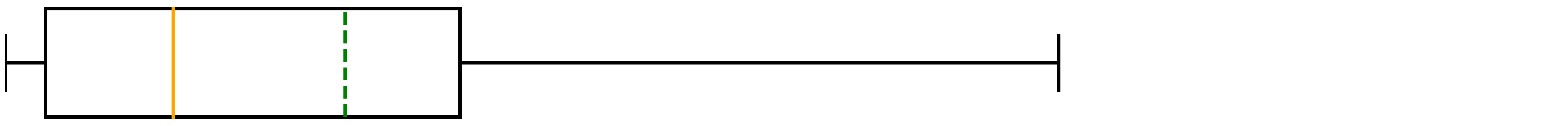} & \includegraphics[width=0.25\textwidth]{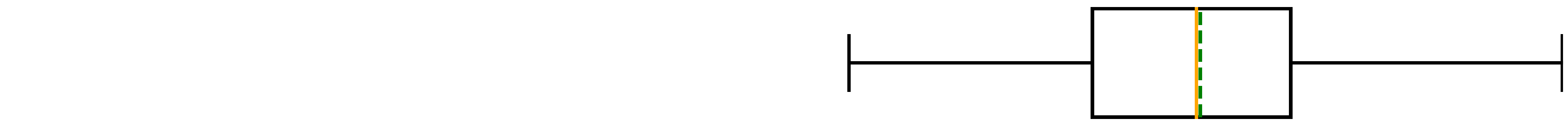}  & \includegraphics[width=0.25\textwidth]{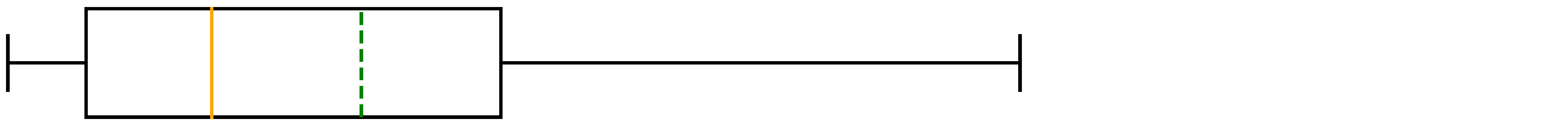} & 1.867 & 10.973 & 0.062 & 0.038 & 0.855 & 0.874 & 0.172 & 0.293 \\
 & & \resizebox{3.0mm}{3.0mm}{\myforecasting}\hspace{0.5mm}TimesFM & \includegraphics[width=0.25\textwidth]{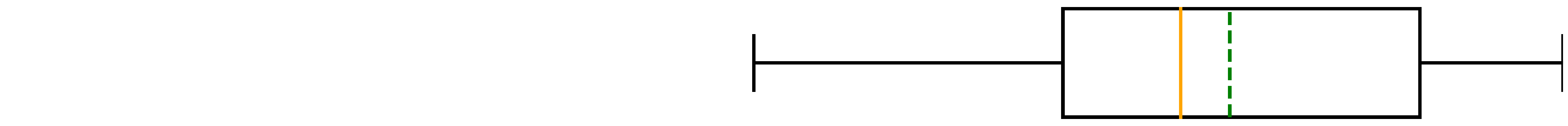} & \includegraphics[width=0.25\textwidth]{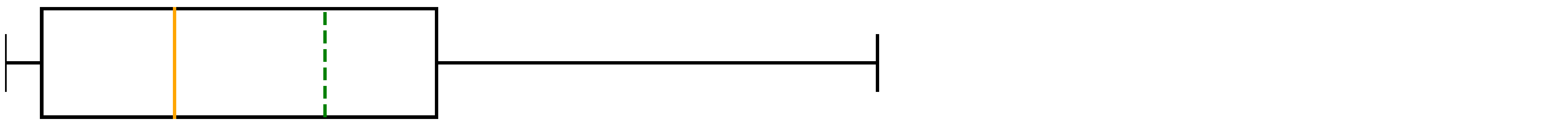} & \includegraphics[width=0.25\textwidth]{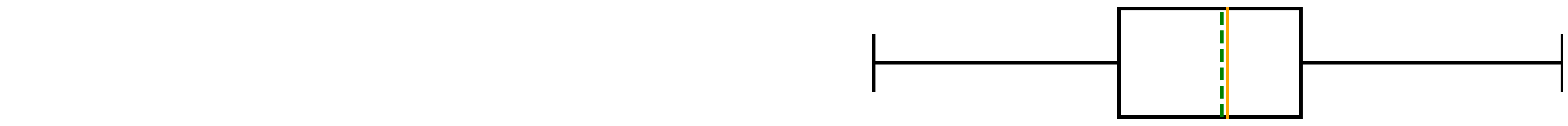}  & \includegraphics[width=0.25\textwidth]{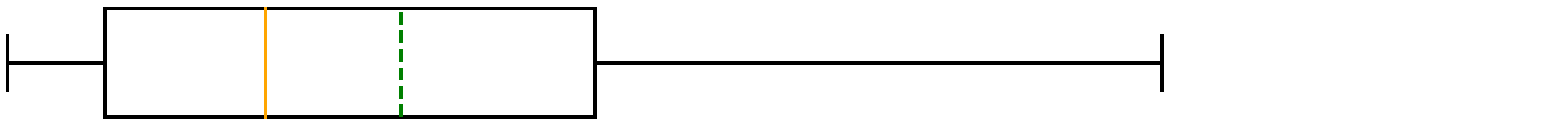} & 9.312 & 141.400 & 0.138 & 0.212 & 0.898 & 0.905 & 6.113 & 6.301 \\
 & & \resizebox{3.0mm}{3.0mm}{\myforecasting}\hspace{0.5mm}Chronos & \includegraphics[width=0.25\textwidth]{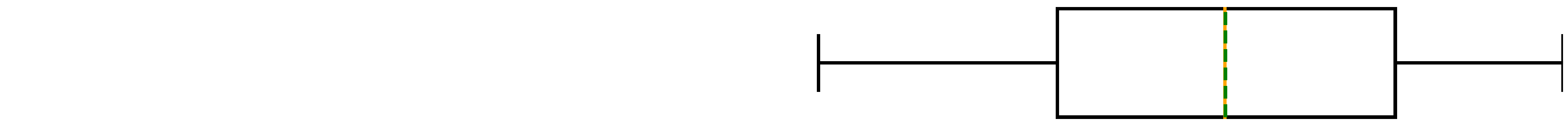} & \includegraphics[width=0.25\textwidth]{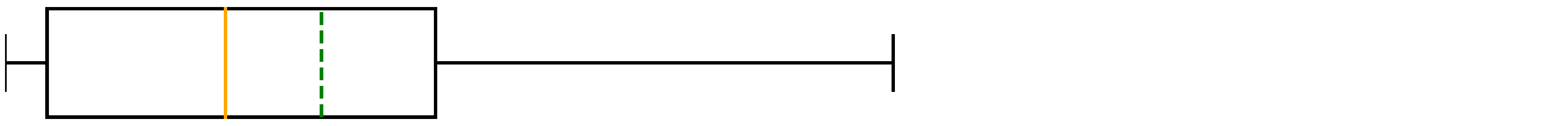} & \includegraphics[width=0.25\textwidth]{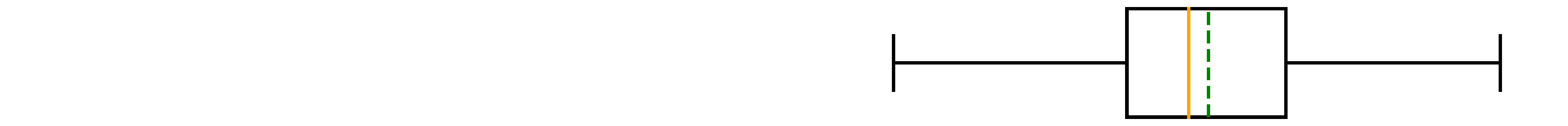}  & \includegraphics[width=0.25\textwidth]{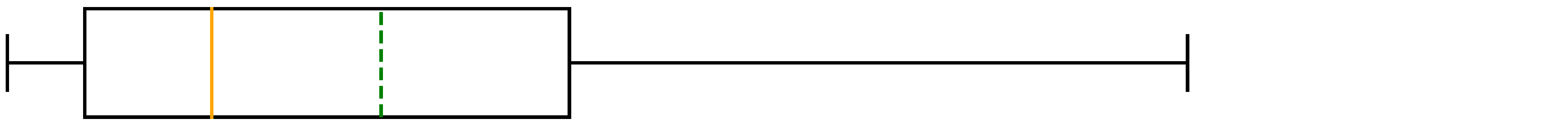} & 4.440 & 252.742 & 0.342 & 0.474 & 1.040 & 1.046 & 0.506 & 0.506 \\
 & & \resizebox{3.0mm}{3.0mm}{\myreconstruction}\hspace{0.5mm}Dada & \includegraphics[width=0.25\textwidth]{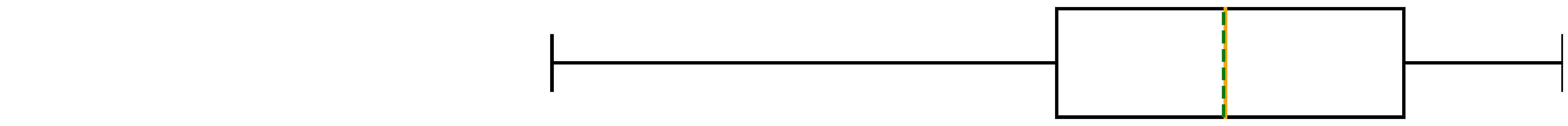} & \includegraphics[width=0.25\textwidth]{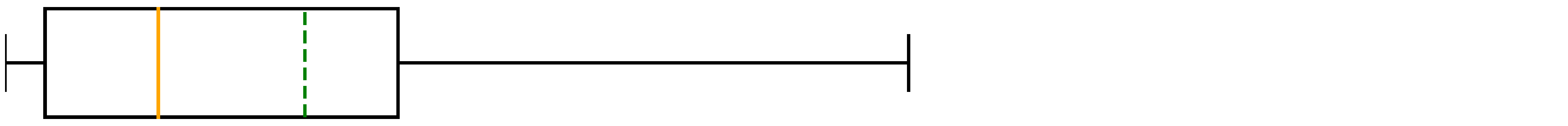} & \includegraphics[width=0.25\textwidth]{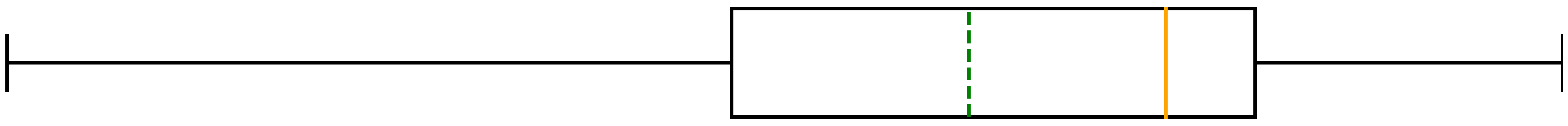}  & \includegraphics[width=0.25\textwidth]{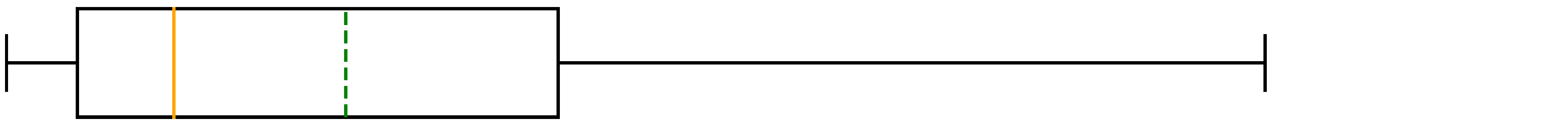} & 2.675 & 13.647 & 0.098 & 0.050 & 1.059 & 1.088 & 0.867 & 4.029 \\
 & & \resizebox{3.0mm}{3.0mm}{\myforecasting}\hspace{0.5mm}Timer & \includegraphics[width=0.25\textwidth]{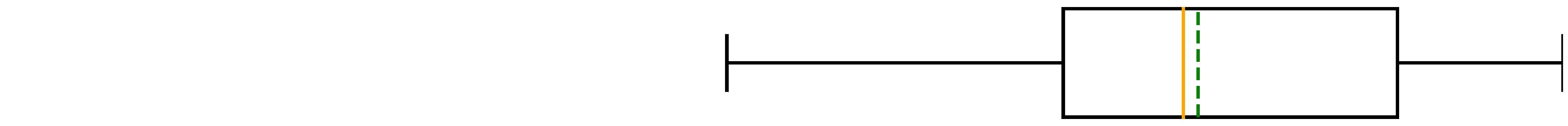} & \includegraphics[width=0.25\textwidth]{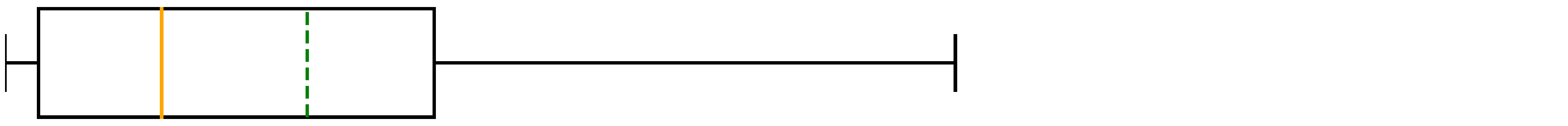} & \includegraphics[width=0.25\textwidth]{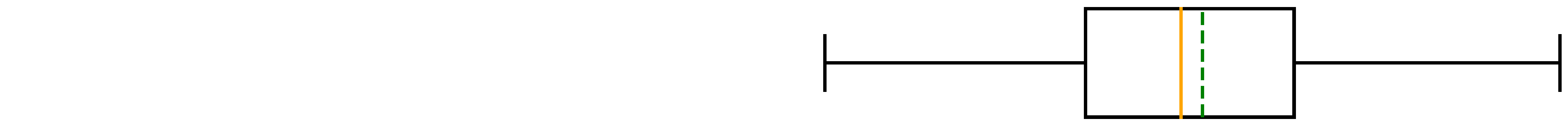}  & \includegraphics[width=0.25\textwidth]{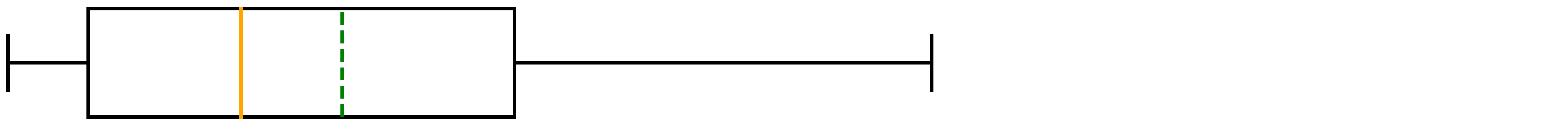} & 1.780 & 8.662 & 0.041 & 0.042 & 0.895 & 0.958 & 0.438 & 0.566 \\
 & & \resizebox{3.0mm}{3.0mm}{\myreconstruction}\hspace{0.5mm}Moment & \includegraphics[width=0.25\textwidth]{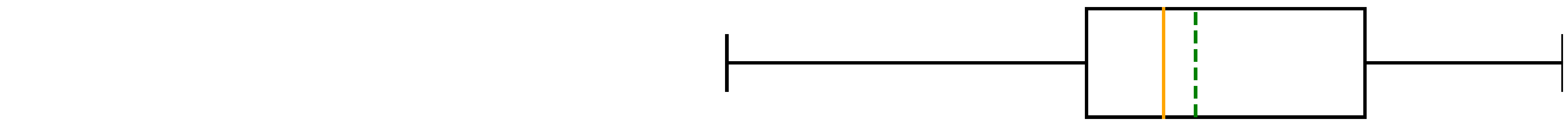} & \includegraphics[width=0.25\textwidth]{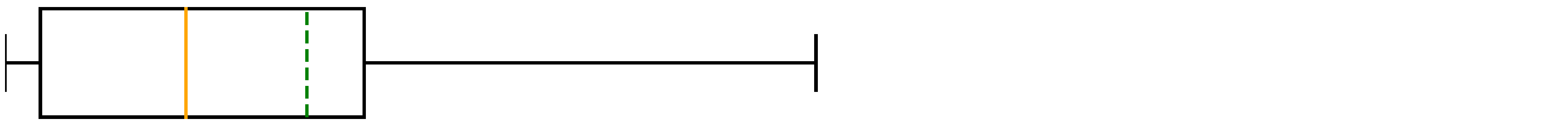} & \includegraphics[width=0.25\textwidth]{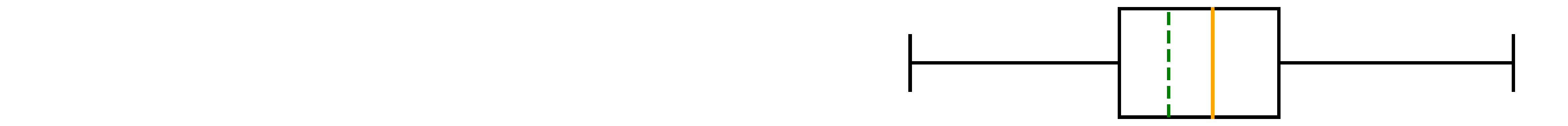}  & \includegraphics[width=0.25\textwidth]{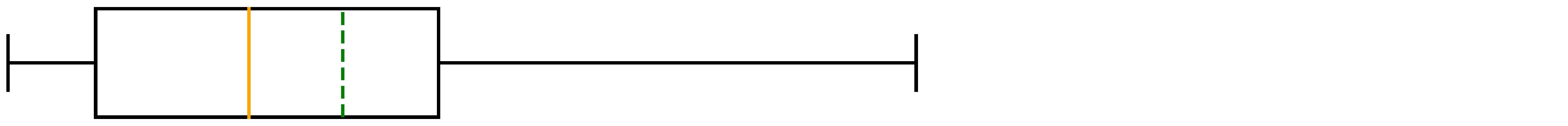} & 13.525 & 239.935 & 0.098 & 0.101 & 1.142 & 1.147 & 4.271 & 5.201 \\
 & & \resizebox{3.0mm}{3.0mm}{\myforecasting}\hspace{0.5mm}TTM & \includegraphics[width=0.25\textwidth]{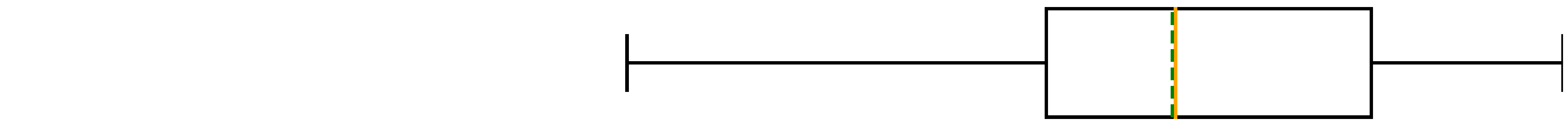} & \includegraphics[width=0.25\textwidth]{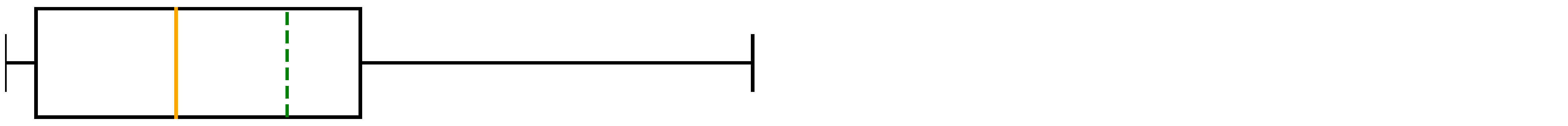} & \includegraphics[width=0.25\textwidth]{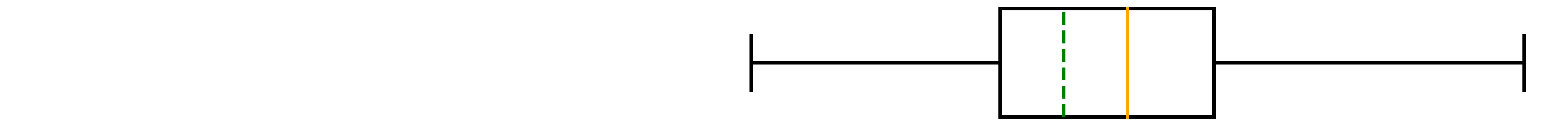}  & \includegraphics[width=0.25\textwidth]{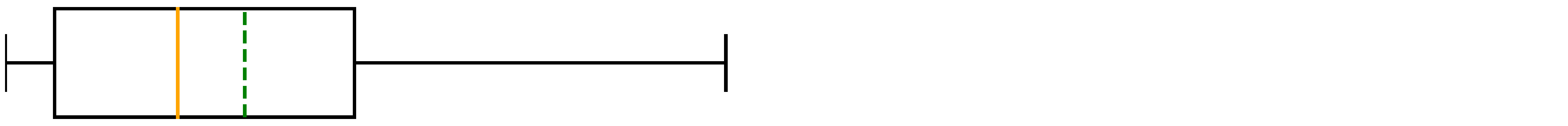} & 3.144 & 10.436 & 0.037 & 0.032 & 0.785 & 0.821 & 0.205 & 0.457 \\
\cmidrule(lr){1-15}
\multirow{7}{*}{\rotatebox{90}{\textbf{Zero Shot}}}& \multirow{7}{*}{\rotatebox{90}{\textbf{Pre-trained}}} &
\resizebox{3.0mm}{3.0mm}{\myreconstruction}\hspace{0.5mm}UniTS & \includegraphics[width=0.25\textwidth]{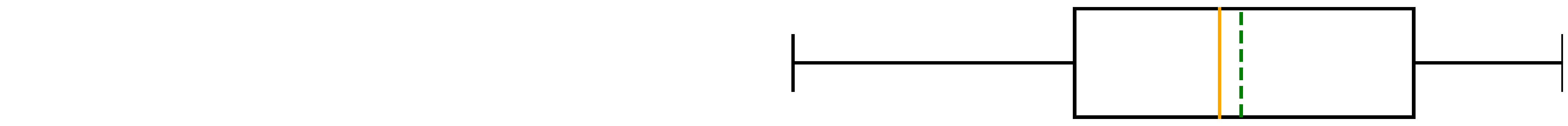} & \includegraphics[width=0.25\textwidth]{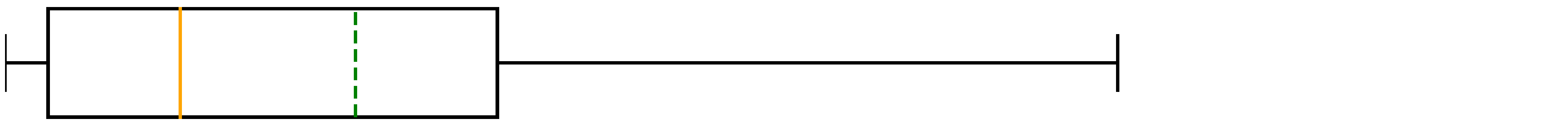} & \includegraphics[width=0.25\textwidth]{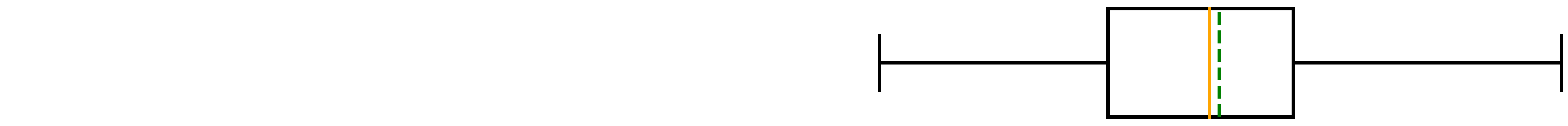}  & \includegraphics[width=0.25\textwidth]{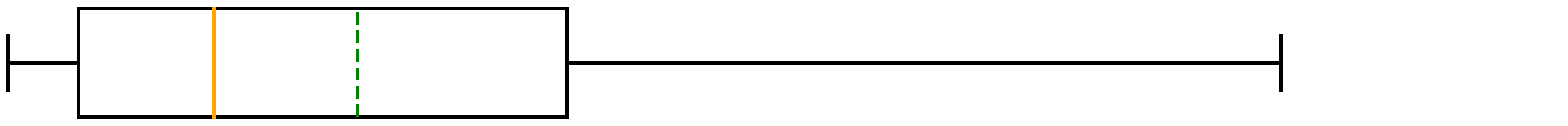} & 0.000 & 0.000 & 0.362 & 0.354 & 0.769 & 0.773 & 0.129 & 0.108 \\
 & & \resizebox{3.0mm}{3.0mm}{\myforecasting}\hspace{0.5mm}TimesFM & \includegraphics[width=0.25\textwidth]{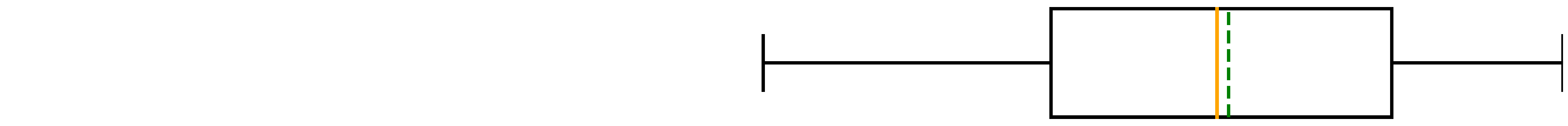} & \includegraphics[width=0.25\textwidth]{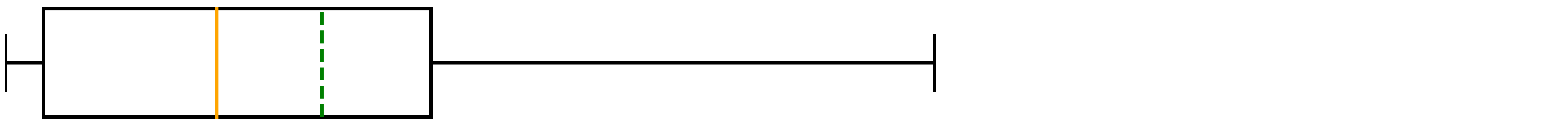} & \includegraphics[width=0.25\textwidth]{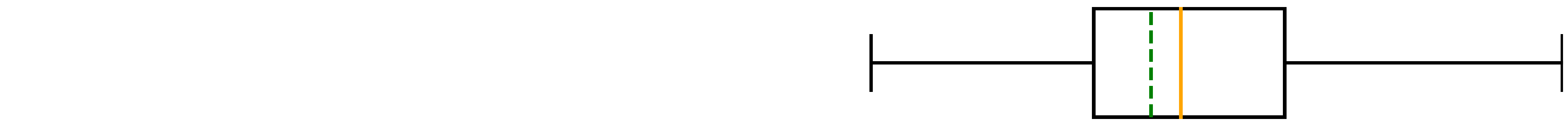}  & \includegraphics[width=0.25\textwidth]{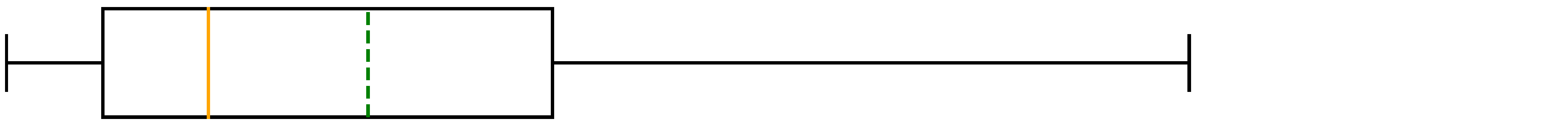} & 0.000 & 0.000 & 1.295 & 2.025 & 0.819 & 0.856 & 0.924 & 1.594 \\
 & & \resizebox{3.0mm}{3.0mm}{\myforecasting}\hspace{0.5mm}Chronos & \includegraphics[width=0.25\textwidth]{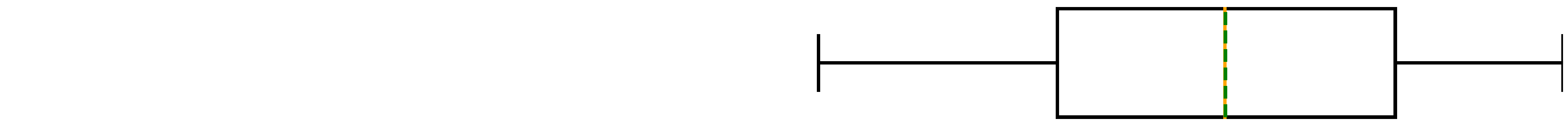} & \includegraphics[width=0.25\textwidth]{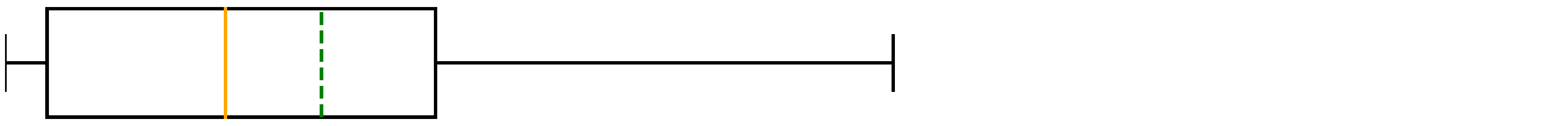} & \includegraphics[width=0.25\textwidth]{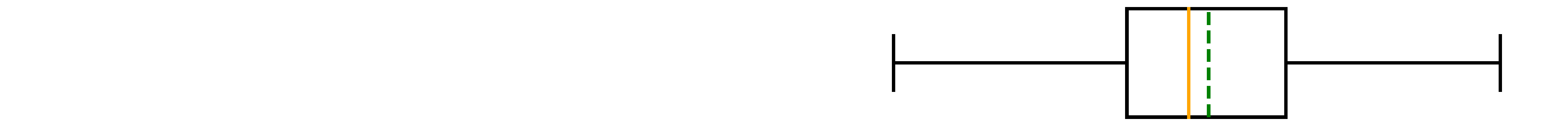}  & \includegraphics[width=0.25\textwidth]{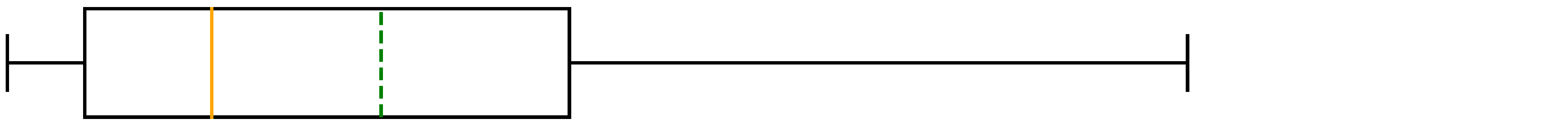} & 0.000 & 0.000 & 1.403 & 0.959 & 0.870 & 1.047 & 0.410 & 0.506 \\
 & & \resizebox{3.0mm}{3.0mm}{\myforecasting}\hspace{0.5mm}Timer & \includegraphics[width=0.25\textwidth]{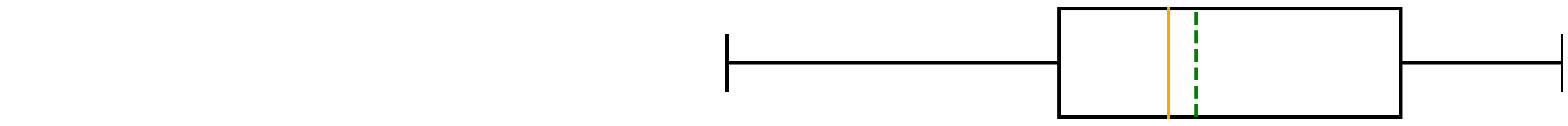} & \includegraphics[width=0.25\textwidth]{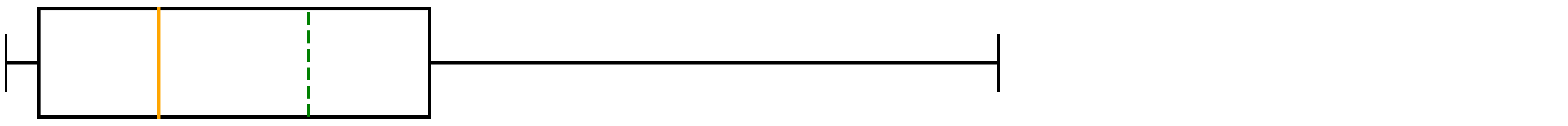} & \includegraphics[width=0.25\textwidth]{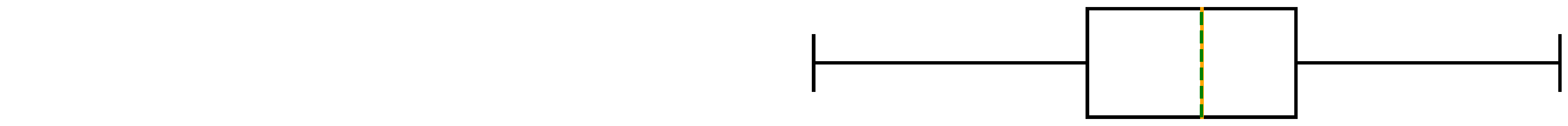}  & \includegraphics[width=0.25\textwidth]{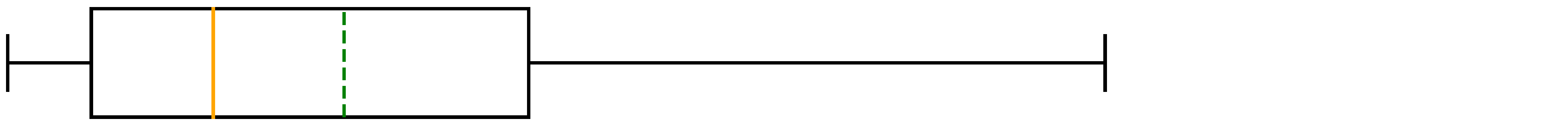} & 0.000 & 0.000 & 0.715 & 1.370 & 0.818 & 0.840 & 0.250 & 0.254 \\
 & & \resizebox{3.0mm}{3.0mm}{\myreconstruction}\hspace{0.5mm}Moment & \includegraphics[width=0.25\textwidth]{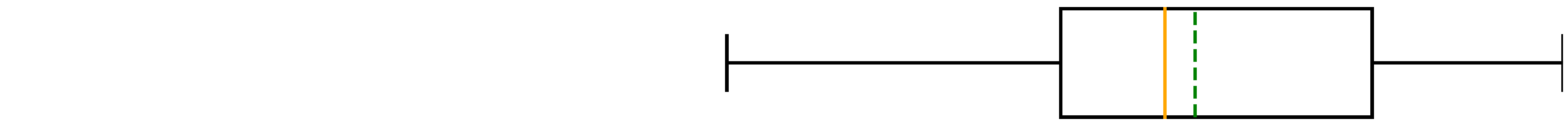} & \includegraphics[width=0.25\textwidth]{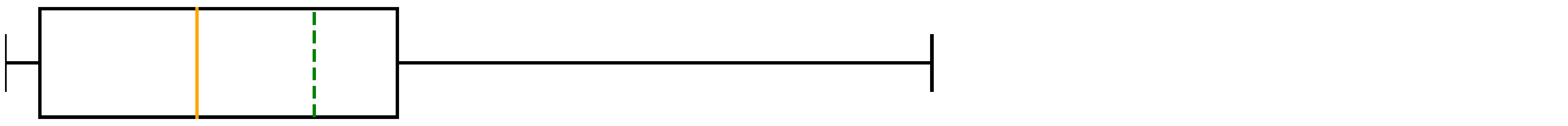} & \includegraphics[width=0.25\textwidth]{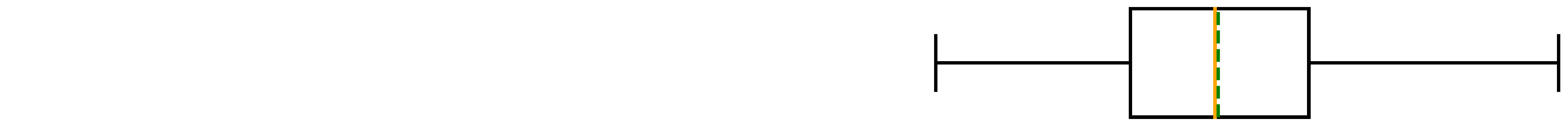}  & \includegraphics[width=0.25\textwidth]{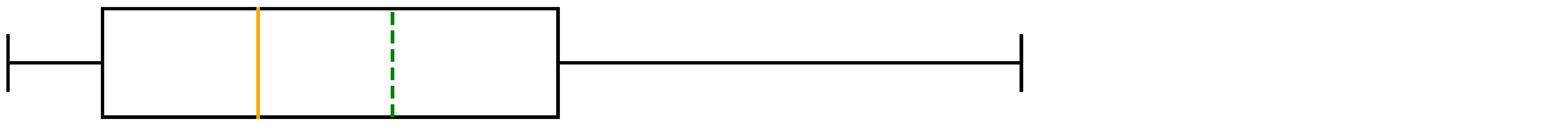} & 0.000 & 0.000 & 1.324 & 1.756 & 1.105 & 1.110 & 2.412 & 1.576 \\
 & & \resizebox{3.0mm}{3.0mm}{\myreconstruction}\hspace{0.5mm}Dada & \includegraphics[width=0.25\textwidth]{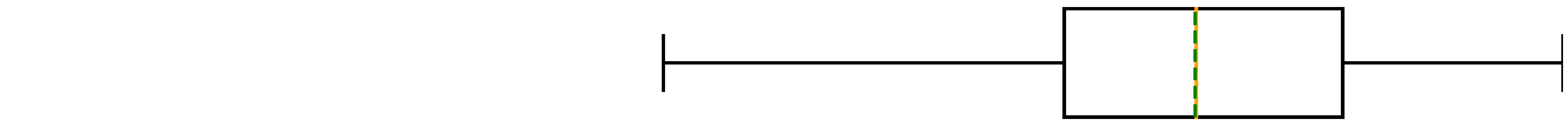} & \includegraphics[width=0.25\textwidth]{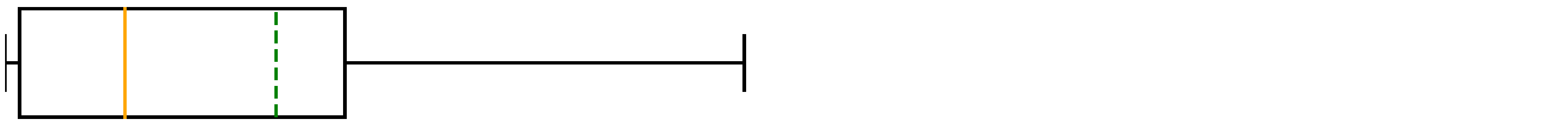} & \includegraphics[width=0.25\textwidth]{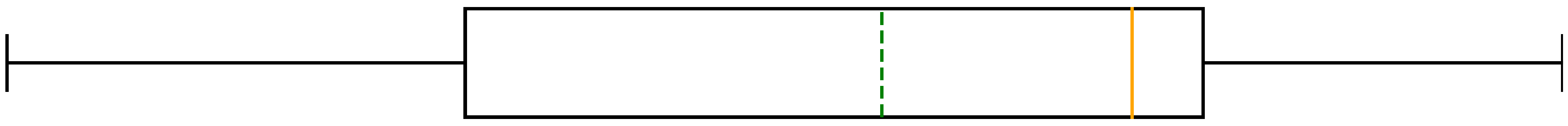}  & \includegraphics[width=0.25\textwidth]{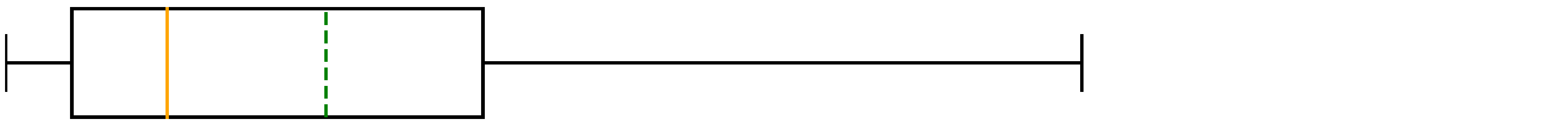} & 0.000 & 0.000 & 0.892 & 0.907 & 0.885 & 0.910 & 0.600 & 1.152 \\
 & & \resizebox{3.0mm}{3.0mm}{\myforecasting}\hspace{0.5mm}TTM & \includegraphics[width=0.25\textwidth]{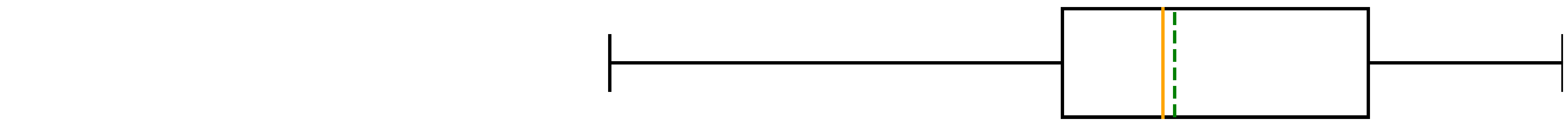} & \includegraphics[width=0.25\textwidth]{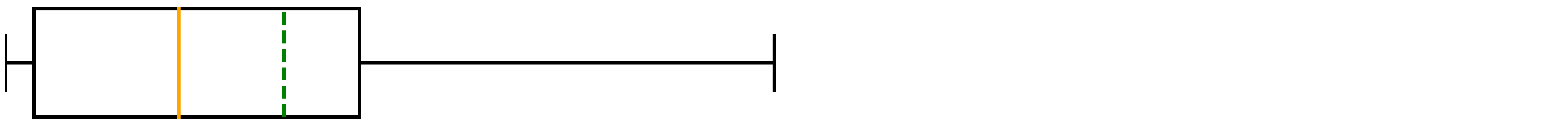} & \includegraphics[width=0.25\textwidth]{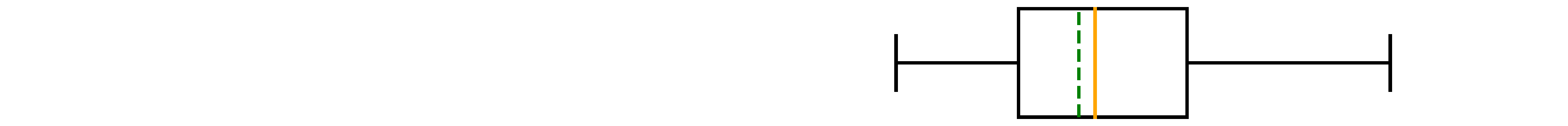}  & \includegraphics[width=0.25\textwidth]{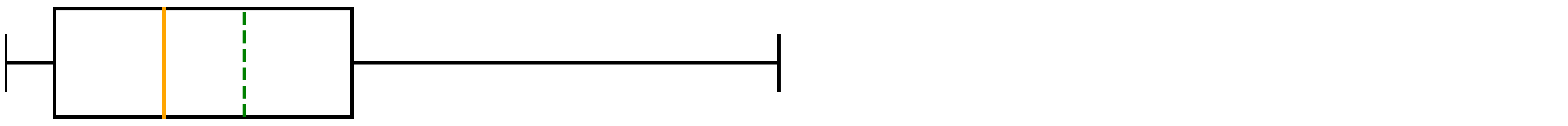} & 0.000 & 0.000 & 0.522 & 0.638 & 0.726 & 0.745 & 0.166 & 0.123 \\
& &  & \includegraphics[width=0.25\textwidth]{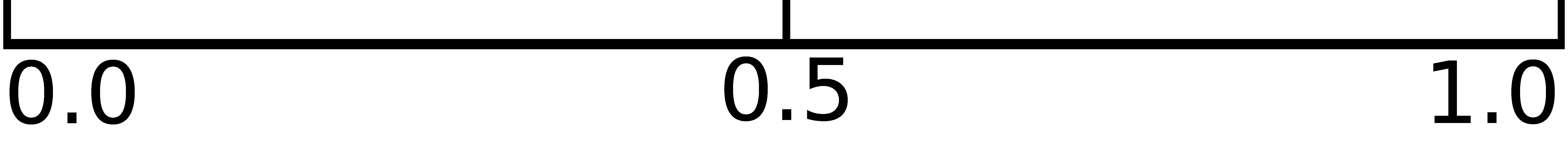} & \includegraphics[width=0.25\textwidth]{figures/main-table/x-lim.png} & \includegraphics[width=0.25\textwidth]{figures/main-table/x-lim.png} & \includegraphics[width=0.25\textwidth]{figures/main-table/x-lim.png} \\
\bottomrule
\multicolumn{15}{l}{
\begin{tabular}[l]{@{}l@{}}
For training and inference time, as well as CPU and GPU memory usage, we respectively report the recorded results on the multivariate dataset NYC and the univariate dataset GAIA. \\ Classification from the methodological perspective: distance-based (\textcolor{orange}{\textbf{orange}}), density-based (\textcolor{magenta}{\textbf{red}}), prediction-based (\textcolor{cyan}{\textbf{blue}}), contrast-based (\textcolor[rgb]{0,0.5,0}{\textbf{green}})\\
\resizebox{3.0mm}{3.0mm}{\myproximity}\hspace{0.5mm}proximity-based, 
\resizebox{3.0mm}{3.0mm}{\myclustering}\hspace{0.5mm}clustering-based, 
\resizebox{3.0mm}{3.0mm}{\mydiscord}\hspace{0.5mm}discord-based, 
\resizebox{3.0mm}{3.0mm}{\mydistribution}\hspace{0.5mm}distribution-based, 
\resizebox{3.0mm}{3.0mm}{\mygraph}\hspace{0.5mm}graph-based,
\resizebox{3.0mm}{3.0mm}{\mytree}\hspace{0.5mm}tree-based,
\resizebox{3.0mm}{3.0mm}{\myencoding}\hspace{0.5mm}encoding-based, 
\resizebox{3.0mm}{3.0mm}{\myforecasting}\hspace{0.5mm}forecasting-based, 
\resizebox{3.0mm}{3.0mm}{\myreconstruction}\hspace{0.5mm}reconstruction-based, 
\resizebox{3.0mm}{3.0mm}{\mycontrast}\hspace{0.5mm}contrast-based
\end{tabular}
}
\end{tabular}}
\end{table*}



\section{Experiments}
\label{EXPERIMENTS}
\subsection{Experimental Setup}
\label{Experimental Setup}
\subsubsection{Datasets and comparison methods}
We utilize all the datasets included in TAB, which encompass 29 multivariate datasets and 1,635 univariate time series. More details are given in Tables~\ref{Multivariate datasets} and \ref{Univariate datasets}. The anomaly types and characteristic types for both multivariate and univariate time series are available in our code repository.

\subsubsection{Implementation details}
\label{Implementation details}
We unify all the inconsistencies mentioned in Section~\ref{sec:Introduction} and describe the process of hyperparameter tuning. In addition, we provide a clear explanation of the experimental environment to ensure the reproducibility of the experiments and the reliability of the results.

\begin{itemize}[left=0.1cm]
\item \textbf{Inconsistent Dataset Splitting Issue:} Regarding dataset splitting, we follow the initial splitting provided by the raw data \textcolor{black}{for both multivariate and univariate data}. If the raw data does not include splitting information, we adopt a consistent splitting method, ensuring a relatively low anomaly rate in the training set. \textcolor{black}{The training and validation set comprises 50\% of the total data, and the test set accounts for 50\%. Additionally, the last 20\% of the training and validation set is extracted as the validation set. Details on ``dataset level'' data splitting for multivariate and univariate datasets are provided in Tables~\ref{Multivariate datasets} and~\ref{Univariate datasets}. The splitting details at the ``time series level'' are provided in the code repository.}
\item \textbf{Drop-last \& Point Adjustment Issues:} During the testing and evaluation process, we abandon drop-last and point adjustment operations.
\item \textbf{Different Thresholds Issue:} Since different TSAD datasets are sensitive to the threshold, to avoid the threshold problem as is mentioned in Section~\ref{sec:Introduction}, \textcolor{black}{we consider threshold the range [0.1, 0.5, 1, 2, 3, 5, 10, 15, 20, 25] for both multivariate and univariate time series. All the numbers are percentages.} Then, we conduct metric calculations at all thresholds and report the best results.
\item \textbf{Inconsistent Algorithm Outputs Issue:} If a algorithm requires the use of a rolling window during testing, we adopt a non-overlapping window rolling approach. Reasons can be found in Section~\ref{Performance comparison of different post-processing methods}.
\item \textbf{Hyperparameter Tuning:} For each method, we adhere to the parameters specified in the original papers. Additionally, we perform a hyperparameter search over multiple sets to ensure fairness---see Table~\ref{Overview of comparison methods.}. We then select the best results from these evaluations to contribute to a comprehensive and fair assessment of each method. \textcolor{black}{The results of evaluations are close to those reported in the original papers of each TSAD method.}
\item \textbf{Experiment Environment:} All experiments are conducted in Python 3.8 using PyTorch~\cite{paszke2019pytorch} and executed on an NVIDIA Tesla-A800 GPU. The training process is guided by L2 loss and utilizes the ADAM optimizer. Initially, the batch size is set to 256, and in cases of insufficient memory (OOM), it can be halved (minimum of 16). To ensure reproducibility and convenience in experimentation, both the datasets and code are available at \url{https://github.com/decisionintelligence/TAB}.
\end{itemize}

\subsubsection{Performance comparison of different post-processing methods}
\label{Performance comparison of different post-processing methods}
\color{black}{Table ~\ref{post-processing} presents the performance results of different methods using window overlapping post-processing methods and window non overlapping post-processing methods on different datasets. We found that in most cases, the two post-processing methods do not affect the performance of the method, but there are still a small number of cases where the performance of the method is affected by the post-processing method. To keep consistent evaluation, we need to select either overlapping or non-overlapping for all the compared methods (non-learning, machine learning, and deep learning methods). However, some traditional methods cannot use window overlapping methods for anomaly detection. Moreover, non-overlapping method is widely used in existing works~\cite{yang2023dcdetector, xu2021anomaly}. Therefore, we finally have adopted window non-overlapping method.}

\begin{table}[ht!]
\caption{Average accuracy for 6 multivariate time series.}
\label{post-processing}
\centering
\resizebox{1\columnwidth}{!}{
\begin{tabular}{c|c|ccccccccccccc}
\toprule
\textbf{Dataset} & \textbf{Metric} & \textbf{ATrans} & \textbf{DC} & \textbf{DLin} & \textbf{NLin} & \textbf{Patch} & \textbf{TsNet} \\ \midrule

\textbf{CalIt2} & Overlap & 0.483 & 0.499 & \underline{0.761} & 0.694 & \uuline{0.791} & \textbf{0.798} \\
 & Non-overlap & 0.491 & 0.527 & \underline{0.752} & 0.695 & \textbf{0.808} & \uuline{0.771} \\ \midrule
\textbf{Daphnet} & Overlap & 0.469 & 0.486 & \underline{0.727} & 0.715 & \uuline{0.739} & \textbf{0.773} \\
 & Non-overlap & 0.489 & 0.501 & \underline{0.728} & 0.715 & \uuline{0.741} & \textbf{0.754} \\ \midrule

\textbf{MSL} & Overlap & 0.494 & 0.502 & \uuline{0.626} & 0.594 & \textbf{0.637} & \underline{0.615} \\
 & Non-overlap & 0.508 & 0.504 & \uuline{0.624} & 0.592 & \textbf{0.637} & \underline{0.613} \\ \midrule

\textbf{PSM} & Overlap & 0.496 & 0.499 & \underline{0.581} & \uuline{0.586} & 0.578 & \textbf{0.589} \\
 & Non-overlap & 0.498 & 0.499 & 0.580 & \underline{0.585} & \uuline{0.586} & \textbf{0.592} \\ \midrule
\textbf{SKAB} & Overlap & 0.495 & 0.532 & \uuline{0.563} & \underline{0.558} & 0.555 & \textbf{0.592} \\
 & Non-overlap & 0.513 & 0.522 & \underline{0.593} & 0.583 & \uuline{0.597} & \textbf{0.620} \\ \midrule
\textbf{SMAP} & Overlap & \textbf{0.504} & \uuline{0.499} & 0.398 & 0.434 & 0.449 & \underline{0.455} \\
 & Non-overlap & \uuline{0.504} & \textbf{0.516} & 0.397 & 0.434 & 0.448 & \underline{0.453} \\


\bottomrule
\multicolumn{8}{l}{\textbf{Bold} represents the best result (AUC-ROC ), \uuline{double Underline} represents the second}  \\
\multicolumn{8}{l}{best result, and \underline{single Underline} represents the third best result.}
\end{tabular}
}
\end{table}




\subsection{Overall Performance}
\label{52Overall Performance}
Due to space limitations, we only report the score distribution of two representative metrics, VUS-PR (V-PR) and Affiliated F1 (Aff-F1), in the main text. For results of other metrics, please refer: \url{https://decisionintelligence.github.io/OpenTS/Benchmarks/overview/}.

\begin{figure}[t]
\centering
\begin{subfigure}{1\linewidth}
    {\includegraphics[width=0.99\textwidth]{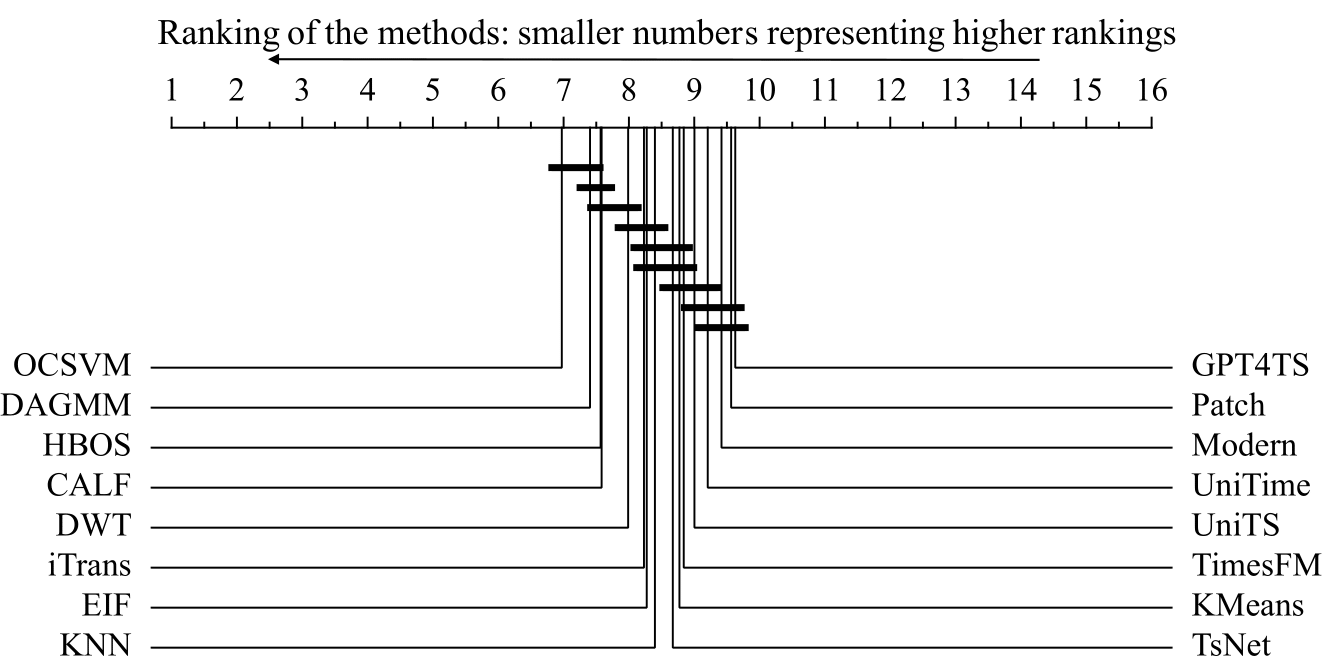}}
    \caption{All univariate datasets: Critical difference diagram ($\alpha$ = 0.05)}
    \label{fig:pattern_seansonal}
\end{subfigure}

\begin{subfigure}{1\linewidth}
    \includegraphics[width=0.99\textwidth]{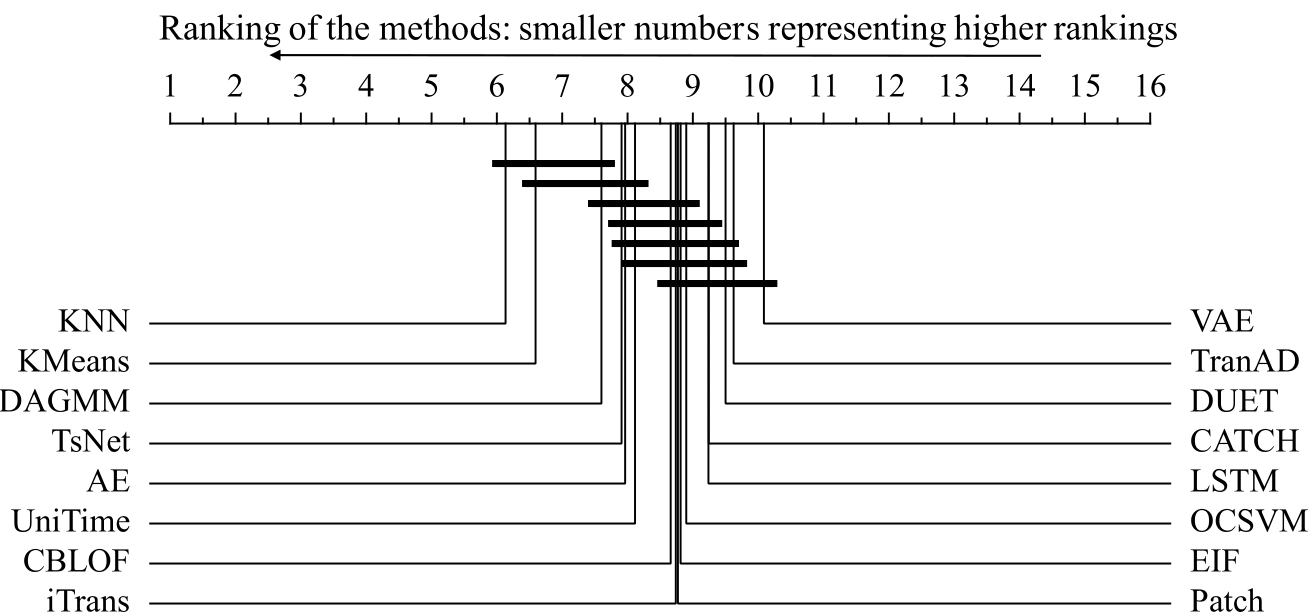}
    \caption{All multivariate datasets: Critical difference diagram ($\alpha$ = 0.05)}
    \label{fig:pattern_trend}
\end{subfigure}
\caption{\textcolor{black}{Illustration of model performance (VUS-PR) on all univariate (a) and multivariate (b) datasets.}}
\label{fig:Critical diagram}
\end{figure}

\subsubsection{Univariate time series anomaly detection}
\label{Univariate time series AD.}
The left part of Table~\ref{main table} presents the results of V-PR and Aff-F1 for 46 methods across 1,635 univariate time series. 
We make four observations: 1) LLM-based and TS pre-trained methods achieve relatively stable average scores and lower bounds on Aff-F1, indicating that they have learned general patterns of time-series data during pre-training, leading to stable performance. Moreover, the differences across the full, few, and zero-shot settings are small, suggesting that the patterns in univariate time series are relatively simple, making them easier to fit and generalize. 2) Traditional DL methods exhibit performance closely tied to their compatibility with the datasets. For example, iTrans and DLin show higher average V-PR and Aff-F1 scores but have larger fluctuations, implying that they excel at detecting specific anomalies but perform less well on others. There are also methods, notably DualTF, ConAD, and DC, that seem to struggle with anomaly detection in univariate time series, leading to overall poor performance. 3) Machine learning (ML) and non-learning (NL) methods exhibit the best average performance in terms of the V-PR and Aff-F metrics. This not only reflects the meaning of traditional methods in theoretical foundations and practical applications but also highlights their significant value and competitiveness at solving modern problems. These results indicate that while pursuing novel methods, we should not overlook the classic methods. When combined with domain knowledge and real-world requirements, traditional methods often demonstrate remarkable performance. \textcolor{black}{4) When considering the different classes of methods, the prediction-based methods (such as CALF, iTrans, and UniTS) usually show relatively better performance than other classes on univariate TSAD tasks. Next flow the density-based methods (such as DWT, S2G, and EIF). The contrast-based methods (such as DC, ConAD) perform the weakest.}



\subsubsection{Multivariate time series anomaly detection} The right of Table ~\ref{main table} presents the results of 40 methods on the 29 public multivariate datasets. 
We have the following observations: 1) Under full-shot and few-shot learning scenarios, time series pre-trained models demonstrate significantly better performance compared to zero-shot learning approaches. This could be because multivariate time series are more complex, and during the pre-training phase, large models tend to capture only basic features and fail to capture the interactions between channels. As a result, fine-tuning leads to more significant improvements by allowing the model to learn these relationships more effectively. 2) Some deep learning models, such as TsNet and CATCH, appear to achieve the best performance in full-shot settings. This may be due to the significant impact of channel interactions in multivariate anomaly detection, so training a model with a reasonable number of parameters for specific datasets helps capture these interactions without overfitting, thereby outperforming even large pre-trained models in the full-shot setting. 3) While deep learning-based methods show promising results, methods based on Non-Learning and Machine Learning, such as KNN, KMeans, and OCSVM, demonstrate exceptional performance and stability in V-PR and Aff-F1. This suggests that there is still significant room for improvement in current deep learning approaches.
\begin{figure}[t]
    \centering
    \includegraphics[width=1\linewidth]{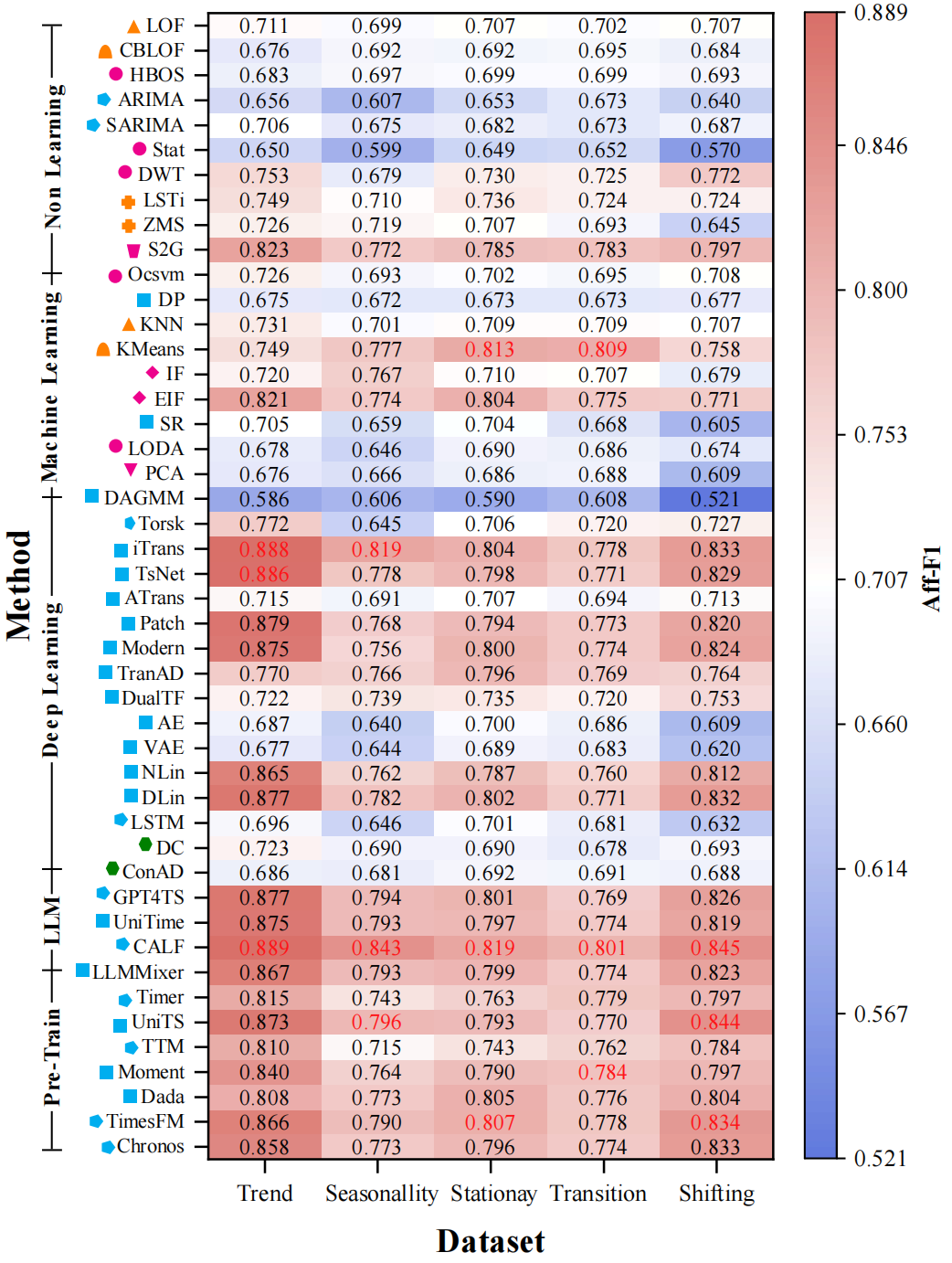}
    \caption{\textcolor{black}{Classification results (Aff-F1) of univariate datasets based on characteristic.}}
    \label{fig:Classification results of univariate datasets based on features.}
\end{figure}

\begin{figure*}[t]
\centering
\begin{subfigure}{0.3\linewidth}
    \includegraphics[width=1.0\textwidth]{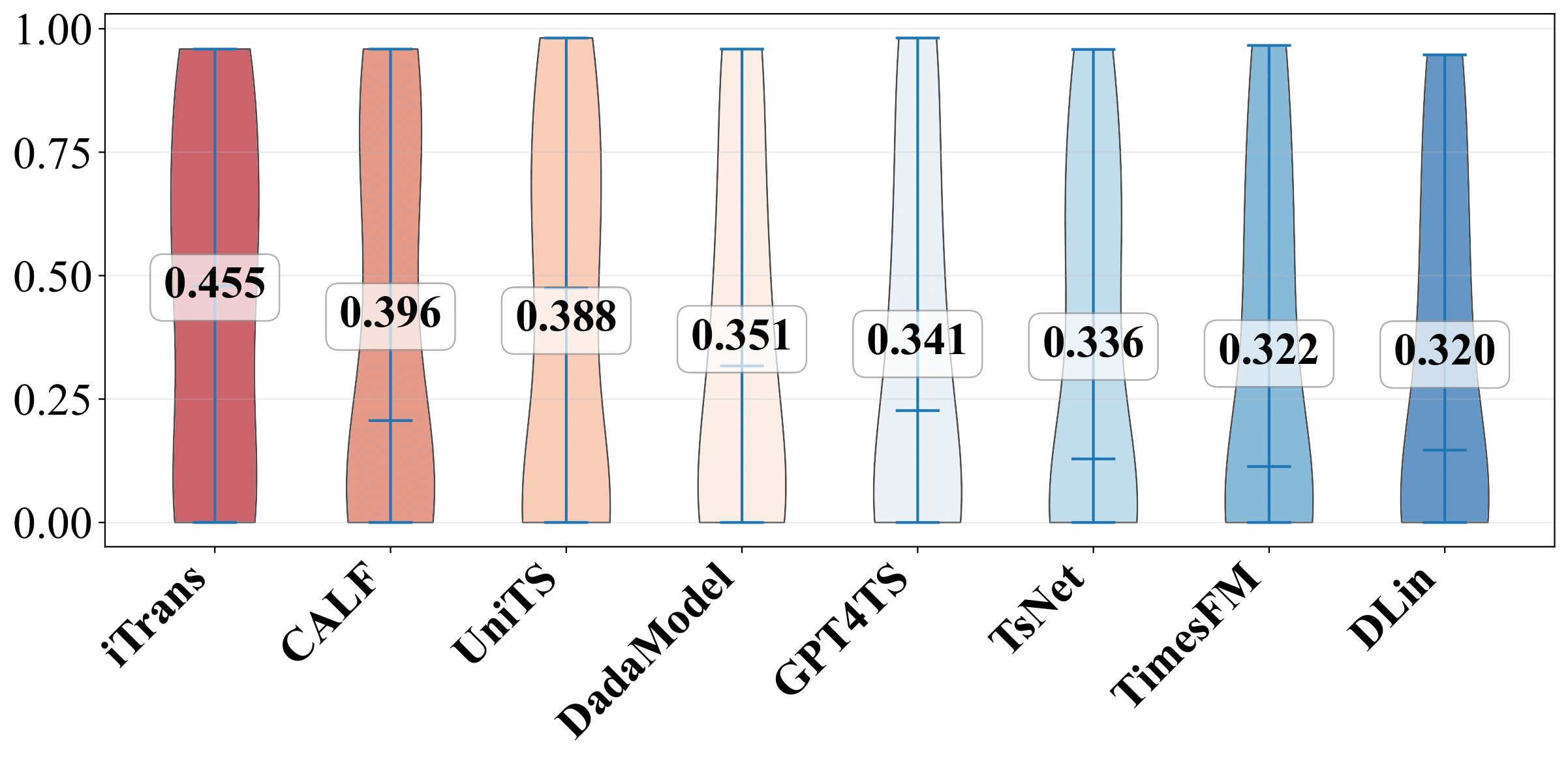}
    \caption{Global}
    \label{fig:pattern_trend}
\end{subfigure}
\begin{subfigure}{0.3\linewidth}
    \includegraphics[width=1.0\textwidth]{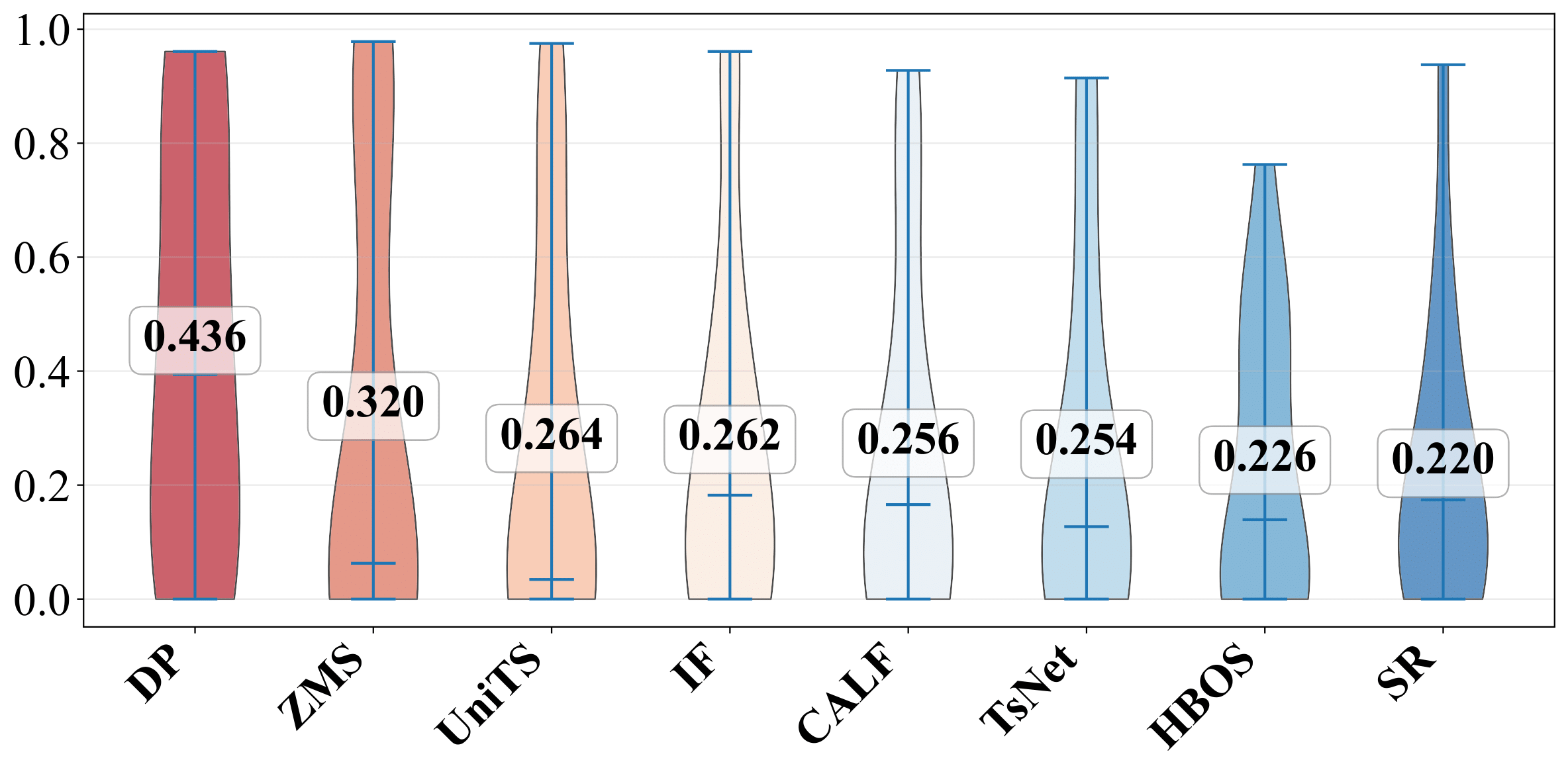}
    \caption{Contextual}
    \label{fig:pattern_seansonal}
\end{subfigure}
\begin{subfigure}{0.3\linewidth}
    \includegraphics[width=1.0\textwidth]{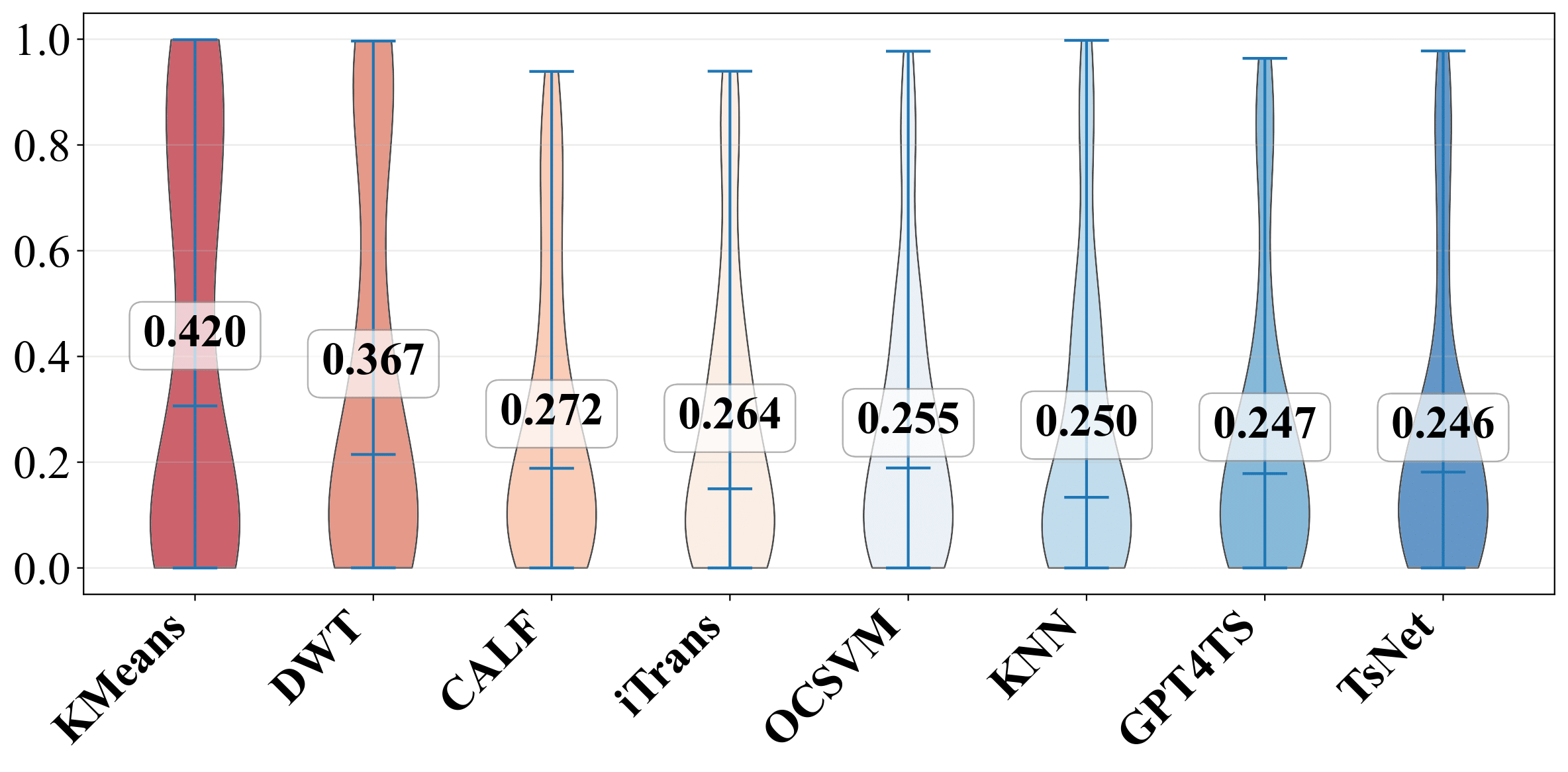}
    \caption{Shapelet}
    \label{fig:pattern_shift}
\end{subfigure}

\vspace{1mm}
\begin{subfigure}{0.3\linewidth}
    \includegraphics[width=1.0\textwidth]{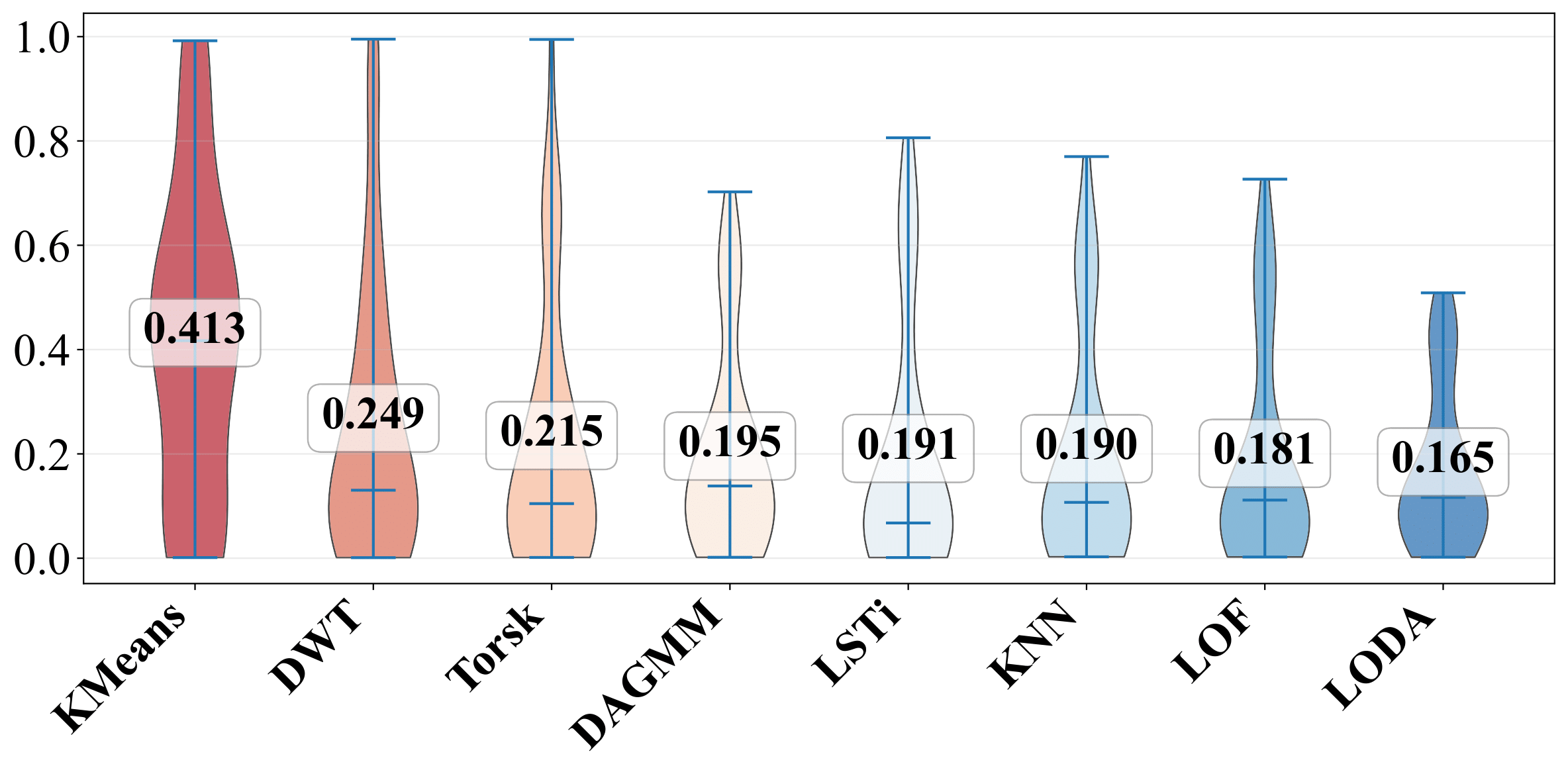}
    \caption{Seasonal}
    \label{fig:pattern_shift}
\end{subfigure}
\begin{subfigure}{0.3\linewidth}
    \includegraphics[width=1.0\textwidth]{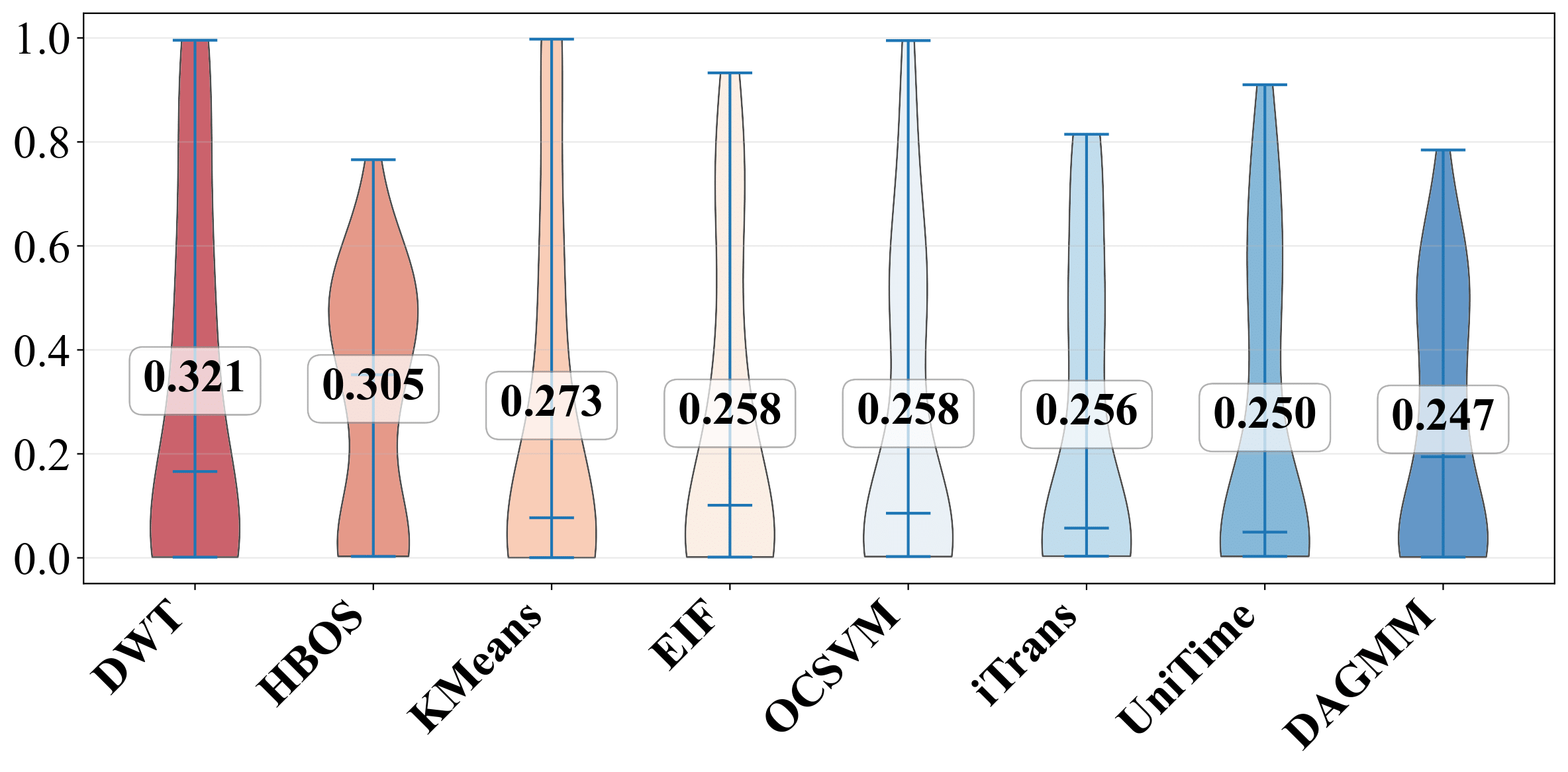}
    \caption{Trend}
    \label{fig:pattern_shift}
\end{subfigure}
\begin{subfigure}{0.3\linewidth}
    \includegraphics[width=1.0\textwidth]{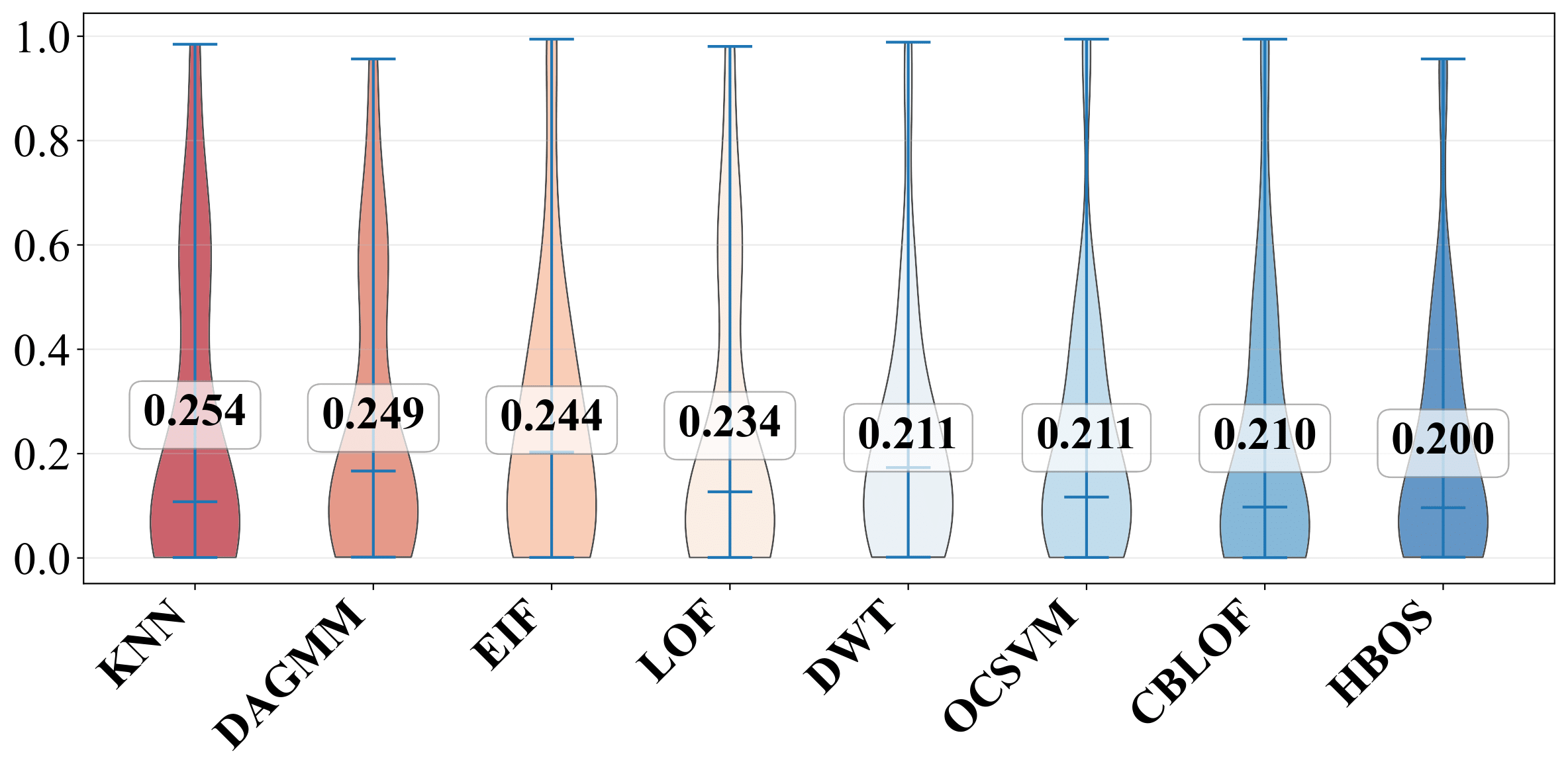}
    \caption{Mix}
    \label{fig:pattern_pattern}
\end{subfigure}

\caption{\textcolor{black}{Classification of univariate datasets based on anomaly type, including the 8 algorithms with the best average performance (VUS-PR) for each category.}}
\label{fig:pattern_eg}
\end{figure*}


\subsubsection{Statistical validation}
\textcolor{black}{To validate the statistical differences in the performance results for the automated solutions, we employ the Wilcoxon test~\cite{wilcoxon1992individual} to obtain pairwise comparisons across datasets. Additionally, to compare multiple automated solutions across datasets, we use the Friedman test~\cite{friedman1937use} followed by the post-hoc Nemenyi test~\cite{nemenyi1963distribution} at a 95\% confidence level. We plot the critical difference (CD) diagrams, where the numbers at the top indicate the ranking of the methods, with smaller numbers representing higher rankings. Methods without significant performance differences are connected by horizontal lines. Specifically, since we cover 48 methods, displaying them all would result in overly dense lines, which would reduce readability. Therefore, we now only present the top 16 ranked methods and have added a legend to ensure that the information is displayed more clearly and intuitively. As shown in Figure \ref{fig:Critical diagram}, the non learning and machine learning methods exhibit particularly strong performance on the univariate time series anomaly detection tasks, with OCSVM, HOBS, and DWT achieving the excellent results. On the multivariate datasets, the deep learning and machine learning methods demonstrate clear advantages, with models such as KNN, KMeans, DAGMM, TsNet, and AE showing excellent performance.}



\subsection{Method Recommendations}
\label{53Method Recommendations}
To study the performance of different methods over different anomaly types and data characteristics, 
we further classify all the univariate time series according to anomaly type and data characteristic, respectively. \textcolor{black}{The quantity distribution of univariate datasets under six different anomaly categories and five time series characteristics is not uniformly distributed since the number of open-source univariate time series datasets in real scenarios is limited. But we have tried our best to make them as evenly distributed as possible under different categories. For anomaly types, we use manual expert annotation to label them. Specifically, the type of anomaly for each time series is determined by a vote of five experienced manual experts in time series anomaly detection.}

%

\subsubsection{Analysis of data characteristics}

To assess performance according to different data characteristics,
Figure~\ref{fig:Classification results of univariate datasets based on features.} reports the average results on datasets with distinct characteristics in terms of the metric Aff-F1. We make the following observations: 
1)~Most methods perform better on time series with \textit{trend}, \textit{stationarity}, and \textit{shifting} characteristics. Time series with \textit{seasonality} and \textit{transition} characteristics make it difficult to detect anomalies. \textcolor{black}{2) LLM-based and time-series pre-trained models typically exhibit good performance on most data with varying characteristics. Notably, they perform well on data with trend and shifting characteristics. Density-based methods, like LODA and PCA, perform a bit worse across most characteristics. Contrast-based methods, including DC and ConAD, exhibit relatively weak performance across all characteristics. This suggests that this type of method has poor adaptability across univariate TSAD tasks. Prediction-based methods, such as iTrans, TsNet, CALF, and UniTS, frequently perform well on data with trend and seasonality characteristics.}

\subsubsection{Analysis on anomaly types}


\textcolor{black}{To assess performance across anomaly types, we partition the time series into six sets according to their anomaly type, i.e., contextual, global, shapelet, seanonal, trend, and mix. \textcolor{black}{Global and contextual anomalies are point anomalies, while trend, shapelet, and seasonal anomalies are subsequence anomalies.}
Figure~\ref{fig:pattern_eg} shows the VUS-PR score distribution of different methods on datasets with different anomaly types. Our observations for the different anomaly types are as follows: 1) Among the six anomaly types, the global point anomalies are the easiest to detect, with methods achieving relatively high VUS-PR values. In contrast, mixed anomalies are the most challenging to identify. 2) Deep learning methods (e.g., iTrans and DP), as well as foundation methods (LLM-based and TS pre-trained methods) like CALF and UniTS, are more suitable for finding point anomalies (including global and contextual anomalies). 3) Non learning methods (e.g., DWT, LOF and HBOS), along with machine learning methods (e.g., KMeans and KNN), are most suited for finding subsequence anomalies (including shapelet, seasonal, and trend anomalies). 4) Density-based methods such as DWT, HBOS, EIF, and OCSVM are appropriate for finding trend anomalies. 5) Distance-based methods (e.g., KMeans and KNN) are suitable for finding shapelet anomalies.}

When given an unseen time series, five formulas in our code repository can be used to calculate the five characteristic values of the time series. Then, based on the relationship between method performance and dataset characteristics described above, users can obtain a better understanding of which method to choose. These observations aid in making more targeted choices and optimizations of methods to better adapt to diverse application scenarios and data characteristics.





\subsection{Trade-off among Accuracy, Memory, and Runtime}
\label{Trade-off Between Accuracy, Cost, and Time}
\textcolor{black}{When applying methods in real-world scenarios, it is often necessary to strike a balance among accuracy, memory, and runtime. Each of these factors plays a crucial role in determining the feasibility and practicality of a method. Table~\ref{main table} presents a comparative analysis of accuracy (V-PR, Aff-F1), memory (CPU and GPU memory usage), and runtime (training and
inference time) for different methods on the multivariate and univariate datasets. }

\textcolor{black}{It can be observed that \textbf{in terms of runtime}, specifically regarding training time, many non learning methods (S2G, LSTi, DWT, etc.) and all TS pre-trained methods \textcolor{black}{under the zero-shot evaluation strategy} are the fastest since they do not require training. The training times for the machine learning methods are longer than those of the non learning methods. The training times for most deep learning methods are relatively long. The training times for the TS pre-trained methods under the full-shot evaluation strategy are the longest, and the training times for LLM-based methods are only shorter than those of TS pre-trained methods. Next, the inference times of the methods in these five categories are very short, with non-learning methods being the fastest, such as HBOS and CBLOF. \textbf{In terms of memory} and specifically regarding GPU memory usage, most non learning methods and machine learning methods do not need to use the GPU, so their GPU memory usage is 0. \textcolor{black}{The methods that consume the most GPU memory are the LLM-based methods because they have a huge number of parameters. Next are the TS pre-trained methods, followed by the deep learning methods.} Next, the CPU memory usage of methods is similar across the five categories. Among them, the usage by the non learning methods is the lowest.} \textcolor{black}{Considering accuracy, memory, and runtime together, for univariate anomaly detection, some non learning methods are good choices, notably S2G, DWT, and KMeans among the machine learning methods. However, most non learning methods do not support multivariate anomaly detection, and the performance of most machine learning methods at multivariate anomaly detection is not satisfactory. For multivariate anomaly detection, if the requirements for runtime and memory are relatively lenient, lightweight deep learning methods can be chosen, notably CATCH and TsNet.}

\section{Conclusions}
\label{sec:Conclusions}

To enable accurate comparison of various TSAD methods, this paper introduces TAB, a unified time series anomaly detection benchmark. First, TAB includes 1,635 univariate time series and 29 multivariate datasets, spanning multiple domains. Second, TAB incorporates state-of-the-art time series anomaly detection methods, including non-learning, machine learning, deep learning, LLM-based, and time series pre-trained approaches. Third, TAB provides a unified and extensible evaluation pipeline for time series anomaly detection, simplifying the process of evaluating TSAD methods and enabling fair comparisons through a consistent experimental setup. Finally, we use TAB to evaluate the performance of 40 multivariate TSAD methods and 46 univariate TSAD methods on all the datasets, employing a wide range of evaluation strategies and metrics. Additionally, we have also launched an online TSAD leaderboard that covers LLM-based and time series pre-trained methods.



\begin{acks}
This work was partially supported by National Natural Science Foundation of China (62472174 and 62372179) and Huawei Cloud Availability Engineering Lab. Jilin Hu is the corresponding author of the work.
\end{acks}

\clearpage

\renewcommand{\arraystretch}{1}

\bibliographystyle{ACM-Reference-Format}
\bibliography{sample}

\end{document}